# Natural Language Processing for Electronic Health Records in Scandinavian Languages: Norwegian, Swedish, and Danish

*Ashenafi Zebene Woldaregay [a], PhD, Jørgen Aarmo Lund [c, j, 1], MSc, Phuong Dinh Ngo [g], PhD, Mariyam Tayefi [g], PhD, Joel Burman [a], MSc, Stine Hansen [a], PhD, Martin Hylleholt Sillesen [i], PhD, Hercules Dalianis [g, h], PhD, Robert Jenssen [c, d, e,1,2], Lindsetmo Rolf Ole [b, f], PhD, Karl Øyvind Mikalsen [a, c, f ,1], PhD*

[a] Norwegian Centre for Clinical Artificial Intelligence (SPKI), University Hospital of Northern Norway, Tromsø, Norway
[b] Clinic of Surgery, Oncology and Women Health, University Hospital of North Norway, Tromso, Norway
[c] Department of Physics and Technology, UiT The Arctic University of Norway, Hansine Hansens veg 18, 9019 Tromsø, Norway
[d] Department of Computer Science, University of Copenhagen, Universitetsparken 1, Copenhagen 2100, Denmark
[e] Norwegian Computing Center, Gaustadalleen 23a, 0373 Oslo, Norway
[f] Department of Clinical Medicine, UiT – The Arctic University of Norway, Tromsø, Norway
[g] Norwegian Centre for E-health Research, Tromsø, Norway
[h] Department of Computer and Systems Sciences, Stockholm University, Stockholm, Sweden
[i] Department of Surgery and Transplantation C-TX, Copenhagen University Hospital, Rigshospitalet, Copenhagen, Denmark
[j] DIPS AS

## Background:

Clinical natural language processing (NLP) refers to the use of computational methods for extracting, processing, and analyzing unstructured clinical text data, and holds a huge potential to transform healthcare in various clinical tasks. The advancement of deep learning, augmented by the recent advent of transformers, has been pivotal in the success of NLP across multiple domains. This success is largely attributed to deep learning systems' end-to-end training capabilities, allowing models to automatically learn features from large amounts of data. Further, recent advancements in instruction tuning have led to the development of Large Language Models (LLMs), e.g. OpenAI's GPT, which can perform tasks described in natural language, and holds promising potential. While these advancements have dramatically enhanced capabilities in processing languages like English, these benefits are not always equally transferable to under-resourced languages. In this regard, this review aims to provide a comprehensive assessment of the state-of-the-art NLP methods for the mainland Scandinavian clinical text, thereby providing an insightful overview of the landscape for clinical NLP within the region.

## Objective:

The study aims to perform a systematic review to comprehensively assess and analyze the state-of-the-art NLP methods for the Scandinavian clinical domain, thereby providing an overview of the landscape for clinical language processing within the Scandinavian languages across Norway, Denmark, and Sweden. Generally, the review seeks to provide a practical outline regarding various modeling options, opportunities, and challenges or limitations, thus, providing a clear overview of existing methodologies and potential avenues for future research and development.

## Method:

A literature search was conducted in various online databases including PubMed, ScienceDirect, Google Scholar, ACM digital library, and IEEE Xplore between December 2022 and February 2023. Further, relevant references to the included articles were also used to solidify our search. Additionally, an updated search was performed between February and March 2024 to incorporate new articles released after the initial search. The search considers peer-reviewed journal articles, preprints, and conference proceedings. Relevant articles were initially identified by scanning titles, abstracts, and keywords, which served as a preliminary filter in conjunction with inclusion and exclusion criteria. After the initial screening and removal of duplicates, the remaining articles went through a full-text eligibility assessment, and, finally, relevant studies were incorporated. The final pool includes articles that conducted clinical NLP in the mainland Scandinavian languages and were published in English between 2010 and 2024. Data was extracted according to predefined categories, established from prior studies and further refined through brainstorming sessions among the authors.

* Corresponding author.
email address: ashenafi.zebene.woldaregay@unn.no

[1] Work done while at the UiT Machine Learning Group, machine-learning.uit.no.
[2] Work done while at Visual Intelligence, www.visual-intelligence.no.

## Results:

The initial search yielded **217** articles. Following the identification stage, **172** articles remained following the removal of duplicates and studies that were outside the study scope. After the initial screening stage, **134** studies remained, which went through full-text eligibility. The full-text eligibility assessment was independently carried out by five of the authors and resulted in **105** studies. After performing an updated search, 8 more articles were included, resulting in a total of **113** articles, which were critically analyzed in the study. Any disagreements among the authors were resolved through discussion. Out of the 113 articles, 18% (n=21) focus on Norwegian clinical text, 64% (n=72) on Swedish, 10% (n=11) on Danish, and 8% (n=9) focus on more than one language. Generally, the review identified positive developments across the region despite some observable gaps and disparities between the languages. Concerning publication trends and co-occurrence, a significant number of studies focus on Swedish clinical text compared to the others, with minimal efforts to study these languages together. There are substantial disparities in the level of adoption of transformer-based models. Research focusing on Swedish and Danish clinical text has shown satisfactory levels, while activities focusing on Norwegian clinical text are minimal. In essential tasks such as de-identification, there is significantly less research activity focusing on Norwegian and Danish compared to Swedish text. Further, the review identified a low level of sharing resources such as data, experimentation code, pre-trained models, and rate of adaptation and transfer learning in the region. Additionally, there are very few efforts to validate models in clinical settings. Regarding active medical expert involvement, the review identified a satisfactory level though the rates vary among the three languages.

## Conclusion:

The review presented a comprehensive assessment of the state-of-the-art Clinical NLP for electronic health records (EHR) text in mainland Scandinavian languages and, highlighted the potential barriers and challenges that hinder the rapid advancement of the field in the region. The review identified a lack of shared resources, such as datasets, pre-trained models and tools, inadequate research infrastructure, and insufficient collaboration as the most significant barriers that require careful consideration in future research endeavors. For example, the disparity in research activities among the three languages underscores the importance of robust research infrastructure, as demonstrated by the significant impact of the research infrastructure "*Swedish Health Record Research Bank*" (Health Bank). Given the low-resource nature of these languages, more research activities should be targeted toward resource development, core NLP tasks, and de-identification. Generally, we foresee the findings presented in this review will help shape future research directions by shedding some light on areas that require further attention for the rapid advancement of the field in the region

# 1. Introduction

The adoption of Electronic Health Record (EHR) systems in hospitals has enabled large-scale collection and digitization of healthcare data, allowing population health analysis and predictive modeling with novel and potentially transformative applications in healthcare (1-3). While EHR systems generally attempt to store data in structured and standardized formats, unstructured free-text notes still make up significant portions of EHR records. These notes afford users the necessary flexibility to provide comprehensive and relevant information for other healthcare workers, especially in complex care situations where preexisting guidelines are challenging to apply. However, they also contain valuable clinical information that can serve as validation for other healthcare data sources, aid for EHR data entry and administrative duties, and support for clinical decision-making. Inferring this information is the main goal of research into clinical NLP methods, which apply computational methods to analyze and process unstructured clinical text (4).

The last decade of NLP research has seen the breakaway success of *deep learning* methods, which forgo the rule and feature design involved in traditional NLP methods, in favor of having the model itself discover semantic and syntactic relationships in large corpora of unannotated text. One key event was the publication of the Transformer network design in 2017, which builds on *attention* blocks that allow the networks to capture complex dependencies between terms. Transformer networks still form the backbone of state-of-the-art NLP models and have been shown to perform well in various clinical NLP tasks (17-20), including information extraction (18, 21-23), classification (24-27), and context analysis (28-30). These networks also benefit from training with domain-specific corpora: models specifically developed for clinical text, such as ClinicalBERT (31) and BlueBERT (146), outperform equivalent general-purpose models in information extraction and other clinical NLP tasks.

Recent research into instruction tuning has allowed the development of Large Language Models (LLMs) capable of performing tasks specified in natural languages, such as the OpenAI GPT models, the Meta LLaMa models, and Google's Gemini (5-7). The ability to perform tasks without specifically annotated training corpora has led to a groundswell of proposed applications for LLMs, with healthcare being no exception: LLM-based solutions have been proposed for clinical tasks such as patient triage and treatment plan generation, as well as documentation tasks such as summarization and clinical narrative structuring (8-10). However, the disconnect between the proposed tasks and the LLMs' language modeling objectives introduces its own challenges, such as "hallucinations" arising from the model picking superficially probable but incorrect answers. The scale of the models also introduces privacy and data security concerns.

However, despite the favorable conditions for general NLP research, the clinical domain presents unique challenges for developing and implementing clinical NLP. Data access is the most readily apparent challenge: clinical notes contain uniquely sensitive information about patients, along with potentially directly personally identifying details, requiring access control and privacy-preserving measures to preserve patient privacy (11). The clinical notes themselves are shaped by the realities of healthcare work: as a means of communicating with other healthcare professionals, often containing jargon and abbreviations specific to the profession, as well as misspellings due to time pressure, making it more difficult to apply methods developed and evaluated with general NLP datasets (11). Depending on the task, getting data annotated may require medical expertise, requiring effort from healthcare workers already under a heavy workload. Beyond model development, challenges also arise when implementing NLP methods in clinical settings. On a technical level, clinical NLP models must work reliably at scale, and with minimal latency, especially in applications directly related to patient treatment. Operationally, the NLP applications must integrate into existing workflows, ideally as seamlessly as possible. From the user's perspective, uninterpretable "black box" models can be perceived as untrustworthy, leading to lower adoption and use (11, 12).

While deep learning methods have drastically enhanced language processing capabilities in high-resource languages like English, Spanish, and German, these capabilities do not necessarily transfer equally to low-resource languages, where large corpora of unannotated text are not as readily available (13). Prior reviews have examined clinical NLP in languages other than English (13, 14), but these reviews cover a wider range of languages, often unrelated to each other, and primarily focus on NLP method development over clinical applications. To the best of our knowledge, this is the first comprehensive review covering the development of clinical NLP for the mainland Scandinavian languages – Danish, Norwegian, and Swedish – which are linguistically related, and broadly similar in vocabulary and grammar (15-17). These languages are among the most under-resourced, especially compared to resource-rich languages like English. This review aims to comprehensively document the state-of-the-art and the developments in clinical NLP methods for the mainland Scandinavian languages, while also providing a practical outline of existing NLP methods and future research opportunities. To provide a better picture of the overall research trends, the review examined four major topics:

I. *Publication insight and trends*: examines the number of publications, co-occurrence of languages, and collaboration of authors among each respective country,
II. *Data characteristics*: examines the type of data sources, reported data quality issues, access to data for public use, and the type of EHR documents targeted in the study,
III. *NLP techniques*: examines NLP techniques employed in the studies, including the availability of codes and pre-trained models,
IV. *Clinical characteristics*: examines the clinical settings the studies were conducted, including the specific healthcare domain targeted, and whether medical professionals were involved in developing or evaluating the method.

Generally, the review identified that the number of publications per year on clinical NLP for mainland Scandinavian languages is trending upwards, aside from a peak in 2014 and a lull between 2018 and 2019. However, the research task focus differs significantly across languages, leaving, for instance, less published research into de-identification for Norwegian text than for Swedish. Despite the similarities between the mainland Scandinavian languages, only a few of the reviewed studies applied adaptation or transfer learning techniques, and only one reviewed study incorporated all three languages. Under a third of the reviewed studies (31.9%) make their datasets available to other researchers, but even fewer studies (18.6%) make their experiment code or NLP models available.

# 2. Methods

The study aims to perform a systematic review to comprehensively assess and analyze the state-of-the-art natural language processing methods for the Scandinavian clinical domain, thereby providing an insightful overview of the landscape for clinical language processing within the mainland Scandinavian languages: Norwegian, Danish, and Swedish. Further, the review seeks to provide a practical outline regarding various modeling options, opportunities, and challenges or limitations, thus, providing a clear overview of existing methodologies and potential avenues for future research and development. To this end, a literature search was conducted in the research aggregators PubMed, ScienceDirect, Google Scholar, ACM digital library, and IEEE Xplore between December 2022 and February 2023. Additionally, an updated search was performed between February and March 2024 to incorporate new articles released after the initial search. Further, relevant references to the included articles were also used to solidify our search. In addition, research outputs from distinguished researchers from our networks were also used. The search considered peer-reviewed journal articles, preprints, and conference proceedings. Different search strings were used including "*clinical decision support*", "*natural language processing*", "*Norway*", "*Norwegian*", "*Danish*", "*Denmark*", "*Swedish*", "*Sweden*", "*artificial intelligence*", "*machine learning*", "*deep learning*", "*transformers*", "*neural network*", "*clinical text*", "*clinical notes*", "*EHR*", *and* "*clinical NLP*". Further, these strings and sub-strings were combined using "*OR*", *and* "*AND*" for a better search strategy. Relevant articles were initially identified by scanning titles, abstracts, and keywords, which served as a preliminary filter in conjunction with inclusion and exclusion criteria. After removing the duplicates, studies that passed the initial screening went through a full-text eligibility assessment, and, studies fulfilling the inclusion criteria were encompassed. Studies that perform clinical NLP in the mainland Scandinavian languages and were published between 2010 and 2024 in the English language are included in the study. The process was carried out and reported according to the widely used PRISMA guidelines for reporting systematic review (18). Information was extracted from the included articles using predefined categories established through a series of brainstorming sessions among the authors.

## 2.1. Inclusion and exclusion criteria:

Studies were included in the review if and only if they met the following criteria:

- Studies should use natural language processing for extracting, processing, or analyzing medical texts written in Norwegian, Swedish, or Danish. Clinical text in our context encompasses but is not limited to, electronic health records (EHR) (e.g., progress notes, lab reports, pathology reports, discharge summaries, and others).
- Studies should be published between 2010 and 2024.
- Peer-reviewed journal articles, preprints, and conference proceedings are included.
- Studies published in English.

Studies that fell outside the defined scope, including those limited to only structured data, were excluded.

## 2.2. Categorization and Data Collection

This section describes the categories used in extracting information from the articles included in the study. These categories are solely determined to assess and evaluate the included articles and are pre-defined based on previous knowledge, brainstorming, discussion among the authors, and literature (4, 13). A more detailed description of these categories can be found in Appendix A and the result section. These variables were selected so as to capture an overview of the clinical NLP research dynamics and assist in analyzing trends, identifying gaps, and understanding research directions. The definition of other relevant categories, which were extracted but not directly used in the literature analysis, can also be found in Appendix A.

- ***Clinical Application and Technical Objective***: The clinical challenges or healthcare goals the study intends to address, and, if there is any technical innovation being pursued from an NLP perspective.
- ***Author's country and Date of publication:*** The author's country of affiliation, and article publication date.
- ***Language***: Language of the clinical text processed in the study.
- ***NLP Task***: High-level categorization of the NLP task implemented in the study.
- ***Type Description***: Specific categorization of the NLP tasks implemented in the study.
- ***Model and Method***: Family and type of model either used or developed in the study.
- ***Code Availability, and pre-trained models*:** Whether the study had released any accompanying code or pre-trained models related to the experiments conducted.
- ***Adaptation & Transfer learning***: Whether the study employed adaptation and transfer learning such as data augmentation, or domain adaptation.
- ***Data Sources and Names***: The origin and name, where available, of the dataset employed in the study.
- ***Other data*:** Non-free text data utilized in the study along with free text data.
- ***Type of EHR documents***: Where reported, the specific type of EHR documents utilized in the study.
- ***Feature Representation Techniques***: The specific methods employed to convert the preprocessed text data into a numerical representation that the NLP models can understand and analyze.
- ***Disease area/Health domain*:** The disease or health domain targeted by the study, wherever possible, roughly mapped onto the classification defined by the Clinical Data Interchange Standards Consortium (19).
- ***Departments/Clinical Unit*:** The particular healthcare department or clinical unit from which the study dataset was retrieved.
- ***Data Availability*:** Accessibility of the dataset, whether public, restricted to certain users, or available upon request.
- ***Domain Expert*:** Involvement of clinical domain experts in dataset preparation or analysis process.
- ***Data quality and related issues***: Type of data quality and other related issues reported in the study and measures to mitigate them.
- ***Implementation Status*:** State of deployment and integration, or trial in clinical settings.

Based on the categories defined above, five authors (*AZW, PDN, MT, JB, and JAL*) independently performed the data extraction process. To mitigate bias, all authors carried out random checking of each other's results and as a final confirmation, *AZW* performed a thorough inspection of the team's results. Any ambiguity that arose at this stage was resolved through discussion.

## 2.3. Literature Analysis and Evaluation

Literature evaluation was carried out based on the predefined categories mentioned above to provide a comprehensive overview of the landscape for state-of-the-art clinical NLP in mainland Scandinavian languages: Swedish, Norwegians, and Danish. For the sake of brevity, the analysis was conducted by grouping these predefined categories into four groups: *Publication Insight and Trends*, *Clinical Characteristics*, *NLP Techniques*, and *Data Characteristics*. It is worth noting that evaluations of certain features in the analysis could exceed the total number of articles reviewed, given the fact that an article could have more than one contribution.

Further detailed information regarding how the specific values in a category were grouped into sub-categories can be found in Appendix A.

***Publication Insight and Trends***: This group delves into these three predefined categories – *language, author's country, and date of publication*. It focuses on analyzing the general publication trends across the three languages over the years, examining the co-occurrence of languages within publications, and exploring collaborations between authors from the three mentioned countries and others, thereby providing a broader understanding of the research dynamics, language use, and international cooperation in the field.

***Clinical Characteristics***: This group delves into these four predefined categories - *disease area/health domain, departments/clinical unit, domain expert and annotation, and implementation status*, thereby providing analysis and insight into the overall focus of the included studies regarding the type of disease and department or clinical unit considered, and the extent of pilot testing or other forms of evaluation performed within the clinical settings along with the extent of experts involvement in data annotation. For the sake of brevity and ease, the disease area/health domain was further grouped into hematologic and immune disorders, cardiovascular disorders, gastrointestinal disorders, behavioral and lifestyle factors, infectious diseases, cancer and systemic disorders, musculoskeletal and connective tissue disorders, neurological and psychiatric disorders, general conditions and procedures, surgical and postoperative complications, family and genetic history, specific conditions (Appendix A). By the same token, the departments/clinical units were further grouped into general services, intensive and critical care, surgical specialties, medical specialties, oncology and hematology, diagnostic and therapeutic services, women's health, pediatrics, other specialties, and administrative (Appendix A).

***NLP Techniques***: This group thoroughly examines six predefined categories - *NLP task, type description, feature representation techniques, model and methods, code availability, and adaptation & transfer learning*, thereby providing detailed analysis and insight into the specific NLP task addressed, the type of methodology and feature representation employed for the specific task under consideration, the availability of resource for replicating or building upon the work, and the adaptation of existing pre-trained models to the language of interest. For the sake of brevity and ease, the *model and methods* reported in the included studies are categorized into rule-based systems, interpretable methods, ensemble methods, traditional/statistical learning methods, deep/nonlinear learning methods, hybrid methods, transformers and large language models (Appendix A). By the same token, the feature representation techniques reported were categorized into Statistical features, Orthographic word features, Ontologies, Learned word-level embeddings, and Learned sentence/document embeddings (Appendix A).

***Data Characteristics*****:** This group thoroughly examines five predefined categories - *data sources and names, type of EHR documents, language, data availability, and data quality and related issues*, thereby providing detailed analysis and insight into the complexity and diversity of the datasets used in clinical NLP research. This examination aids in understanding the scope of data types, the linguistic variety, accessibility for research purposes, and potential challenges related to data integrity and reliability, crucial for advancing NLP applications in healthcare. For the sake of brevity and ease, the data quality and related issues reported in the studies are mapped into one or more of the following categories: misspelled, multi-word, abbreviation, reference resolution, lexical variation, incomplete data, inconsistent data, noisy data, duplicate data, formatting issues, handwritten text, data sparsity, lack of public data and others (Appendix A).

# 3. Results

## 3.1. Relevant literature

The initial search, conducted by the first author *AZW*, yielded **217** articles. Following the identification stage, *AZW* removed duplicates and studies that resided outside of the study scope, resulting in **172** articles. Further, *AZW* performed the initial screening stage resulting in **134** articles, which remained for further full-text assessment. The full-text eligibility assessment was independently carried out by the five authors, *AZW, PDN, MT, JB, and JAL*. To mitigate bias, the five authors provided reasons for inclusions and exclusions at this stage and reached an agreement through discussions. Any disagreements among the authors were resolved through discussion. The full-text eligibility assessment results in 105 studies, which are included in the study. After performing an updated search, 8 more articles were included, resulting in a total of 113 articles, as shown in Figure 1 and Table 1, that were critically analyzed in the study.

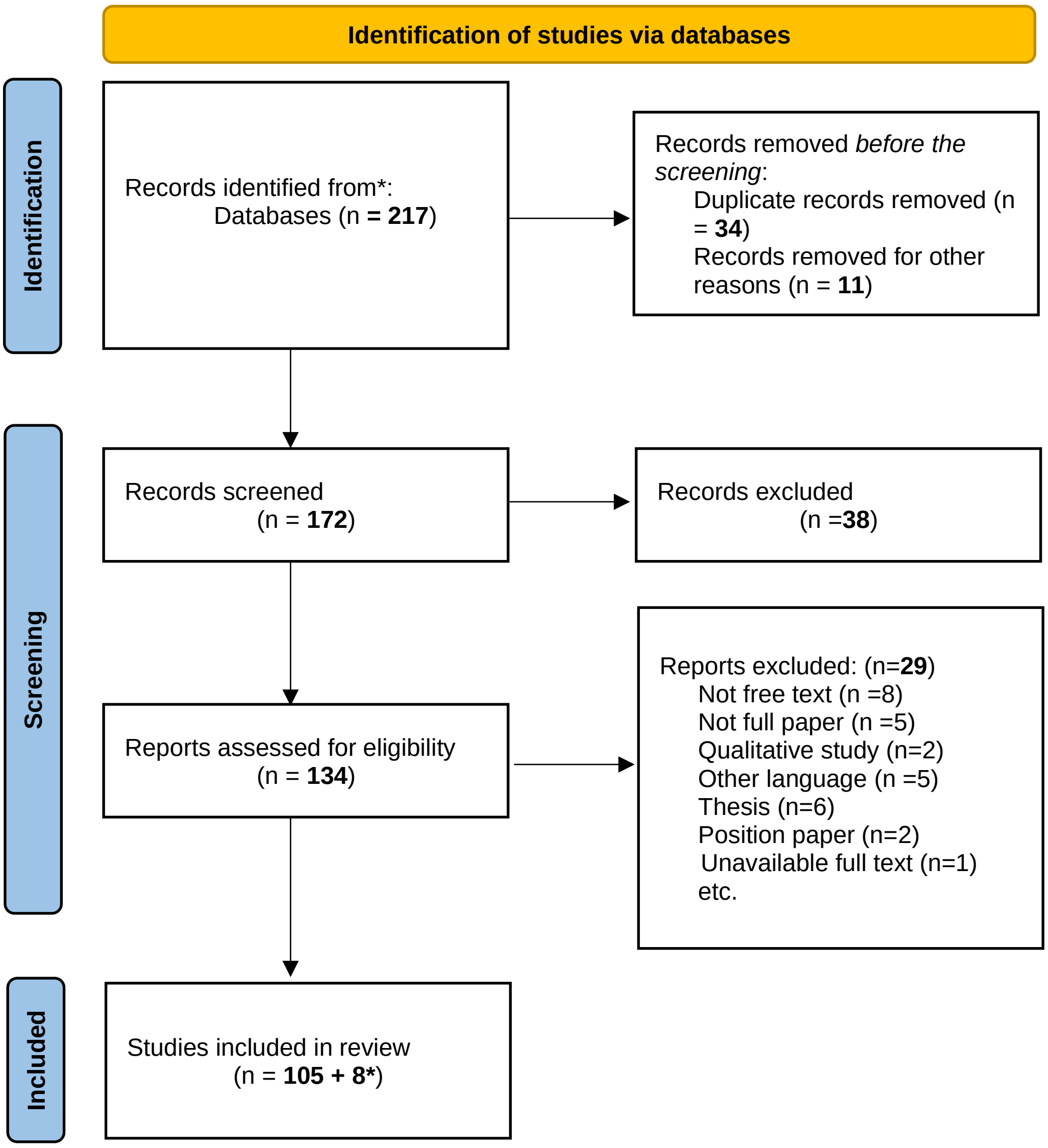

*Figure 1: The review process - identification, screening, and eligibility. * Denotes articles included after the update search.*

*Table 1: Language coverage, and category of proposed model for the included articles. "NA" depicts resource (dataset) construction. (see Appendix A for more information).*

| Refs. | Language | Main Model/Method |
|---|---|---|
| (20-25) | Norwegian | Rule-Based Systems |
| (26-30) | Norwegian | Interpretable Methods |
| (31-37) | Norwegian | Traditional/Statistical Learning Methods |
| (38) | Norwegian | Deep/Nonlinear Learning Methods |
| (39) | Norwegian | Hybrid Methods |
| (40) | Norwegian | Transformers and Large Language Models |
| (41) | Norwegian, Danish, Swedish | Transformers and Large Language Models |
| (42-51) | Swedish | Rule-Based Systems |
| (52) | Swedish, Finnish | Rule-Based Systems |
| (53) | English, French, German, Swedish | Rule-Based Systems |
| (54) | Swedish | Interpretable Methods |
| (55-82) | Swedish | Traditional/Statistical Learning Methods |
| (83) | Swedish, Finnish | Traditional/Statistical Learning Methods |
| (84) | English, Swedish | Traditional/Statistical Learning Methods |
| (85-90) | Swedish | Ensemble Methods |
| (91-98) | Swedish | Deep/Nonlinear Learning Methods |
| (99, 100) | Spanish, Swedish | Deep/Nonlinear Learning Methods |
| (101, 102) | Swedish | Hybrid Methods |
| (103-118) | Swedish | Transformers and Large Language Models |
| (119) | Spanish, Swedish | Transformers and Large Language Models |
| (120) | Spanish, Swedish, English | Transformers and Large Language Models |
| (121) | Swedish | NA |
| (122, 123) | Danish | Rule-Based Systems |
| (124) | Danish | Traditional/Statistical Learning Methods |
| (125-127) | Danish | Deep/Nonlinear Learning Methods |
| (128-132) | Danish | Transformers, and Large Language Models |

Task-specific categorization of the included articles (13):

- *Information extraction* (medical concepts, findings/symptoms, drugs/adverse events, specific characteristics, named entity recognition, and relations): encompasses a set of techniques that are designed to identify, categorize, and extract key pieces of information from unstructured clinical text data sources (20-23, 26-29, 33-36, 39, 47, 51, 54, 55, 62, 63, 68-70, 72-74, 76, 85-87, 89, 91, 92, 96, 97, 99, 100, 102, 108, 109, 115, 116, 123, 124, 127, 129, 130).
- *Classification* (risk stratification of medical conditions, coding, and cohort stratification): encompasses a set of techniques that are designed to categorize various types of health-related data (unstructured and structured) into predefined classes or groups (25, 26, 28-32, 35, 36, 38, 39, 54, 65, 66, 76, 79, 87, 88, 93, 104-106, 110-120, 124, 126, 127, 130-132).
- *De-identification*: refers to a set of methods employed to remove personal identifiers from a clinical text data (37, 40-42, 56, 57, 62, 64, 67, 78, 90, 94, 95, 98, 101, 103, 107, 110-112, 128).
- *Context Analysis* (negation detection, uncertainty/assertion, temporality, and abbreviation): encompasses a set of methods that are designed to extract and interpret the nuanced meanings or context of text data beyond the literal content (44, 46, 48-53, 60, 61, 68, 69, 71, 75, 77, 79, 82, 84, 110, 112, 118, 122).
- *Core NLP* (part-of-speech tagging, parsing, and segmentation): refers to a set of foundational computational techniques designed to process and analyze natural language, thereby helping to understand the structure and meaning, and they serve as the building blocks for more complex NLP tasks and applications (45, 59, 80).
- *Resource development* (content analysis, lexicons, corpora, and annotation, machine-readable dictionaries, models, and methods): refers to a set of techniques that focus on developing and compiling accessible resources that can serve as a building block for other complex and advanced NLP tasks and applications (22, 23, 27, 33, 34, 37, 40, 43-45, 47, 48, 50, 52, 53, 58, 60, 68-72, 75-77, 81-85, 92, 121, 123, 125, 128-130, 132).

## 3.2. Comprehensive Analysis and Evaluation of the literature

As described earlier, the analysis and evaluation of the included articles were performed based on the above-predefined categories, which were further grouped for the analysis including *publication insight and trends*, providing an overview related to the publications dynamics, *data characteristics* providing an overview related to the dataset used in model development and evaluations, *clinical characteristics* providing overview related to the clinical aspect of the dataset and study objective along with extent of expert involvement in data preparation, and *NLP techniques* providing overview related to the class of NLP models and task category studied.

### 3.2.1. Publication Insight and Trends

This section provides insights into the general publication trends among the three languages depicting the total volume of publications per year, language co-occurrence depicting the extent these languages appear together in a publication, and collaborative network depicting the overall inter and intra-collaborations of researchers from each respective country.

#### *Publication Trend*

The overall publication trend for the three languages is shown in Figure 2 below. As can be seen from the figure, the number of publications focusing on Swedish clinical text shows a slight fluctuation over the years but maintains a generally positive trend over the years, while publications focusing on Norwegian clinical text have seen similar trends but to a lesser extent. However, publications focusing on Danish clinical text have seen the highest fluctuations over the years, despite starting publishing early in 2011, and have no publication between 2015 and 2020. The three languages began appearing in publications at different timeframes: Swedish was the first, emerging in 2010, followed by Danish in 2011, and Norwegian, which appeared last, in 2014. In general, Swedish accounts for the majority of the publications every year, and this trend seems to become a fair share among the three languages starting from 2023, indicating the overall increased research activity in clinical NLP among the Scandinavian languages. The pie chart depicts the distributions of publications across the languages over the years, revealing the enormous clinical NLP research activity in the Swedish language, accounting for 69.2%, followed by Norwegian at 20.6%, and Danish at 10.3%. The success of Swedish in this domain could be attributed to several contributing factors including early initiatives, adoption and commitment, the presence of a strong research infrastructure with access to data and computing power, good governmental intiatives and institutional support including funding, a good interdisciplinary collaboration among universities, research institutions, healthcare providers, and international partners. On the other hand, certain contributing factors such as population size can be determinantal, with Sweden having a population twice as large as Norway, and Denmark.

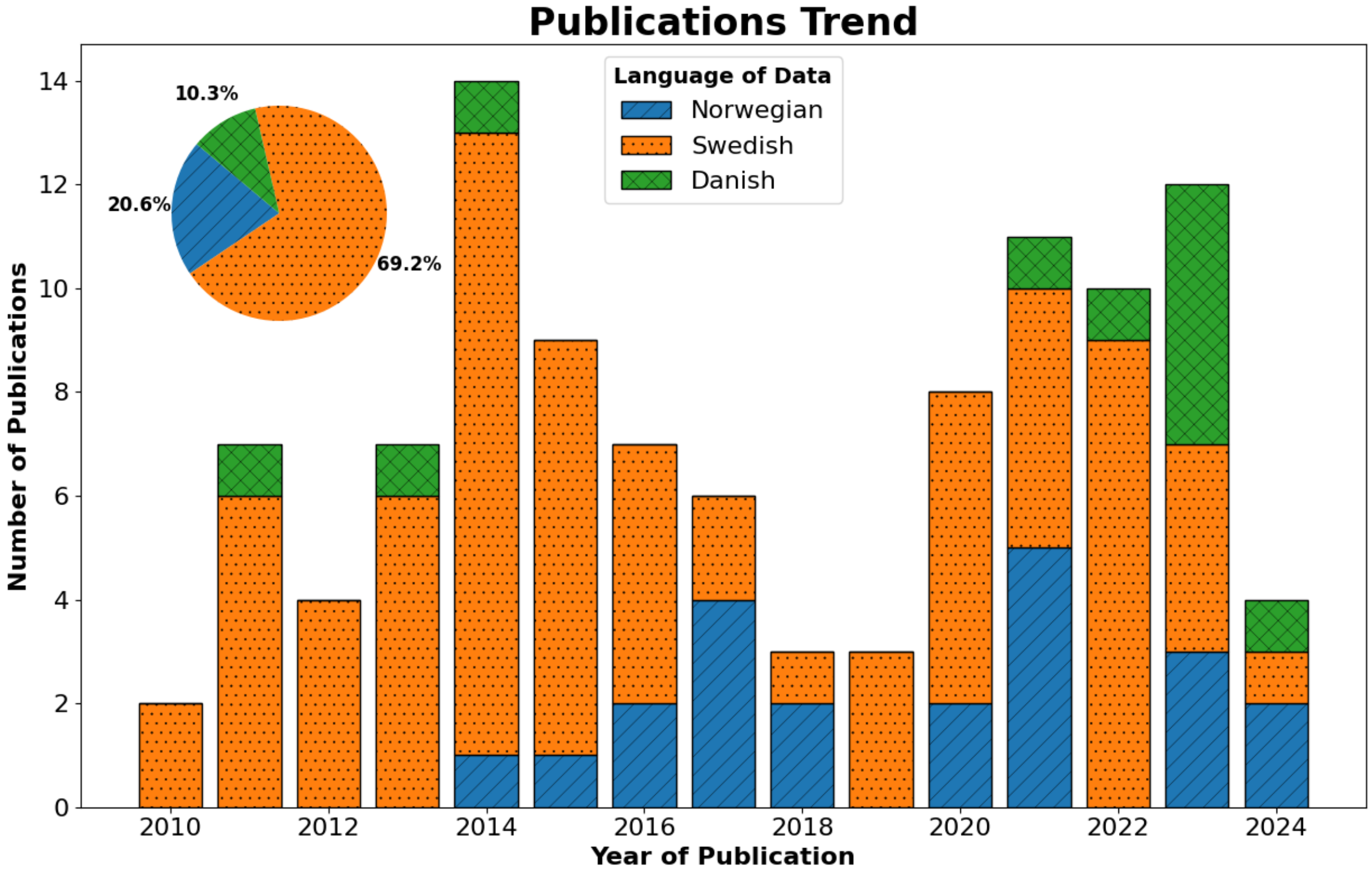

*Figure 2: Publications trend among the three countries: Norwegian, Swedish, and Danish.*

#### *Language Co-occurrence*

In a natural language processing setting, language co-occurrence, i.e. analysis of two or more languages, in a publication can signify several important aspects of research and development. This could involve analysis related to cross-cultural research to understand cultural and linguistic nuances for adaptation and transfer learning, and diverse data insights for uncovering language-specific patterns, characteristics, and phenomena along with shared similarities (83). One such aspect is the prospect of enhanced transfer learning and adaptation, where developed tools and knowledge from one language can be easily applied or transferred to another. It could also pave the way for addressing data limitations by leveraging data augmentation techniques with closely related language. Furthermore, this could facilitate addressing resource disparities through leveraging resources from well-supported languages to less-supported languages. In this regard, the result demonstrated that out of 113 articles, only 8% (n=9) articles considered multiple languages.

The language co-occurrence diagram, shown below in Figure 3, provides insight into how frequently pairs of languages appear together in a publication over the years, emphasizing only the three languages: Norwegian, Swedish, and Danish. Language co-occurrence counts the presence of a pair of languages appearing in publication over the years. For example, given a set of languages $\mathcal{S}$ with predefined criteria group $\mathbb{G}_i$ (Swedish, Norwegian, Danish), generate unique two-element permutations $(s_1, s_2)$ such that $s_1, s_2 \in \mathcal{S}$ and at least one $s_i$ belong $\mathbb{G}_i$. Each newly generated permutation element $p_i$ is added to the pairs list $\mathcal{U}$ if $p \notin \mathcal{U}$. As can be seen from Figure 3, Swedish has shown extensive co-occurrence with other languages and has been studied along with Norwegian (n=1), Danish (n=1), English (n=3), German (n=1), Spanish (n=4), Finnish (n=2) and French (n=1). Norwegian and Danish, on the other hand, have the least frequent co-occurrence only appearing together once and each having one occurrence with Swedish. The most intriguing observation is related to the least frequent co-occurrence of the three languages – Swedish, Norwegian, and Danish. There is little to no effort in studying these languages together, even though these languages share linguistic similarities with countries that have similar healthcare systems, cultural contexts, social values, healthcare challenges, and possibly a culture of cross-border cooperation. Those single pairs of occurrences, as shown in Figure 3, involving French, German, Norwegian, Danish, and Swedish could suggest initial explorations encompassing either a pilot study, proof of concept study, or the beginning stages of expanding research to include more diverse linguistic data.

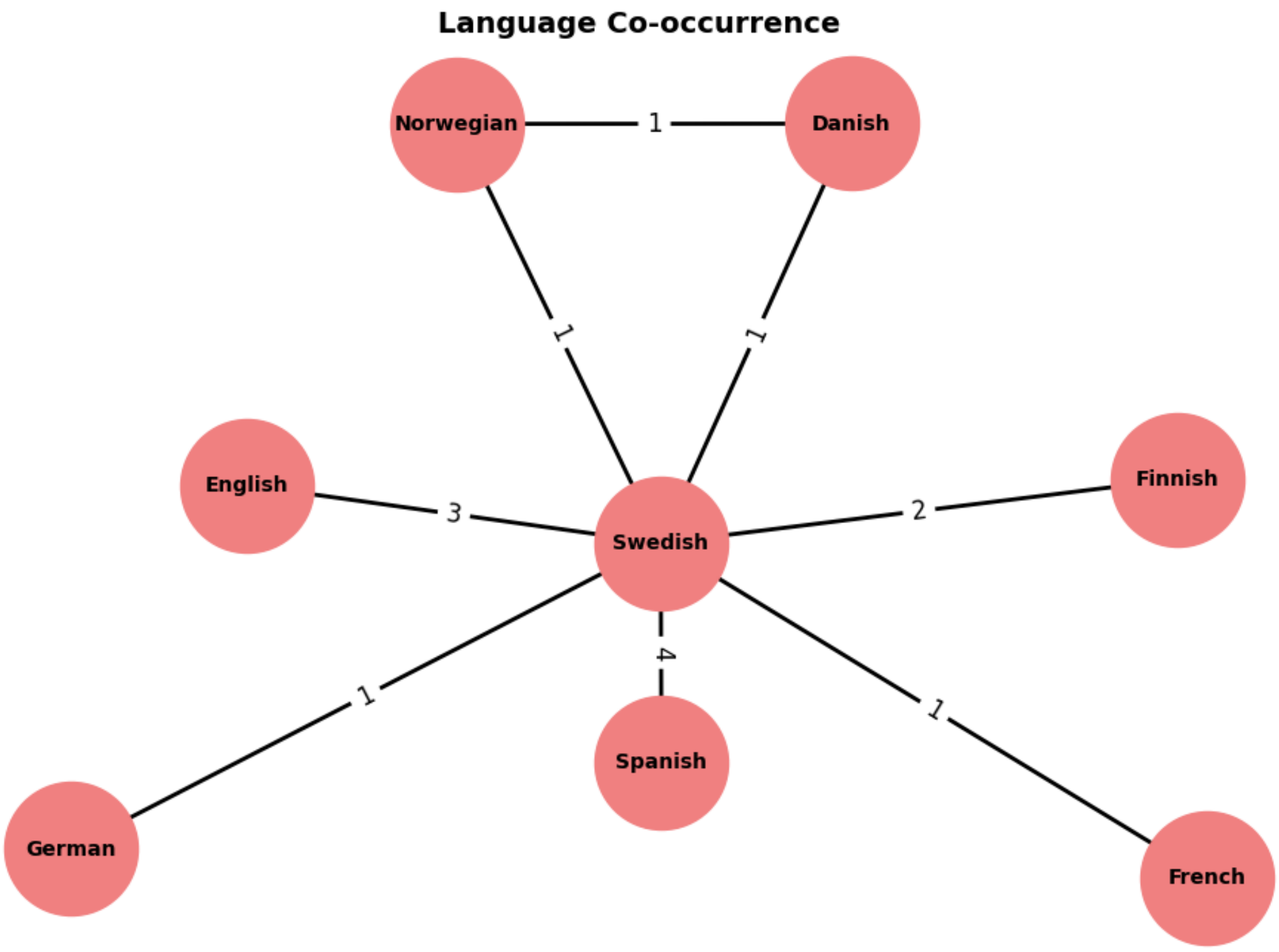

*Figure 3: Language co-occurrence in a publication over the years.*

## *Collaboration Network*

The collaborative network depicts the frequency of authors from a certain country appearing with either authors of the same country, depicted in a self-loop, or authors of another country depicted in the edge connecting the countries. The network counts a pair of authors from either the same country or two different countries appearing in publications over the years. In our context, the author's country refers to the individual's affiliation. Self-loop depicts only those articles, whose authors' affiliations are in the same country, in other words, articles that fully depict intra-collaborations within a country. Edge represents inter-collaborations between countries, and considers only distinct or unique countries from an article. For example, if there are four authors in one article, where 2 from Sweden, 1 from Norway, and 1 from the USA, then only the 3 authors are considered in the analysis discarding the duplicate and 1 author from Sweden in this case, and only the connection to the 3 countries – Sweden, Norway, and Denmark are displayed on the network. The existence of active and frequent collaborations among the different research groups and institutions, both within and across borders, could be a key driver for the success of a project and is crucial in a field like NLP, where interdisciplinary and international cooperation can lead to significant advancements by fostering knowledge exchange, and innovations.

Out of the 113 reviewed articles, 22% (n =25) articles include authors from different countries depicting inter-collaborations, and where 78% (n=88) articles depict intra-collaboration within the same country. Similarly, as can be seen in Figure 4, there is a strong intra-national collaboration in all three countries, where Sweden (n=67) represents the highest, followed by Norway (n=11), and Denmark (n=10). Sweden appears as the most frequent collaborator internationally, with ties to Finland (n=2), Norway (n=12), Australia (n=1), Spain (n=4), the USA (n=3), Italy (n=1), Denmark(n=1), and France (n=1). This centrality implies that Sweden could be a leading force in research, possibly providing pivotal datasets, resources, or expertise in the field. Despite the small number of publications compared to Sweden, Norway also appears to gain frequent collaborators across the globe with ties to Finland (n=2), Sweden (n=12), Australia (n=1), Spain (n=6), the USA (n=3), Italy (n=1). On the other hand, Denmark appears to have limited collaborators with ties only to Sweden (n=1), and Spain (n=1), and this could perhaps imply either the presence of a strong independent national research community in clinical NLP or a lack of resources and funding to create the necessary collaboration across the globe. Another possible explanation is that Denmark potentially has a journal, which is less based on free text.

There appears a robust partnership between Norwegian and Swedish researchers (n=12), and given their geographic proximity and linguistic similarities, it is expected that these two nations to have strong research ties. However, it is surprising that Denmark has very few ties with Norwegian, and Swedish researchers, only having one connection with Sweden (n=1), even though these countries share strong linguistic similarities. This kind of connection, with only one connection (n=1), might depict one-time or project-specific partnerships rather than ongoing research relationships. Overall, cross-Nordic interactions depict a productive research network involving Sweden, Norway, and Finland but Denmark. Generally, Sweden displayed a significant international collaboration and high-frequency intra-national collaboration compared to the other languages. Sweden's high level of collaboration could suggest the existence of one or more research and innovation hubs spearheading multiple projects. In this regard, the "*Swedish Health Record Research Bank*", an infrastructure incorporating a large EHR dataset from Karolinska University Hospital, is among the facilitating factors. Similarly, the results suggest that, in addition to Sweden, Norway is maybe developing its research and innovation hubs.

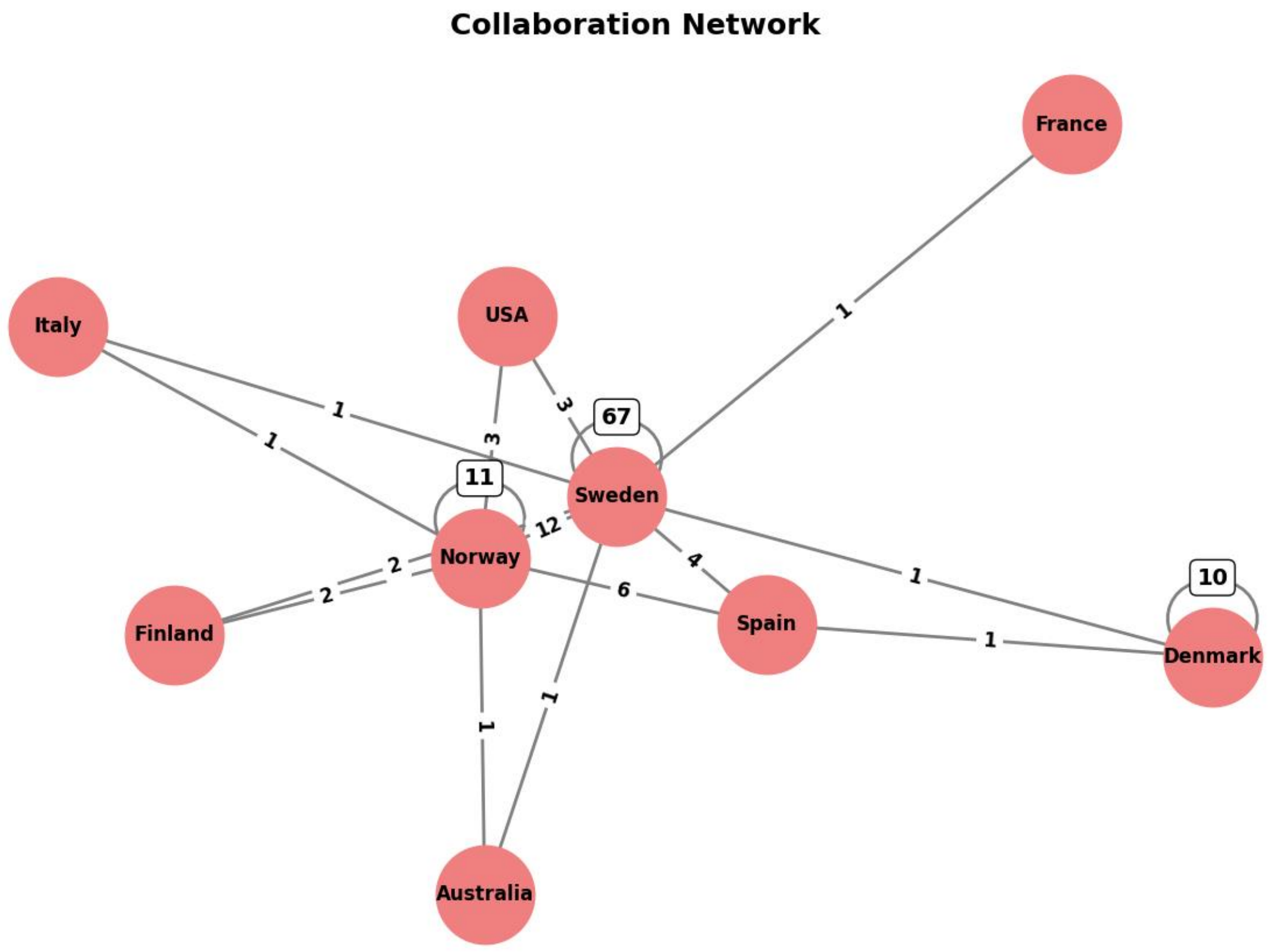

*Figure 4: Collaborations Network. Self-loop indicates the number of publications appearing only among authors in that country.*

### 3.2.2. Data Characteristics

This section presents an overview of the data sources and names, type of EHR documents, language, data availability, data quality, and related issues, and further provides insight into the scope of data types, the linguistic variety, accessibility for research purposes, and potential challenges related to data integrity and reliability.

#### *Data sources*

Health institutions' participation in providing datasets for research and development is vital to developing clinically relevant NLP models that are well-adapted to real-world clinical and hospital settings or environments. In general, the reviewed studies have relied on four major types of data sources: *EHR data from hospitals*, *registry data that may provide information about specific patients or larger population insights, non-EHR data including medical journals, and synthetic data/artificially generated data*, which bypass *privacy concerns while still contributing valuable information for model development*. In those studies utilizing registry data, the review identified that registry information is mainly used to augment EHR data for classification and symptom extraction tasks. For example, Jensen et. al. used the Norwegian death registry to construct disease and event trajectories for cancer patients from symptoms to death (30).

The Karolinska University Hospital stands out with a substantial contribution to Swedish datasets (n =65), as shown in Figure 5, with very little contribution from other similar hospitals in Sweden. The success behind the huge contribution of the Karolinska University Hospital could be linked to the initiative to release massive EHR data research infrastructure known as "*The Swedish Health Record Research Bank*" (133). Among the hospitals in Denmark, Odense University Hospital (n = 6) provides the highest contributions followed by Sct. Hans Hospital (n=2). In Norway, the University Hospital of North Norway has the highest contribution (n=5), followed by Akershus University Hospital (n=3) and Sørlandet Hospital Trust (n=3). There appears to be a more equitable distribution of contributions among healthcare institutions in Norway compared to Sweden and Denmark. Utilizations of registry data and non-EHR data sources to augment EHR datasets are also observed in some of the studies, specifically involving Swedish (registry: n =4, non-EHR: n =7 ) and Norwegian languages (registry: n =4, non-EHR: n =2 ). However, the use of synthetic datasets is observed only in Norwegian (n=6).

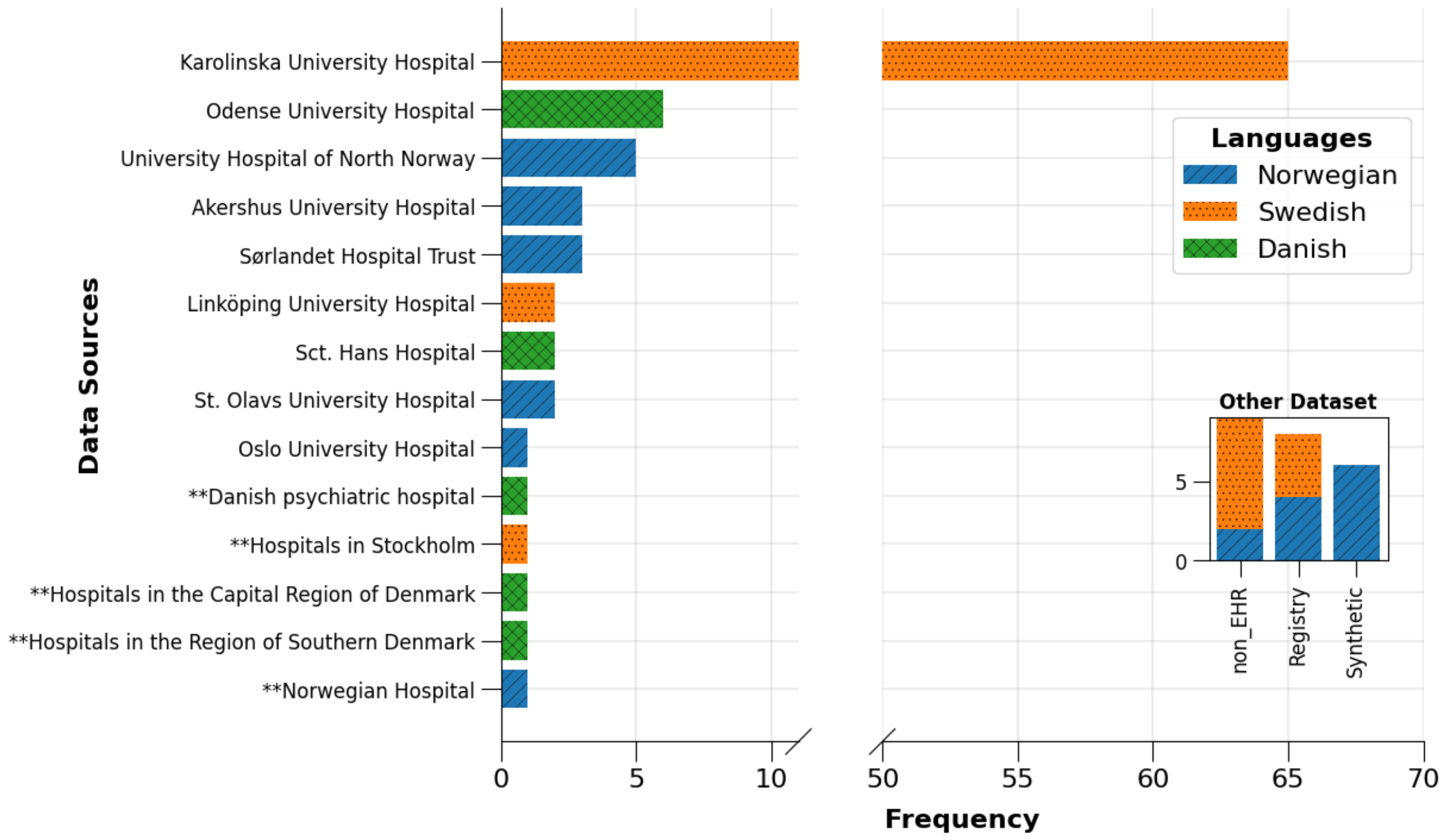

*Figure 5: Frequency of data sources by institutions used in the study. ** denotes mention of a data source without specific identifiers of the source.*

## *Data Quality and Other Related Issues*

High-quality data is of utmost importance and a foundational step for model development (134). Failure to address the existing data quality issues at the early stage of model development can be costly, and it could affect many aspects including accuracy and reliability, model training and validation, generalization ability, and finally model adoption and use (134, 135). Therefore, it is crucial to address, mitigate, and improve the quality of the data through preprocessing, understanding of the clinical context, and performing an ongoing validation. There are several types of *data quality and other related issues* reported in the reviewed studies, and for the sake of clarity, these reported issues were mapped into one or more of the following keywords: *misspelled, multi-word, abbreviation, reference resolution, lexical variation, incomplete data, inconsistent data, noisy data, duplicate data, formatting issues, handwritten text, lack of public data, data sparsity, and Others* (3, 13). A detailed definition of these keywords can be found in Appendix A. It should be noted that the extracted number of issues from the studies could be low since authors often don't report all issues encountered in model development and should be interpreted accordingly.

As can be seen from Figure 6, the category *others* have the highest frequency (n = 29) suggesting the wide variety of data quality issues that are not even easily categorized into the above-predefined keywords and highlighting the complexity and uniqueness of challenges in clinical text data. It should be noted that although the *others* category is the highest, it encompasses more than eighteen individual issues, and if these issues were compared as individual categories, the frequency would be much smaller.

The next top frequent issue is *inconsistent data* (n = 22), followed by *incomplete data* (n = 11). The fourth and fifth most frequent issues are *lexical variation* (n = 8) and *formatting issues* (n = 7), which reflect the diverse ways in which medical professionals document patient information. The presence of *misspelled* words and *noisy data* along with the *lack of public data* ranked as the six most frequent issues with all appearing in six occurrences emphasizing the need for robust preprocessing methods and privacy concerns. The seventh most frequent issues are *abbreviation* and *data sparsity*, each appearing in five occurrences, followed by *reference resolution* and *handwritten* text each appearing in three occurrences. The less frequent issues include *multi-word* expressions (n = 2) and *duplicate data* (n = 1).

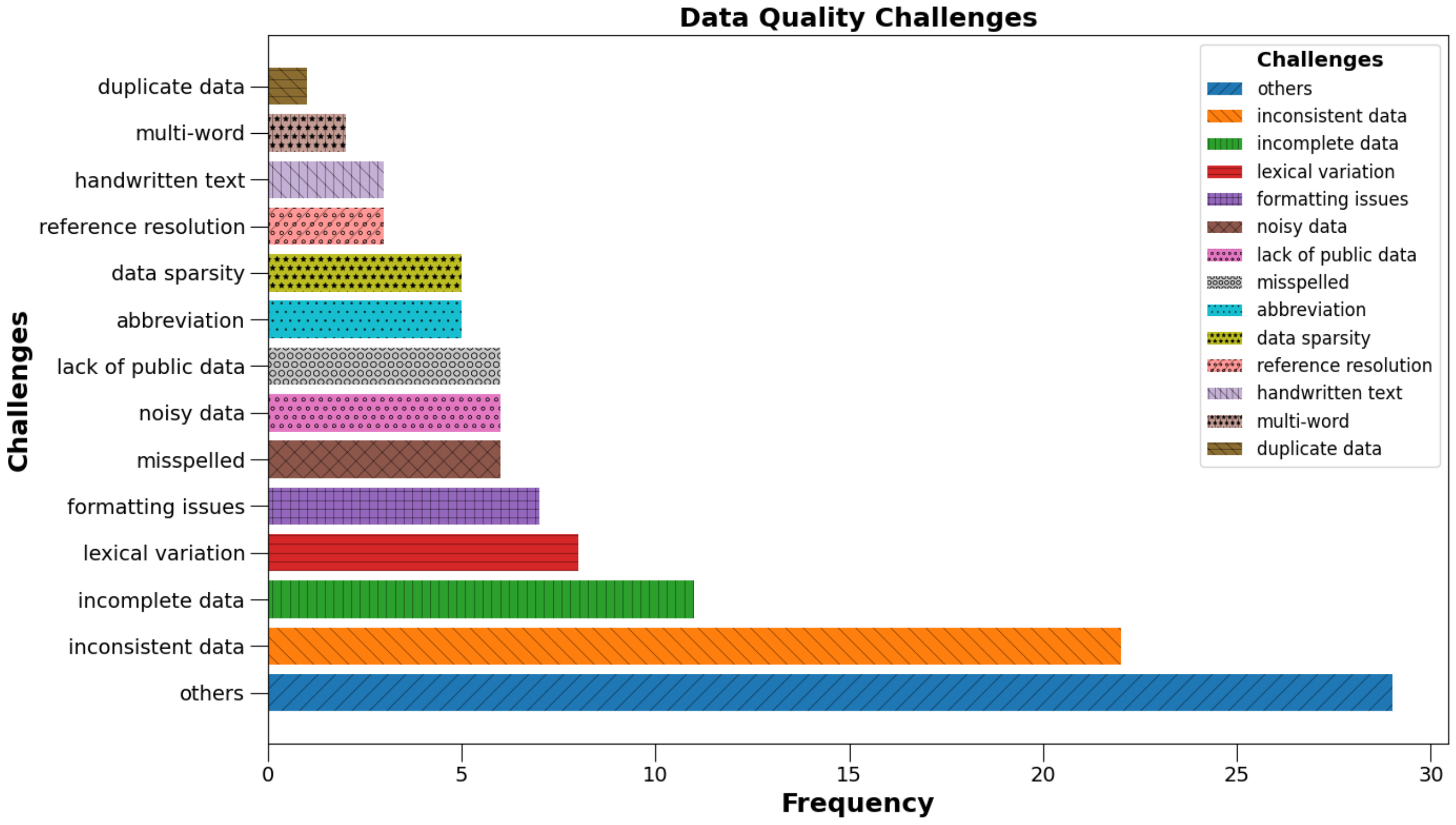

*Figure 6: Data quality and other related issues reported in the studies.*

### *Type of EHR Documents*

EHR encompasses a wide range of document types that record comprehensive information on various aspects of patient care. The reviewed studies have utilized various kinds of EHR documents for NLP model development, as shown in Table 2. In Table 2, studies that didn't specifically mention the type of documents are categorized under "Diverse/Not specific" categories, and studies that barely refer to "clinical notes" without a detailed description of the type of document are categorized as "clinical notes". Accordingly, clinical notes are the most utilized document type overall (19.7%) and across each language group, with the highest percentage in Swedish (62.5%) and the lowest in Norwegian (17.5%). The progress note is the second most utilized document overall (17.73%), and evenly distributed but with a slight lean towards Norwegian (38.88%) and Swedish (41.67%). The third most utilized document overall is the medical history and physical examination report (15.76%), with a strong presence in Swedish (56.25%) followed by Norwegian (37.5%). The other relatively well-utilized documents overall include discharge summaries (10.84%), and laboratory and diagnostic reports (10.34%), where Swedish has the highest usage of discharge summaries (59.09%), and Norwegian has the highest usage of laboratory and diagnostic reports (57.14%). Some document types have notably low utilization across the data set, such as surgical reports (4.43%), emergency department reports (2.46%), allergy and adverse event reports (1.97%), medication and prescription information (1.48%), and nutritional assessments and plans (0.99%).

*Table 2: Electronic Health Record (EHR) documents utilized in the studies. "Diverse/Not specific" category implies studies that didn't specifically mention the type of EHR document analyzed.*

| Type of Documents | Overall | Norwegian | Swedish | Danish |
|---|---|---|---|---|
| | Utilization (Percentage and Count) | | | |
| Allergy and Adverse Event Reports | 1.97% (4) | 75% (3) | 25% (1) | - |
| Clinical Notes | 19.7% (40) | 17.5% (7) | 62.5% (25) | 20% (8) |
| Consultation and Referral Reports | 4.9% (10) | 90% (9) | - | 10% (1) |
| Discharge Summary | 10.8% (22) | 31.8% (7) | 59.1% (13) | 9.1% (2) |
| Diverse/Not specific | 9.4% (19) | - | 100% (19) | - |
| Emergency Department Reports | 2.5% (5) | - | 75% (4) | 25% (1) |
| Laboratory and Diagnostic Reports | 10.3% (21) | 57.1% (12) | 38.1% (8) | 4.7% (1) |
| Medical History and Physical Examination Report | 15.8% (32) | 37.5% (12) | 56.3% (18) | 6.3% (2) |
| Medication and Prescription Information | 1.5% (3) | 66.7% (2) | 33.3% (1) | - |
| Nutritional Assessments and Plans | 1% (2) | - | 100% (2) | - |
| Progress Notes | 17.7% (36) | 38.9% (14) | 41.7% (15) | 19.4% (7) |
| Surgical Reports | 4.4% (9) | 55.5% (5) | 11.1% (1) | 33.3% (3) |

#### *Access to Clinical Text Data*

Access to clinical data is of utmost importance and cornerstone for model development. However, data sharing remains a significant challenge in the clinical domain for various reasons including but not limited to privacy concerns, ethical considerations, and the logistical challenges of anonymizing and sharing large datasets. The review has identified studies that mentioned access to data either as readily available or that can be accessed upon request by the authors. Overall, only about just under the third of the studies (31.86%, n = 36/113) reported that their clinical text data are available to external researchers. On the other hand, Sweden (34.57%, n = 28/81) and Norwegian (31.82%, n = 7/22) show the highest availability rate with nearly over a third of the studies indicating data availability, and only one study reporting data availability in Denmark (8.33%, n = 1/11).

### 3.2.3. Natural Language Processing Techniques

This section presents an overview and insight into the class of NLP model/methods employed in the studies, the diversity of NLP tasks and sub-tasks considered, feature representation techniques applied, availability of code and pre-trained models along with adaptation and transfer learning.

#### *Natural Language Processing Methods*

For the sake of brevity and interpretability, NLP methods employed in the studies are roughly grouped into *rule-based, interpretable methods, traditional/statistical learning methods, ensemble methods, deep/nonlinear learning methods, hybrid methods (combining deep learning and other models), and transformers and large language models* based on method complexity, see Appendix A. The evolution of these methods over the years is shown in Figure 7. Generally, traditional/statistical learning methods are the most frequently used method accounting for 31.6% (n = 78) followed by rule-based systems (23.9%, n = 59). The third most frequent method is transformers and large language models (14.2%, n = 35) followed by interpretable methods (12.6%, n = 31), and deep/nonlinear learning methods (11.3%, n = 28). Ensemble methods and hybrid methods are the least frequent accounting for 4.9% (n = 12) and 1.6% (n = 4) respectively. A more country-specific analysis can be found in Appendix A.

**Initial period (2010-2014)**: During this period as can be seen from Figure 7, there's a clear dominance of rule-based systems (41.97%, n = 34) and traditional/statistical learning methods (53.1%, n = 43) reflecting the initial stages of NLP development, where simpler models that relied on handcrafted rules and statistical features were prevalent. There are some activities of using ensemble methods (2.5%, n =2), and interpretable methods (2.5%, n = 2) that started to appear in 2014.

**Mid period (2015-2019)**: The adoption of ensemble methods (12.8%, n = 7) and deep/nonlinear learning methods (16.3%, n = 9) starts to be noticeable from 2015 onwards, marking the beginning of the transition towards more complex models that can capture the nuances of language better than traditional methods. This period marks the initial exploration of deep learning techniques. Further, interpretable methods (20.0%, n = 11) continue to grow, albeit slowly, indicating a consistent interest in maintaining some level of interpretability in clinical NLP solutions. Generally, despite the introduction of deep/nonlinear methods, traditional/statistical learning methods (30.0%, n= 17) remain dominant, and rule-based systems (20.0%, n = 11) are still noticeable but to a lesser extent compared to the initial period.

**Recent period (2020-2024):** A significant shift towards transformers and large language models (31.53%, n = 35), and continued interest in deep/nonlinear learning methods (17.12%, n = 19) from 2020 onwards. The substantial jump in these methods by 2022 and onwards suggests a rapid adoption of state-of-the-art NLP technologies, likely driven by their success in general language understanding tasks. The emergence of hybrid methods (3.6%, n = 4) in 2023 depicts the interest in combining various methodologies to leverage the strengths of each for tackling complex clinical text analysis tasks. Interpretable methods (16.22%, n = 18) maintain a consistent appearance across the years. Traditional/statistical learning methods (16.22%, n = 18), and ensemble methods (2.7%, n = 3) showed a significant decline compared to the previous period, however, rule-based systems (12.6%, n = 14) showed a consistent decrease.

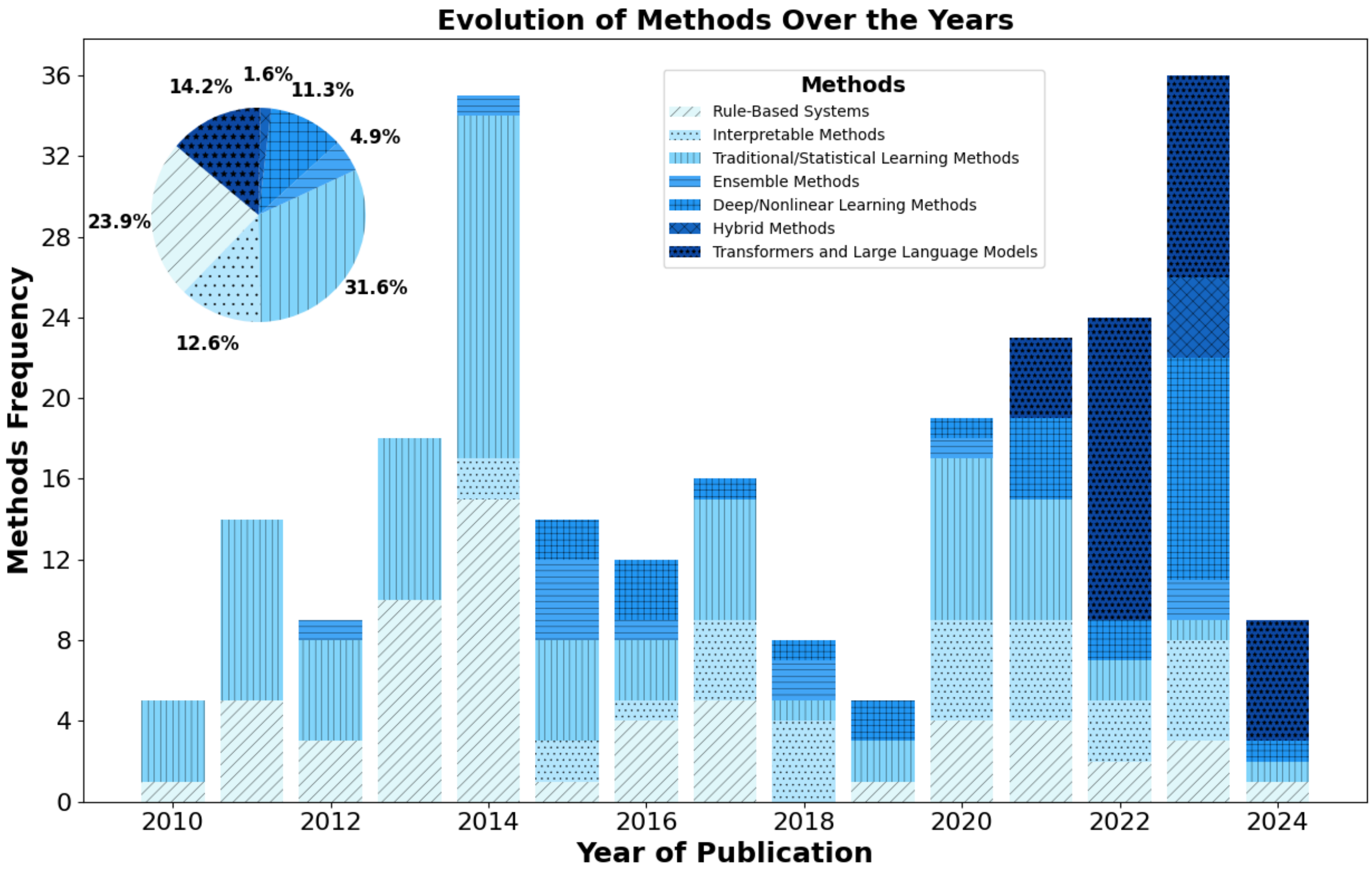

*Figure 7: Evolution of NLP methods for Scandinavian over the years. The colour gradients depict model complexity.*

## *Natural Language Processing Tasks Trend*

This section provides an overview and insights into the various NLP tasks and strategies employed in the studies. For the sake of brevity and interpretability, the NLP strategies employed in the studies are categorized into *core NLP, resource development, de-identification, classification, context analysis, and information extraction* (13, 38, 136, 137). It is worth noting that a single study can represent three or more tasks, and the number of NLP tasks could exceed the number of reviewed articles. A more detailed description of these specific tasks and their sub-categories along with detailed results corresponding to country-specific development can be found in Appendix A.

*Information extraction* (33.8%, n = 67) represents a significant portion of the studied strategy across all the NLP tasks, and named entity recognition (44.3%, n = 43), a fundamental task for identifying clinical entities in text, is the most focused strategy within this category followed by findings/symptoms (16.5%, n = 16) and drugs/adverse events (15.5%, n = 15). Medical concepts (11.3%, n = 11) and relations (9.28%, n = 9) are also fairly represented, where studies focusing on specific characteristics (3.1%, n = 3) are less.

*Classification* (24.2%, n = 48) tasks take up a notable share of the studied strategy, with a significant emphasis on risk stratification of medical conditions from EHR text (58.4%, n = 28) followed by coding (39.6%, n = 19), and studies focusing on cohort stratification (2.1%, n = 1) are rare.

*Resource development* (18.7%, n = 37) represents a fair portion of the studied strategy, emphasizing the importance of creating foundational resources like lexicons, corpora, and machine-readable dictionaries. The high frequency of corpora and annotation (45.2%, n = 19) efforts suggests a strong emphasis on developing annotated datasets, which are critical for training and evaluating NLP models. Efforts towards developing new methods (30.9%, n = 13) are also significant followed by lexicons (19%, n = 8), while efforts to release pre-trained models (2.4%, n = 1), and create machine-readable dictionaries (2.4%, n = 1) are very seldom.

*Context Analysis* (11.1%, n = 22) focuses on understanding the context around clinical mentions in the text. Negation detection (46.4%, n = 13) and uncertainty/assertion (28.5%, n = 8) are the most represented strategies, crucial for accurately interpreting the clinical significance of statements in records. Other strategies like temporality (10.7%, n = 3) and abbreviation (14.2%, n = 4) are also present.

*De-identification* (10.6%, n = 21) represents a tenth of the studied strategy, highlighting the significant challenge of privacy concerns in working with clinical texts and the need for methods that can reliably anonymize patient information effectively on both small and large-scale datasets.

*Core NLP* tasks are relatively underrepresented, with only a small fraction (1.5%, n = 3) of the efforts dedicated to fundamental NLP capabilities like part of speech tagging (40%, n = 2) and parsing (60%, n = 3).

### *Analysis of Model usage in various Natural Language Processing Tasks*

This section presents an insight into the adoption rate of different types of models/methods in realizing specific clinical NLP tasks, as shown in Table 3. As can be seen from the table, research focusing on classification tasks places high emphasis on interpretability as demonstrated by the dominance of interpretable methods. The dominance of a rule-based system and traditional/Statistical learning methods in context analysis, information extraction, and core NLP tasks demonstrated the research efforts in incorporating domain knowledge in feature engineering. There is also a higher adoption rate of deep/nonlinear learning methods, and transformers and large language models in classification and information extraction tasks.

*Table 3: Utilization of different NLP models/methods across various NLP tasks.*

| Models/Methods | Classification | Context Analysis | Core NLP | De-identification | Information Extraction |
|---|---|---|---|---|---|
| Deep/Nonlinear Learning Methods | 44.2% (19) | - | - | 7.0% (3) | 48.8% (21) |
| Ensemble Methods | 25.0% (4) | - | - | 6.25% (1) | 68.8% (11) |
| Hybrid Methods | 40.0% (2) | - | - | 20.0% (1) | 40.0% (2) |
| Interpretable Methods | 57.5% (27) | 2.2% (1) | - | - | 40.4% (19) |
| Rule-Based Systems | 15.3% (11) | 31.9% (23) | 8.3% (6) | 11.1% (8) | 33.3% (24) |
| Traditional/Statistical Learning Methods | 17.6% (18) | 18.6% (19) | 2.0% (2) | 14.7% (15) | 47.1% (48) |
| Transformers and Large Language Models | 39.7% (23) | 5.2% (3) | - | 19.0% (11) | 36.2% (21) |

### Classification Tasks

Classification methods encompass the set of approaches that are employed to categorize various clinical text data into predefined classes or groups (13, 38). A classification method encompasses several key tasks, including *risk stratification of medical conditions from EHR text* (70.4%, n = 76), *coding* (27.8%, n = 30), and *cohort stratification* (1.9%, n = 2), each addressing different aspects of clinical data categorization and analysis. The overall adoption rate of the NLP model/methods corresponding to these classification tasks is depicted in Table 4. The high adoption rate of interpretable methods in classification tasks depicts the research interest in generating an explainable or interpretable model, which provides insight into how the model reaches at such a decision for classification. In *coding*, transformers and large language models (36.7%, n = 11) have the highest adoption rate, and traditional/statistical learning methods (26.7%, n = 8) and interpretable methods (20%, n = 6) have a significant presence. Little efforts are focused on the C*ohort stratification task*, and the existing efforts rely exclusively on traditional/statistical learning methods (100%, n = 2). In *risk stratification of medical conditions from EHR text* interpretable methods see the most significant adoption (27.6%, n = 21), followed closely by deep/nonlinear learning methods (21.1%, n = 16). There is also a notable usage of traditional/statistical learning methods (15.8%, n = 12), and transformers and large language models (15.8%, n = 12).

*Table 4: Adoption rate of different NLP methods for classification tasks.*

| Model/Method | Classification Task | | |
|---|---|---|---|
| | Coding | Cohort stratification | Risk stratification of medical conditions from EHR text |
| **Deep/Nonlinear Learning Methods** | 3.3% (1) | - | 21.1% (16) |
| **Ensemble Methods** | - | - | 5.3% (4) |
| **Hybrid Methods** | - | - | 5.3% (4) |
| **Interpretable Methods** | 20% (6) | - | 27.6% (21) |
| **Rule-Based Systems** | 13.2% (4) | - | 9.2% (7) |
| **Traditional/Statistical Learning Methods** | 26.7% (8) | 100% (2) | 15.8% (12) |
| **Transformers and Large Language Models** | 36.7% (11) | - | 15.8% (12) |

## Information Extraction Task

Information extraction methods encompass the set of approaches that are designed to identify, categorize, and extract key pieces of information from unstructured clinical text data sources (13, 136). Information extraction methods comprise several key strategies and focus areas including *medical concepts* (15.5%, n =33), *findings/symptoms* (18.6%, n = 40), *drugs/adverse events* (12.1%, n = 26), *specific characteristics* (4.2%, n = 9), *named entity recognition* (41.9%, n = 90), *relations* (7.9%, n = 17). The overall adoption rate of the NLP model/methods corresponding to these information extraction tasks is depicted in Table 5. Generally, the high adoption rate of rule-based systems and traditional/Statistical learning methods in this specific task could be due to the research interest in incorporating domain knowledge in feature engineering. Extraction of *medical concepts* predominantly utilized traditional/statistical learning methods (27.3%, n = 9), along with interpretable methods (21.2%, n = 7), deep/nonlinear learning methods (18.2%, n = 6), and rule-based systems (18.2%, n = 6), to varying extents. Moreover, there are some utilizations of transformers and large language models (6%, n = 2), and hybrid models (6%, n = 2). *Findings/Symptoms* predominantly utilized rule-based systems (25%, n = 10), with deep/nonlinear learning methods (22.5%, n = 9) and interpretable methods (22.5%, n = 9) closely following. There are also some utilizations of traditional/statistical learning methods (15%, n = 6), and transformers and large language models (7.5%, n = 3). *Drugs/Adverse Events* predominantly utilized ensemble methods (30.8%, n = 8), with transformers and large language models (19.2%, n = 5), and traditional/statistical learning methods (19.2%, n = 5) also taking a significant share. Extraction of *specific characteristics* predominately utilized interpretable methods (55.6%, n = 5), with rule-based systems (33.3%, n = 3) and traditional/statistical learning methods (11.1%, n = 1) taking a noticeable share. *Named Entity Recognition (NER)* predominantly utilized traditional/statistical learning methods (37.8%, n = 34), where transformers and large language models (16.7%, n = 15), and rule-based systems (18.9%, n = 17) also take a significant share. Moreover, there is a noticeable presence of deep/nonlinear learning methods (7.8%, n = 7), and interpretable methods (10%, n = 9). Extraction of *Relations* predominantly utilized traditional/statistical learning methods (47.1%, n = 8), whereas interpretable methods (23.5%, n = 4), and transformer and large language models (17.7%, n = 3) also have a very significant presence.

*Table 5: Adoption rate of different NLP methods for information extraction tasks.*

| Model/Method | Information Extraction Task | | | | | |
|---|---|---|---|---|---|---|
| | Medical Concepts | Findings/ Symptoms | Drugs/Adverse events | Specific characteristics | Named entity recognition | Relations |
| **Deep/Nonlinear Learning Methods** | 18.2% (6) | 22.5% (9) | 15.4% (4) | - | 7.8% (7) | - |
| **Ensemble Methods** | 3% (1) | 2.5% (1) | 30.8% (8) | - | 4.4% (4) | 5.88% (1) |
| **Hybrid Methods** | 6% (2) | 5% (2) | - | - | 4.4% (4) | 5.88% (1) |
| **Interpretable Methods** | 21.2% (7) | 22.5% (9) | 11.5% (3) | 55.6% (5) | 10% (9) | 23.5% (4) |
| **Rule-Based Systems** | 18.2% (6) | 25% (10) | 3.9% (1) | 33.3% (3) | 18.9% (17) | - |
| **Traditional/Statistical Learning Methods** | 27.3% (9) | 15% (6) | 19.2% (5) | 11.1% (1) | 37.8% (34) | 47.1% (8) |
| **Transformers and Large Language Models** | 6% (2) | 7.5% (3) | 19.2% (5) | - | 16.7% (15) | 17.7% (3) |

## Context Analysis Task

Context analysis encompasses a set of methods that are designed to extract and interpret the nuanced meanings of text data beyond the literal content (13, 137). Context Analysis comprises several key tasks that allow for a deeper understanding of the text by considering how the context of the surrounding affects the meaning of words and phrases and includes *negation detection* (39.1%, n = 27), *uncertainty/assertion* (21.7%, n = 15), *temporality* (4.4%, n = 3), *abbreviation* (17.4%, n = 12), and *content analysis* (17.4%, n = 12). The overall adoption rate of the NLP model/methods corresponding to these context analysis tasks is depicted in Table 6. Similar to the information extraction task, there is high interest in incorporating domain knowledge during the crafting of the input features as demonstrated by the high adoption of rule-based systems and traditional/Statistical learning methods for this task. The *negation detection* task is predominated by rule-based systems (48.2%, n = 13) and traditional/statistical learning methods (40.7%, n = 11), while there are some utilizations of transformers and large language models (7.4%, n = 2), and interpretable methods (3.7%, n = 1). *Uncertainty/Assertion* predominantly utilized traditional/statistical learning methods (66.7%, n = 10), followed by rule-based systems (26.7%, n = 4), and there is small utilization of transformers and large language models (6.7%, n = 1). *Temporality analysis, Content Analysis,* and *Abbreviation* decoding are primarily performed with rule-based systems and traditional/statistical learning methods, but to a varying extent.

*Table 6: Adoption rate of different NLP methods for context analysis tasks.*

| **Model/Method** | **Context Analysis Task** | | | | |
|---|---|---|---|---|---|
| | Negation detection | Uncertainty/ Assertion | Temporality | Abbreviation | Content analysis |
| **Deep/Nonlinear Learning Methods** | - | - | - | - | - |
| **Ensemble Methods** | - | - | - | - | - |
| **Hybrid Methods** | - | - | - | - | - |
| **Interpretable Methods** | 3.7% (1) | - | - | | - |
| **Rule-Based Systems** | 48.2% (13) | 26.7% (4) | 66.7% (2) | 41.7% (5) | 41.7% (5) |
| **Traditional/Statistical Learning Methods** | 40.7% (11) | 66.7% (10) | 33.3% (1) | 58.3% (7) | 58.3% (7) |
| **Transformers and Large Language Models** | 7.4% (2) | 6.7% (1) | - | - | - |

## *Feature representation techniques*

There are various ways of representing text data, and for the sake of brevity, the feature representation techniques reported in the reviewed studies are categorized into *statistical features* (e.g. n-grams, skipgrams, bag-of-words, tf-idf, co-occurrence matrix, and others), *orthographic word features* (e.g. tags, anchor specification, and others), *ontologies* (e.g. dictionary, ontology, and others), *learned word-level embeddings* (e.g. context vectors, Word2vec, GloVE, Fasttext, Starspace, and others), and *learned sentence/document embeddings* (e.g. topic vectors, and others) (138-141).

The findings demonstrate that *statistical features* (42.9%, n = 100) are the most utilized among the feature representation techniques, reflecting the preference for techniques that are more or less easy to interpret. Statistical features describe the distribution and frequency of words within a text, in the form of frequency counts, the presence or absence of specific terms, and other numerical representations derived from the text data (140, 141). The second most common technique is *learned word-level embeddings* (26.2%, n = 61), which involves learning representations of words in a continuous vector space where semantically similar words are mapped to nearby points, reflecting the growing interest in capturing semantic relationships and meanings of words as they appear in clinical text (140, 141). The third most common technique is *ontologies* (13.7%, n = 32), which are a structured framework that represents a set of concepts within a domain and the relationships between those concepts (140, 142), and the moderate use depicts the interest in incorporating domain-specific knowledge in the NLP models. The fourth most common technique is *learned sentence/document embeddings* (11.6%, n = 27), which is like word-level embeddings, but instead, entire sentences or documents are represented as vectors in a continuous space. The fifth most common technique is *orthographic word features* (5.6%, n = 13), which relate to the specific characteristics of word forms themselves, including spelling, capitalization, punctuation, and other aspects of the way words are written (140).

The landscape of feature representation techniques utilized by different kinds of models is depicted in Table 7. Statistical features (52.5%, n = 32) and ontologies (41.0%, n = 25) are dominant in a rule-based system, whereas statistical features (59.6%, n = 65) and learned word-level embeddings (29.4%, n = 32) are favored by interpretable methods. Like the interpretable methods, ensemble methods utilize often statistical features (50.0%, n = 21) and learned word-level embeddings (40.5%, n = 17). By the same token, statistical features (40.1%, n = 59) and learned word-level embeddings (39.5%, n = 58) are more prevalent in traditional/statistical learning methods, but ontologies (7.5%, n = 11), and orthographic word features (8.8%, n = 13), are also present but to a lesser extent. Deep/nonlinear learning methods commonly utilize learned word-level embeddings (52.0%, n = 51), and statistical features (44.9%, n = 44), and the strong preference for word-level embeddings underscores the importance of capturing intricate, nonlinear semantic patterns in text, which deep learning models excel at modeling. Transformer and large language models utilize learned sentence/document embeddings (100%, n = 37) as the sole feature representation techniques. On the other hand, hybrid models, which presumably combine different modeling techniques, show a strong reliance on learned word-level embeddings (46.7%, n = 14), and statistical features (33.3%, n = 10), and also used the others but to a lesser extent.

Table 7: Examining feature representation techniques utilized in the studies.

| Model/Method | Feature representation techniques | | | | |
|---|---|---|---|---|---|
| | Statistical Features | Ontologies | Orthographic Word Features | Learned Word-level Embeddings | Learned Sentence/ Document Embeddings |
| **Rule-Based Systems** | 52.5% (32) | 41.0% (25) | 3.3% (2) | 3.3% (2) | - |
| **Interpretable Methods** | 59.6% (65) | 3.7% (4) | 6.4% (7) | 29.4% (32) | 0.9% (1) |
| **Ensemble Methods** | 50.0% (21) | 4.8% (2) | 4.8% (2) | 40.5% (17) | - |
| **Traditional/Statistical Learning Methods** | 40.1% (59) | 7.5% (11) | 8.8% (13) | 39.5% (58) | 4.1% (6) |
| **Deep/Nonlinear Learning Methods** | 44.9% (44) | - | 1.0% (1) | 52.0% (51) | 2.0% (2) |
| **Transformers and Large Language Models** | - | - | - | - | 100.0% (37) |
| **Hybrid Methods** | 33.3% (10) | 3.3% (1) | 6.7% (2) | 46.7% (14) | 10.0% (3) |

#### *Resource Availability - Code and pre-trained model*

A high state of resource sharing and accessibility in terms of experimental code and pre-trained model plays a crucial role in NLP model development (143, 144). The resource availability presented here is based on whether the authors of the studies reviewed reported access to the code used in model development and if there is a pre-trained model released. In this regard, the result demonstrated that, overall, there is low availability of codes and pre-trained models. Across all languages, there's low overall availability of codes and pre-trained models, with only 18.6% affirming availability (Yes: 21, No: 92). This could indicate a significant gap in open-access resources for clinical NLP applications in Scandinavian texts, which could hinder the progress and reproducibility of research in this field. In comparison, despite the small number of publications, Danish exhibit the highest availability rate (25%) among the three (Yes: 3, No: 9), followed by Norwegian with 22.7% (Yes: 5, No: 17). On the other hand, the Swedish exhibit a lower availability rate 16.1% (Yes: 13, No: 68).

#### *Adaptation and Transfer learning*

In model development, adaptation and transfer learning are essential tools to address challenges related to data scarcity, and low computational resources, especially in low-resource language settings (145). Given the close linguistic similarities among mainland Scandinavian languages (15), one could expect that successful techniques and models in one language could be readily adapted to another, potentially boosting the efficiency of research and application development in the region. However, the result of the assessment regarding adaptation and transfer learning in the region indicates otherwise. Accordingly, the overall rate of adaptation and transfer learning in the region is relatively low at 19.5% (Yes:22, No:91). Among the three countries, Danish exhibits a significant research activity to leverage adaptation and transfer learning at 66.7% (Yes:8, No:4). Norwegian 13.6% (Yes:3, No:19) and Swedish 16.1% (Yes:13, No:68) exhibit relatively low rate of adaptation and transfer learning.

### 3.2.4. Clinical Characteristics

This section presents an overview of the clinical aspects presented in the studies such as whether the studies had performed some kind of implementation and testing of the model within a clinical setting, overall distribution of focus area in terms of disease area or health domain, and types of hospital departments or clinical unit considered along with the extent of medical experts in data preparations.

#### *Applied in Clinical Settings*

Despite the rapid advancement and widespread adoption of artificial intelligence (AI) and machine learning (ML) models across various industries, integrating trained models seamlessly into clinical settings continues to pose a significant challenge (146). In this regard, the reviewed studies were assessed to provide an overview of the efforts to validate the developed models clinically. The findings revealed that only a small fraction of the studies 5.31% (Yes:6, No:107) had tried to conduct some form of clinical validation, highlighting the significant challenges in the healthcare domain. Specifically, Norway 9.01% (Yes:2, No:20) demonstrated a slightly higher tendency to test the models in a clinical setting than Sweden 4.9% (Yes:4, No:77) and Danish. In contrast, Danish shows no implementation (0%, n=0), with all surveyed studies (100%, n=12) not testing the model in real-world clinical settings.

#### *Disease Area/Health Domain*

Analysis of the type of disease area or health domain the study focuses on could provide several interesting insights into research gaps and opportunities, and neglected disease areas, and could guide future research directions. In this regard, an assessment of the reviewed studies was conducted to provide an overview of the disease area or health domain the studies focus on. For the sake of brevity, the disease areas mentioned in the studies were categorized into 13 distinct and representative groups: *Infectious diseases,*

*Adverse events*, *Cardiovascular disorders*, *Gastrointestinal disorders*, *Musculoskeletal and connective tissue disorders*, *Cancer and systemic disorders*, *Surgical and postoperative complications*, *Hematologic and immune disorder*, *Neurological and psychiatric disorder*, *Behavioural and lifestyle factors*, *Family and genetic history*. The analysis of the relevant studies was performed by counting the frequency of disease areas or health domains reported within the studies dataset. Studies that did not specifically mention the disease area or health domain are categorized under *General conditions and Procedures*. In this regard, as can be seen from the pie chart in Figure 8, the majority of the studies for some reason didn't report the type of disease area considered (28.9%, n=76/263). Apart from this, overall, the findings revealed that infectious diseases (12.2%, n = 32), adverse events (9.9%, n = 26/263), cardiovascular disorders (8.0%, n = 21/263), gastrointestinal disorders (8.7%, n = 23/263), and musculoskeletal and connective tissue disorders (10.3%, n = 27/263) are the most studied groups. Others receive fair consideration including cancer and systemic disorders (4.6%, n=12/263), surgical and postoperative complications (4.2%, n = 11/263), and earn, nose and throat (ENT) (5.7%, n = 15/263). Others with fairly small consideration include neurological and psychiatric disorders (2.6%, n = 7/263), hematologic and immune disorders (2.3%, n = 6/263), behavioral and lifestyle factors (1.5%, n = 4/263), and family and genetic history (1.1%, n = 3/263).

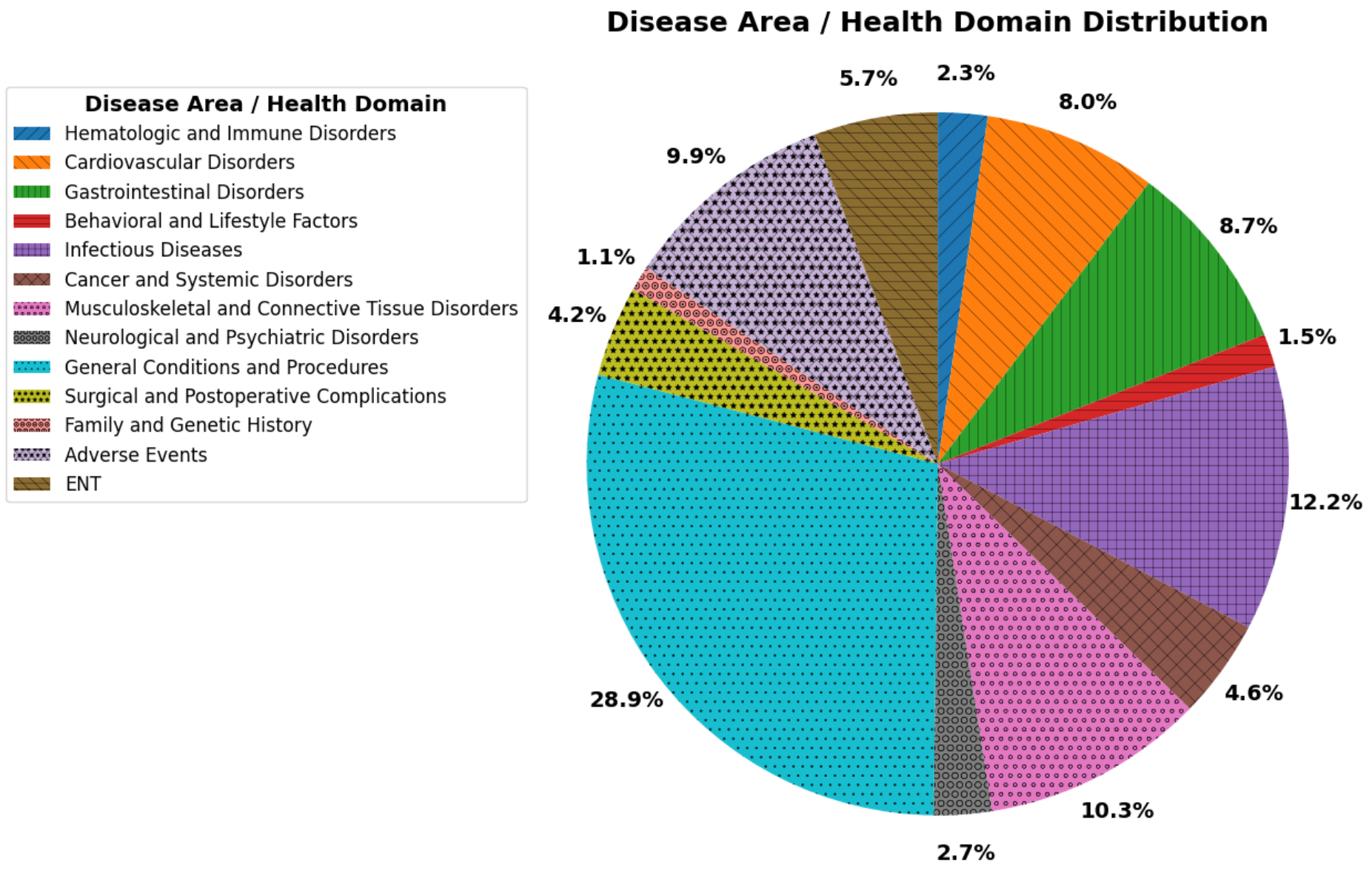

*Figure 8: Distribution of disease area or health domain the studies considered. ENT stands for ear, nose, and throat.*

Assessment of the disease area or health domain considered in the study along with the various NLP tasks is given in Table 8 below. As can be seen from the table, information extraction (18.6%, n =13) is particularly dominant in *adverse events*, which might be related to the need to accurately capture, categorize, and report any unexpected medical adverse events that require medical intervention including drug-drug interactions. On the other hand, classification (27.5%, n=11) is dominant in *Gastrointestinal disorders*, which might be related to the overlapping symptoms present in gastrointestinal disorders (147). Context analysis is fairly present in all the disease areas, and a little higher in *Gastrointestinal disorders and Infectious diseases* demonstrating its overall significance.

Table 8: NLP Tasks focus on areas of disease types or health domains.

| Disease area/Health domain | NLP Tasks | | | | |
|---|---|---|---|---|---|
| | Information extraction | Classification | De-identification | Context analysis | Resource development |
| ***Hematologic and Immune Disorders*** | 4.3% (3) | 2.5% (1) | - | 9.1% (1) | 3.2% (1) |
| ***Cardiovascular Disorders*** | 10.0% (7) | 7.5% (3) | 10.7% (3) | 9.1% (1) | 22.6% (7) |
| ***Gastrointestinal Disorders*** | 5.7% (4) | 27.5% (11) | 10.7% (3) | 18.2% (2) | 3.2% (1) |
| ***Behavioral and Lifestyle Factors*** | 1.4% (1) | 2.5% (1) | - | 9.1% (1) | 3.2% (1) |
| ***Infectious Diseases*** | 14.3% (10) | 17.5% (7) | 17.9% (5) | 18.2% (2) | 22.6% (7) |
| ***Cancer and Systemic Disorders*** | 8.6% (6) | 10.0% (4) | 3.6% (1) | 9.1% (1) | - |
| ***Musculoskeletal and Connective Tissue Disorders*** | 17.1% (12) | 7.5% (3) | 25.0% (7) | 9.1% (1) | 9.7% (3) |
| ***Neurological and Psychiatric Disorders*** | 4.3% (3) | 2.5% (1) | - | 9.1% (1) | 3.2% (1) |
| ***Surgical and Postoperative Complications*** | 4.3% (3) | 12.5% (5) | - | - | 9.7% (3) |
| ***Family and Genetic History*** | 1.4% (1) | - | 3.6% (1) | - | 3.2% (1) |
| ***Adverse Events*** | 18.6% (13) | 10.0% (4) | 7.1% (2) | 9.1% (1) | 16.1% (5) |

## *Hospital Departments/Clinical Unit*

This section aims to present an overview of the hospital departments or clinical units that provide a dataset for the studies. For the sake of brevity, the analysis was performed by grouping the departments reported in the studies into *intensive and critical care, surgical specialties, medical specialties, oncology and hematology, infectious diseases and pathology, diagnostic and therapeutic services, general services, women's health, pediatrics, other specialties, and administrative*, see Appendix A. The analysis was conducted by counting the frequency of the departments reported as a data source. Studies that didn't report this information were categorized under general services. In this regard, as shown in Figure 9, surgical specialties are the most widely reported source of data (23.0%, n = 89/387), followed by medical specialties (19.1%, n = 74/387), intensive and critical care (11.1%, n = 43/387). Some groups take a small share including infectious diseases and pathology (5.4%, n = 21/387), diagnostic and therapeutic services (4.1%, n = 16/387), oncology and hematology (3.4%, n = 13/387), women's health (2.6%, n = 10/387), other specialties (7.3%, n = 28/387), pediatrics (0.5%, n = 2/387), and administrative (5.7%, n = 22/387).

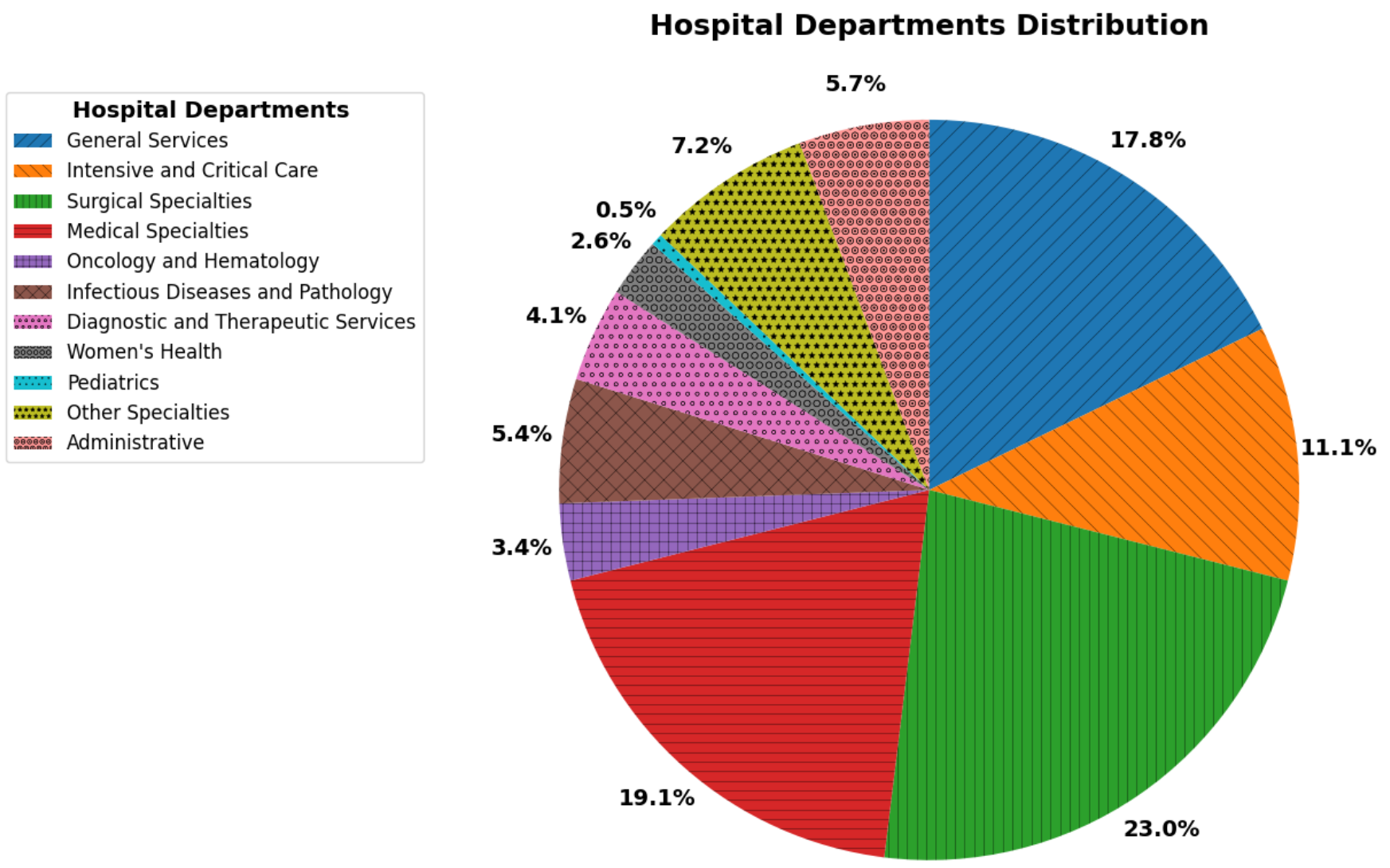

Figure 9: Distribution of hospital departments as data source reported in the studies.

### *Data Annotation*

Data annotation is a crucial step in model development, and the active participation of medical experts in this process is critical in setting the quality of the ground truth right (148). In this regard, this section presents an analysis that is performed by counting the active involvement of medical experts (occupation and experience in years) in data preparation in each study. It aims to provide insight into the extent to which the study involves domain experts in the dataset's preparation or analysis process, highlighting their contribution to data curation, annotation, or validation. It underscores the level of specialized knowledge applied to ensure the accuracy, relevance, and reliability of the data. Accordingly, the findings revealed that, overall, the active participation of medical experts is satisfactory at 64.6% (Yes:73, No:40). It should be noted that those studies that didn't mention the active participation of medical experts could most probably be using second-hand data that were already annotated, or something else. Specifically, Norwegian 90.9% (Yes:20, No:2) and Danish studies 91.7% (Yes:11, No: 1) reported a significant level of active rate of medical expert involvement compared to Swedish 54.3% (Yes:44, No: 37). This may be due attributed to Swedish studies' heavy reliance on a single infrastructure, namely *Health Record Research Bank*, which is already pre-annotated.

## Discussion

The advancement of deep learning, along with the recent introduction of transformers, coupled with the exponential growth of digital data has contributed to the success of NLP in various domains (149-152). Despite the advancements of NLP methods for resource-rich languages like English, these advancements are not equally transferable to low-resource languages (13). The objective of this review was to provide a comprehensive assessment of the state-of-the-art NLP techniques for the mainland Scandinavian clinical domain.

Although there are promising developments, the adoption of these technologies to clinical text is slower compared to other types of text. For example, the adoption of transformer-based models like ClinicalBERT (153), and BlueBERT (154) for English clinical texts occurred one year after the introduction of BERT (155) in 2018. On a similar note, the review identified the adoption of transformer-based models to Scandinavian clinical text is slower than in English, the earliest being in 2021. SweDeCin-BERT was introduced in 2022 (115). However, among the three languages, there exist noticeable gaps in adopting the transformer-based model, where research focusing on Swedish and Danish clinical text exhibits a satisfactory level of adoption, whereas there is little to none for Norwegian clinical text. Generally, this slower uptake compared to English could be attributed to several factors but mainly due to the availability of shared resources such as publicly available datasets, e.g. MIMIC (156), and necessary tools for processing text.

The most recent advances in Large Language Models (LLMs) including the introduction of ChatGPT (Chat Generative Pre-trained Transformer) have sparked intense discussion among experts on the potential use of these models in healthcare settings (157-159). The diverse potential applications range from clinical tasks, such as diagnostic assistance, patient triage, and automatic treatment plan generation, to documentation tasks, such as summarization, de-identification, coding, and automatic structuring of clinical narratives, and finally in medical research and education (8, 10, 159, 160). However, despite these promising applications, certain challenges need to be addressed before actual use including risks associated with ethics and privacy, incorrect and biased decisions, error accountability, technology misconceptions, and others (159, 161). To this end, several pieces of research have been conducted to evaluate these LLMs under different clinical tasks mainly for resource-rich languages like English (8-10, 160). For instance, Liu et. al. recently published an article focusing on the development of a de-identification framework to evaluate GPT-4 based on zero-shot learning, and despite some of the shortcomings, the study demonstrated the application of GPT-4 for a de-identification task (162). On a similar note, however, the review identified very limited research activities focusing on the utilization of these models in various clinical tasks for mainland Scandinavian clinical tasks. Nevertheless, there was one recent article that demonstrated the use of the GPT model in synthetic clinical data generation and de-identification tasks for Norwegian clinical text (40). Lund et. al. employed GPT for the generation of Norwegian discharge summaries and evaluated and compared the GPT model along with other baseline models for the de-identification task, where the GPT model is promoted to either annotate or redact personal identifying information based on instruction tunning (40).

The trend of publication for the three countries varies over the years, and the number of publications focusing on Swedish clinical text is significant compared to the other two. Despite these gaps, publications focusing on Norwegian and Danish clinical text appeared to catch up notably after 2023, indicating the increased activities in clinical NLP research. There could be several contributing factors to the success of publications focusing on Swedish clinical text, but the prominent factor is the presence of a strong research infrastructure with access to data– "*Swedish Health Record Research Bank*" (133, 163), and others like early commitment and the presence of a research ecosystem consisting of various stakeholders are also among the contributing factors. An ongoing national initiative like "*Health Data Sweden*" (164), which aims to further develop and strengthen the research ecosystem, is expected to further boost clinical NLP research activities.

There exists literature that points out the advantage of utilizing language similarities, especially for the mainland Scandinavian languages (15-17). For example, Faarlund et. al. even argue that these languages are basically the same (17). Further, Sahlgren et. al. promote joint Nordic LLMs considering the language and cultural similarities as their argument cases (16). In this regard, for instance, studying these languages together can provide evidence for differences and similarities in terms of linguistic characteristics (83) to lay the groundwork for technology transfer, and also can be an effective tool for utilizing data augmentations and transfer learning (15, 41). Despite these potential gains, the most intriguing observation is the least frequent co-occurrence of the three languages – Swedish, Norwegian, and Danish. There is little to no effort in studying these languages together, even though these languages share linguistic similarities with countries that have similar healthcare systems, cultural contexts, and social values.

From another perspective, collaboration and knowledge sharing among different institutions are vital components of a well-functioning research ecosystem, and also a key driver for the success of a project requiring interdisciplinary cooperation in a field like clinical NLP. In this regard, overall, the result depicts a productive research collaboration network among the mainland Scandinavian countries, except Denmark, which has very limited ties, only having one occasion with Sweden depicting a one-time or project-specific collaboration rather than an ongoing research partnership. In the future, we envision more productive and active research collaborations than one-time or project-specific relationships among the mainland Scandinavian countries to facilitate and further foster clinical NLP research and development, given their close similarities in many aspects.

Enhancing reproducibility and the availability of shared resources in clinical NLP research are essential for accelerating the development of clinical NLP. In this context, research endeavors should focus on providing a dataset and, where applicable, its annotated version - corpora, tools, and resources such as dictionaries, lexicons, and annotation standards, pre-trained models, and experimentation code along with clear documentation of procedures and requirements, as these are the main ingredients for rapid development. However, it is often challenging to publicly share datasets due to the sensitivity of health data, which contain patient-identifiable information such as names, addresses, telephone numbers, and others, and the privacy and confidentiality concerns that come along with it. A de-identification procedure performed to preserve individual privacy and confidentiality remains a significant challenge, especially on a large scale. A recent study conducted by Berg et.al. (62) assess the impact of de-identification on data quality for downstream named entity recognition tasks, and highlight the importance of balancing the level of de-identification and the utility of the data. In general, this review identified that the majority of research activity focusing on the de-identification task was conducted for Swedish clinical text, and there was little effort for the other two languages. Regardless, with some constraints, studies still can share pre-trained models, and experimentation code, which undeniably have significant impacts. In this regard, the most intriguing observation is, despite the apparent challenge of sharing clinical text data, the review identifies high availability of data (31.86%, n = 36/113) compared to experimentation code and pre-trained models, with only 18.6% affirming availability (n=21/113). These observations can be attributed to factors such as intellectual property issues, and the need for extensive documentation, and, in some cases, researchers might also withhold models and experimentation code to gain competitive advantages, and possibly if there is a commercial interest. The reflection of this lack of shared resources including experimentation code and pre-trained model can be observed by closely looking at the rate of adaptation and transfer learning in the region, which is relatively low at 19.5% (n =22/113), despite the potential advantages of language similarity as mentioned above.

When it comes to resource development, such as dictionaries, lexicons, or annotation standards, the review identified fair endeavors, where 32% of the studies (n=37 /113) have engaged in such resource development either as their main or an additional contribution. However, this number greatly varies among the three languages, with Danish exhibiting the lowest and Swedish the highest level of activities in this direction. Generally, given the low-resource nature of these languages, extensive research endeavors are expected to enrich the available resources for language processing tasks, potentially opening doors for enormous opportunities similar to those enjoyed by resource-rich languages like English. Regarding the development of annotation standards and methods, there are some research activities (24, 72, 76, 128), but are more of case-specific development than general purposes. The involvement of medical experts in developing annotation guidelines, as well as directly in data curation and annotation processes, is crucial to ensure the integrity of training and validation datasets. In this regard, the review identifies a satisfactory level of active medical expert involvement though the rates vary among the three languages. On a different note, it generally remains challenging to clinically validate models in real-world clinical settings, and most developed models fail to transition into actual clinical use. There are several contributing factors to this including stringent regulatory and compliance requirements, difficulties integrating models with existing systems and workflows, concerns over maintaining patient privacy, lack of standardization across data collection and processing, the inherent complexity of clinical environments, and other similar challenges. By the same token, the review identified that only a small fraction of the studies 5.31% (n = 6/113) had tried to conduct some form of clinical validation, highlighting the significant challenges in the healthcare domain, and there is no noticeable difference between the three languages. In this regard, for example, Laursen MS et. al. (165), developed an AI model capable of detecting hemorrhage events and evaluated the performance gained by assisting medical doctors with AI during chart review in a mock-up setting.

# Conclusion

The review presented a comprehensive assessment of the state-of-the-art Clinical NLP for EHR text in Scandinavian languages and, highlighted the potential barriers and challenges that hinder the rapid advancement of the field in the region. The review identified a lack of shared resources, such as datasets, pre-trained models and tools, inadequate research infrastructure, and insufficient collaboration as the most significant barriers that require careful consideration in future research endeavors. For example, the disparity in research activities among the three languages underscores the importance of robust research infrastructure, as demonstrated by the significant impact of the "*Swedish Health Record Research Bank*". Further, despite the close similarities among these languages, there has been minimal effort to study these languages together, which can potentially open up enormous opportunities. Future research should focus on exploring the potential of data augmentation, adaptation, and transfer learning across the three languages to facilitate progress. Starting in 2021, we observed a high research interest in adopting transformer-based models. However, there is a noticeable gap between the three languages; Swedish and Danish have shown significant activities, while Norwegian has exhibited minimal activities. Despite the recent booming of LLMs like GPT with enormous potential, there have been limited research activities, and the review identified only one study focusing on Norwegian clinical text. Given the low-resource nature of these languages, more research activities should be targeted toward resource development, core NLP tasks, and de-identification. In this regard, research focusing on Swedish clinical text is more prominent, while the other two languages, Danish and Norwegian, require further research efforts in this direction. The availability of research infrastructure, including data and computational resources, is a key driving factor for the success of NLP development, underscoring the need for increased efforts in this area. Generally, we foresee the findings presented in this review will help shape future research directions by shedding light on areas that require further attention for the rapid advancement of the field in the region.

## Acknowledgment

This work was financially supported by the Research Council of Norway (RCN), through its Centre for Research-based Innovation funding scheme (Visual Intelligence, grant no. 309439), and RCN IKTPLUSS, grant no. 303514.

## Conflict of Interest

The authors declare that they have no known competing financial interests or personal relationships that could have appeared to influence the work reported in this paper.

# Appendix A

## Extended Definition of Categorization Used in Data Collection

Categories used in extracting information from the articles included in the study. These categories are solely determined to assess and evaluate the included articles and are pre-defined based on previous knowledge, brainstorming, discussion among the authors, and literature (1, 2).

- ***Clinical Application and Technical Objective***: This category identifies the precise clinical challenges or healthcare goals the study intends to address through Natural language processing (NLP). It also identifies the technical objective defined to address the clinical needs, emphasizing the technical contribution of the study to solve the clinical challenges.
- ***Language*****:** This category identifies and pinpoints the language of the processed clinical text in each of the reviewed studies. It can be either *Swedish, Norwegian, Danish, or other languages* studied along with these three main languages.
- ***NLP Task***: This category defines the high-level grouping of the NLP task considered in each of the reviewed studies. It can take any of the following: *core NLP, resource development, de-identification, information extraction, classification, context analysis, and multi-lingual tasks*. Each group can contain several types of methods outlined under the "*type description"* category below.
- ***Type Description***: This category identifies and pinpoints the specific group of NLP methods, which are the sub-category of each NLP task defined above, used or implemented in each of the reviewed studies and it can be *part-of-speech tagging, negation detection, drugs/adverse events, uncertainty/assertion*, *named entity recognition, and others*.
- ***Model and Method***: This category identifies and pinpoints the type of NLP models or methods proposed, and developed in each of the reviewed studies and it can take any of the following: *rule-based, interpretable methods, ensemble, deep/non-linear learning, hybrid, and transformer and large language mode*l.
- ***Code Availability, and pre-trained models*****:** This category identifies if the studies had released any accompanying code related to the experiments conducted and further looks if there are any reported releases of pre-trained models.
- ***Adaption & Transfer learning***: This category identifies and pinpoints if there is some form of adaptation and transfer learning used including, for example, *data augmentation, transfer learning*, etc.
- ***Data Sources and Names***: This category specifies the origin and name of the dataset used in the study, detailing the institutions or platforms from which it was sourced, and the specific names or identifiers of the datasets used.
- ***Other data*****:** This category identifies, and pinpoints datasets other than clinical free text utilized in the studies and can include *laboratory measurements, codes, registry, and others*.
- ***Type of EHR documents***: This category identifies the specific type of Electronic Health Record (EHR) documents targeted in each of the reviewed studies (e.g., *Progress notes, lab reports, pathology reports, discharge summaries, consultation reports, medication lists, immunization records, and others*).
- ***Text Preprocessing***: This category identifies the preprocessing strategies used to refine and transform the clinical text data into a format more suitable for feature representation. It can be *tokenization, sentence splitter, stop word removal, stemming, lemmatization, PoS-tagger, and others*.
- ***Feature Representation Techniques***: This category outlines the specific methods used to convert the preprocessed clinical text data or words into a numerical representation that the NLP models can understand and analyze. There are various kinds of feature representation available and can include *Word2Vec, Phrase2Vec, GloVe, FastText, TF-IDF, N-grams, Bag-of-words, and other* representation techniques used in NLP.
- ***Disease area/Health domain*****:** This category identifies and pinpoints the specific type of disease or health domain targeted by the reviewed studies. Whenever possible, the reported disease areas are roughly mapped into the existing classifications as defined by the Clinical Data Interchange Standards Consortium https://www.cdisc.org/standards/therapeutic-areas/disease-area.
- ***Departments/Clinical Unit*****:** This category identifies the specific department or clinical unit within the healthcare setting in which the dataset was retrieved for each of the reviewed studies and aims to pinpoint the group of clinical units that were studied most or least.

- ***Study Period*****:** This category captures the dataset's temporal dimensions, providing insights into the specific timeframe over which the data was collected. It outlines the duration, including start and end dates if applicable.
- ***Dataset Size*****:** This category pinpoints the total volume of the dataset used, detailing the number of patients, records, documents, instances, or tokens it comprises.
- ***Training/Test /Validation size*****:** This category details the division of the dataset into training, test, and validation subsets, offering insights into the allocation of data for model development, evaluation, and validation.
- ***Data Availability*****:** This category identifies if the dataset used in each of the reviewed studies is accessible for other researchers, indicating how and where the dataset can be accessed, whether it is publicly available, restricted to certain users, or available upon request.
- ***Annotated Set*****:** This category pinpoints the size of the dataset that had been annotated, detailing the volume of data that has undergone labeling processes.
- ***Domain Expert*****:** This category identifies the involvement of domain experts in the dataset's preparation or analysis process, highlighting their contribution to data curation, annotation, or validation.
- ***Data quality and related issues***: This category identifies the type of data quality and other related issues reported in each of the reviewed studies. The data quality issues can incorporate any of the following: *misspellings, abbreviations, incompleteness, inconsistency, lack of standardization, inconsistencies in data representation, missing data, incorrect data labeling, or others*. Other data-related issues were also considered, e.g. *lack of public data*.
- ***Performance and efficiency***: This category identifies and pinpoints the performance reported in the studies and evaluation methods (performance metrics) used.
- ***Implementation Status*****:** This category provides an overview of the current state of model evaluation in real-world clinical settings and the extent of deployment and integration of these models into clinics.

## Grouping of Values in Categories for the Literature Analysis and Evaluation

After the data collection stage, where all the categories described above have been extracted from each study, for the sake of brevity, the categories were further grouped into sub-categories, and specific values were assigned to each category. The following table provides the delineation of all the values extracted from the articles, and how they were mapped or grouped under each sub-category during the analysis.

- ### *Disease area/health domain, Departments/clinical units, Type of EHR Documents, Models and methods, Feature representation techniques*

Table 1 incorporates the mapping of the values from the main category (*Disease area/health domain, Departments/clinical units, Type of EHR Documents, Models and methods, Feature representation techniques*) into sub-categories.

*Table 1: Mapping disease areas and department/clinical units, model/methods, and feature representation techniques employed in the studies.*

| Category | Sub-category | Values Assigned |
|---|---|---|
| Disease area/health domain | Hematologic and immune disorders | e.g. aplastic anemia, allergies |
| | Cardiovascular disorders | e.g. cardiac disease, venous thromboembolism |
| | Gastrointestinal disorders | e.g. gastrointestinal surgery |
| | Behavioral and lifestyle factors | e.g. patients' smoking status, nutrition |
| | Infectious diseases | e.g. infection, sepsis |
| | Cancer and systemic disorders | e.g. cancer, skin disorder |
| | Musculoskeletal and connective tissue disorders | e.g. musculoskeletal conditions |
| | Neurological and psychiatric disorders | e.g. psychiatric disorder |
| | General conditions and procedures | e.g. adverse events |
| | Surgical and postoperative complications | e.g. postoperative delirium, bleeding |
| | Family and genetic history | e.g. family history |
| | Specific conditions | e.g. oral, throat |
| | General services | e.g. nursing, ambulatory care, outpatient |
| Departments/clinical units | Intensive and critical care | e.g. intensive care unit, emergency, anesthesia |
| | Surgical specialties | e.g. orthopedic, dental, gastrointestinal, general surgery, cardiothoracic, ophthalmic, plastic surgery, neurological, oral and maxillofacial surgery, colorectal surgery, thoracic and vascular Surgery |
| | Medical Specialties | e.g. neurology, geriatric, cardiology, internal medicine, dermatology, urology, psychiatry, rheumatology |
| | Oncology and hematology | e.g. oncology, hematology, infectious diseases, pathology |
| | Diagnostic and therapeutic services | e.g. radiology, diagnostics, drugs |
| | Women's Health | e.g. gynecology, obstetrics |
| | Pediatrics | |
| | Other specialties | e.g. nutrition and dietetics, speech pathology, ENT (otorhinolaryngology), ophthalmology, maxillofacial |
| | Administrative | e.g. admission, discharge |
| Type of EHR Documents | Progress Notes | e.g. Progress, Nurse, and Physician Notes |
| | Medical History and Physical Examination Report | e.g. Admission Notes, Patient History, Assessment Entries, Chief complaint |
| | Medication List and Prescriptions | e.g. Drug information, prescription records |
| | Allergy and Adverse Reaction List | e.g. Adverse event reports |
| | Surgical Reports | e.g. Surgical reports and notes |
| | Laboratory and Diagnostic Test Results | e.g. Laboratory examination notes, radiology reports, pathology reports, diagnostic notes |
| | Consultation and Referral Reports | e.g. Outpatient and transfer notes, referrals |
| | Discharge Summary | e.g. Discharge summaries, notes, and letters |
| | Nutritional Assessments and Plans | e.g. Dietetic notes |
| | Emergency Department Reports | e.g. Emergency department notes, and reports |
| | Diverse/Not specific | Not mentioned in the study |
| | Clinical Notes | e.g. Clinical Notes, text, narrative, EHR notes, narratives, text, journal notes |
| Model and methods | Rule-based systems | e.g. rule-based systems |
| | Interpretable methods | e.g. decision tree, naïve Bayes, KNN, logistic regression |
| | Ensemble methods | e.g. random forest |
| | Traditional/statistical learning methods | e.g. CRFs, SVMs |
| | Deep/nonlinear learning methods | e.g. MLP, LSTM, GRU, CNN |
| | Hybrid methods | e.g. LSTM+CNN |
| | Transformers and large language models | e.g. BERT, GPT |
| Feature representation techniques | Statistical features | e.g. N-grams, Skipgrams, Bag-of-words, TF-IDF, Co-occurrence matrix |
| | Orthographic word features | e.g. Tags, Anchor specification |
| | Ontologies | e.g. Dictionary, Ontology |
| | Learned word-level embeddings | e.g. Context vectors, Word2vec, GloVE, Fasttext, Starspace |
| | Learned sentence/document embeddings | e.g. Topic vectors |

- ***Mapping of extracted values into keywords: Data quality and related issues***

The values extracted on data quality and related issues from the reviewed studies are further mapped into proper keywords, and the table below defines the keywords used in analyzing the reported issues. The categories "Others" are created to include issues that are project-specific and have limited horizons and only happen to the specific data considered in that particular study.

*Table 2: Mapping the data quality and other related issues reported in the studies.*

| Data Quality and Related Issues | Working definition |
|---|---|
| *misspelled* | Issues related to misspellings or typographical errors in the data |
| *multi-word* | Issues related to phrases or terms that consist of multiple words |
| *abbreviation* | Issues related to abbreviations or shortened forms of words |
| *reference resolution* | Issues related to challenges in linking mentions of entities to the same underlying real-world entities. |
| *lexical variation* | Issues related to variability in how a concept or entity is expressed, such as synonyms or different ways of writing the same term, e.g., medical jargon, and acronyms. |
| *incomplete data* | Issues related to missing information or data entries |
| *inconsistent data* | Issues related to inconsistencies in the data, such as conflicting information. |
| *noisy data* | Issues related to data that contains errors, irrelevant information, or is otherwise not clean. |
| *duplicate data* | Issues related to repeated information or entries that should be unique. |
| *formatting issues* | Issues related to problems with the way data is formatted, which could include date formats or numerical formats. |
| *handwritten text* | Issues related to text that is written by hand, often in a way that may be challenging for automated processing. |
| *Data sparsity* | Issues related to a large proportion of the data being sparse. |
| *Lack of public data* | Issues related to the scarcity of publicly available clinical datasets. |
| *Others* | Any other issues or anomalies in the data that are not covered by the specific keywords listed above |

- ***Natural language tasks***

For the sake of brevity and interpretability, the NLP strategies employed in the studies are categorized into *core NLP, resource development, de-identification, classification, context analysis, and information extraction*. A detailed description of these keywords is given below.

- *Core NLP*: This category encompasses techniques like morphology, part-of-speech tagging, parsing, and segmentation.
- *Resource development*: This category encompasses techniques like lexicons, corpora and annotation, machine-readable dictionaries, models, and methods.
- *De-identification*: refers to a set of methods used to remove personal identifiers from the clinical text data.
- *Classification*: This category encompasses several key tasks, including risk stratification of medical conditions from EHR text, indexing and coding, patient-authored text analysis, and cohort stratification, each addressing different aspects of medical data categorization and analysis.
- *Context Analysis*: This category encompasses negation detection, uncertainty/assertion, temporality, abbreviation, and experiencer.
- *Information extraction*: This category encompasses several key techniques and focus areas including medical concepts, findings/symptoms, drugs/adverse events, specific characteristics, named entity recognition, and relations.

## Extended results

- ***Data/Information Extraction***

The following tables provide information extracted from the reviewed studies. Table 3 provides information/categories mapped from the raw extracted information, which is given in Table 4 below. Table 4 provides the raw data extracted and used in the analysis. It is worth noting that all the rest of the variables described above in the categorization section can be found in the Excel file possibly attached to this publication or by contacting the corresponding author.

*Table 3: Information mapped from the raw extraction information (see Table 4).*

| Ref. | NLP Techniques | Model/Method | Feature Representation | Language |
|---|---|---|---|---|
| (3) | Information extraction (Findings/Symptoms, Named entity recognition) | Rule-Based Systems, Traditional/Statistical Learning Methods | Learnt Word-level Embeddings, Ontologies, Statistical Features, Learnt Sentence/Document Embeddings | Swedish |
| (4) | Information extraction (Named entity recognition), De-identification | Rule-Based Systems, Traditional/Statistical Learning Methods | N/A | Swedish |
| (5) | Context Analysis (Negation detection, Uncertainty/Assertion) | Traditional/Statistical Learning Methods | Learnt Word-level Embeddings, Statistical Features, Ontologies | Swedish |
| (6) | Resource development (content analysis) | Traditional/Statistical Learning Methods | Ontologies, Statistical Features | Swedish, Finnish |
| (7) | Information extraction (Named Entity Recognition) | Rule-Based Systems, Deep/Nonlinear Learning Methods | Ontologies, Statistical Features, Learnt Word-level Embeddings. Orthographic Word Features | Swedish |
| (8) | Information extraction (Drugs/Adverse events), Classification (Phenotyping from EHR text) | Ensemble Methods, Interpretable Methods | Statistical Features | Swedish |
| (9, 10) | Information extraction (Named entity recognition), De-identification | Deep/Nonlinear Learning Methods, Traditional/Statistical Learning Methods | Learnt Word-level Embeddings | Swedish |
| (11) | Information extraction (Named entity recognition), De-identification | Rule-based systems, Traditional/Statistical Learning Methods, Deep/Nonlinear Learning Methods | Ontologies, Statistical Features, Orthographic Word Features | Swedish |
| (12) | Information extraction (Named entity recognition), De-identification | Rule-Based Systems | Ontologies | Swedish |
| (13) | Information extraction (Named entity recognition), De-identification | Rule-Based Systems, Traditional/Statistical Learning Methods | Ontologies | Swedish |
| (14) | Information extraction (Named entity recognition), De-identification | Traditional/Statistical Learning Methods | NA | Swedish |
| (15, 16) | Information extraction (Medical Concepts, Findings/Symptoms), Classification (Phenotyping from EHR text) | Traditional/Statistical Learning Methods, Rule-Based Systems | Statistical Features, Learnt Word-level Embeddings | Norwegian |
| (17) | Resource development (Corpora and annotation) | NA | NA | Swedish |
| (18) | Classification (Indexing and coding) | Transformers and Large Language Models | Learnt Sentence/Document Embeddings | Spanish, Swedish, English |
| (19) | Classification (Indexing and coding) | Transformers and Large Language Models | Learnt Sentence/Document Embeddings | Spanish, Swedish |
| (20) | Information extraction (Named entity recognition), De-identification | Transformers and Large Language Models | Learnt Sentence/Document Embeddings | Swedish |
| (21) | Resource development (Corpora and annotation), Information Extraction (Named entity recognition), De-identification | Traditional/Statistical Learning Methods, Rule-Based Systems | NA | Norwegian |
| (22) | Classification (Indexing and coding), Context Analysis (Negation detection) | Transformers and Large Language Models, Interpretable Methods, Rule-Based Systems | Statistical Features, Learnt Sentence/Document Embeddings | Swedish |
| (23) | Information extraction (Specific characteristics, Named entity recognition), Classification (Phenotyping from EHR text) | Interpretable Methods, Rule-Based Systems | Statistical Features, Learnt Word-level Embeddings | Swedish |
| (24-26) | Classification (Indexing and coding) | Transformers and Large Language Models | Learnt Sentence/Document Embeddings | Swedish |
| (27) | Information extraction (Named entity recognition), De-identification | Hybrid Methods | Learnt Word-level Embeddings | Swedish |
| (28) | Classification (Phenotyping from EHR text) | Rule-Based Systems | NA | Norwegian |
| (29) | Classification (Phenotyping from EHR text) | Interpretable Methods, Deep/Nonlinear Learning Methods | Statistical Features | Norwegian |
| (30) | Resource development (Corpora and annotation), Information Extraction (Named entity recognition), De-identification | Rule-Based Systems, Transformers and Large Language Models | Ontologies, Learnt Sentence/Document Embeddings | Danish |
| (31) | Resource development (Corpora and annotation) | Deep/Nonlinear Learning Methods | Learnt Word-level Embeddings | Danish |
| (32) | Information extraction (Medical Concept, and Findings/Symptoms), classification (Phenotyping from EHR text) | Deep/Nonlinear Learning Methods, Interpretable Methods, Ensemble Methods, Hybrid Methods | Ontologies, Statistical Features, Learnt Word-level Embeddings | Norwegian |
| (33) | Resource development (Corpora and annotation), Information Extraction (Named entity recognition, and Relations) | Transformers and Large Language Models | Learnt Sentence/Document Embeddings | Danish |
| (34) | Information extraction (Named entity recognition), Classification, De-identification | Transformers and Large Language Models | Learnt Sentence/Document Embeddings | Swedish |
| (35) | Classification (Phenotyping from EHR text) | Transformers and Large Language Models, Deep/Nonlinear Learning Methods | Learnt Sentence/Document Embeddings, Learnt Word-level Embeddings | Danish |
| (36) | Resource development (Models), Classification (Phenotyping from EHR text) | Transformers and Large Language Models, Hybrid Methods | Learnt Sentence/Document Embeddings, Learnt Word-level Embeddings | Danish |
| (37) | Resource development (Corpora and annotation, and methods) | Rule-Based Systems | NA | Norwegian |
| (38) | Classification (Phenotyping from EHR text) | Interpretable Methods, Ensemble Methods | Statistical Features | Swedish |
| (39) | Resource development (Machine-readable dictionaries), Information Extraction (Named Entity Recognition, Drugs/Adverse events) | Rule-Based Systems | Ontologies | Danish |
| (40) | Information Extraction (Named Entity Recognition, Drugs/Adverse events, Relations) | Interpretable Methods, Ensemble Methods | Ontologies | Swedish |
| (41) | Information extraction (Named Entity Recognition) | Traditional/Statistical Learning Methods, Hybrid Methods | Learnt Word-level Embeddings, Orthographic Word Features, Ontologies | Swedish |
| (42) | Information extraction (Named Entity Recognition) | Transformers and Large Language Models | Learnt Sentence/Document Embeddings | Swedish |
| (43) | Resource development (Lexicons, methods) | Rule-Based Systems | Ontologies; Statistical Features | Swedish |
| (44) | Resource development (Content analysis) | Rule-Based Systems, Traditional/Statistical Learning Methods | Statistical Features | Swedish |
| (45) | Core NLP (Parsing, Part of Speech tagging) | Rule-Based Systems, Traditional/Statistical Learning Methods | NA | Swedish |
| (46) | Information Extraction (Named Entity Recognition) | Traditional/Statistical Learning Methods | Learnt Word-level Embeddings | Swedish |
| (47) | Information Extraction (Named Entity Recognition, Drugs/Adverse events) | Ensemble Methods, Deep/Nonlinear Learning Methods | Learnt Word-level Embeddings, Statistical Features | Swedish |
| (48) | Information extraction (Named entity recognition), De-identification | Traditional/Statistical Learning Methods | Learnt Word-level Embeddings, Orthographic Word Features | Swedish |
| (49) | Classification (Indexing and coding), Context Analysis (negation detection) | Traditional/Statistical Learning Methods, Rule-Based Systems | Statistical Features | Swedish |
| (50) | Classification (Indexing and coding) | Traditional/Statistical Learning Methods | Statistical Features, Learnt Word-level Embeddings | Swedish |
| (51) | Classification (Indexing and coding) | Traditional/Statistical Learning Methods | Learnt Word-level Embeddings | Swedish |
| (52) | Information extraction (Named entity recognition), De-identification | Traditional/Statistical Learning Methods | Ontologies | Swedish |
| (53) | Information extraction (Drugs/Adverse events, Named entity recognition, Relations), Context Analysis (Negation detection, Uncertainty/Assertion, Temporality), Resource Development (Corpora and annotation) | Traditional/Statistical Learning Methods | Learnt Word-level Embeddings, Statistical Features, Ontologies | Swedish |
| (54) | Information Extraction (Medical Concepts, Relations), Context Analysis (Abbreviation), Resource Development (methods) | Traditional/Statistical Learning Methods | Learnt Word-level Embeddings, Ontologies, Statistical Features | Swedish |
| (55) | Resource Development (methods), Information Extraction (Medical concept, Relations) | Traditional/Statistical Learning Methods | Learnt Word-level Embeddings, Statistical Features | Swedish |
| (56) | Resource Development (methods), Information Extraction (Drugs/Adverse events) | Ensemble Methods | Learnt Word-level Embeddings, Statistical Features | Swedish |
| (57) | Information extraction (Drugs/Adverse events) | Deep/Nonlinear Learning Methods, Ensemble Methods | Learnt Word-level Embeddings | Swedish |
| (58) | Resource Development (methods), Information Extraction (Drugs/Adverse events) | Deep/Nonlinear Learning Methods, Ensemble Methods | Learnt Word-level Embeddings | Swedish |
| (59) | Classification (Phenotyping from EHR text) | Deep/Nonlinear Learning Methods, Traditional/Statistical Learning Methods | Statistical Features, Learnt Word-level Embeddings | Swedish |
| (60) | Classification (Phenotyping from EHR text) | Rule-Based Systems, Interpretable Methods | Statistical Features | Norwegian |
| (61) | Information extraction (Medical Concepts) | Transformers and Large Language Models | Learnt Sentence/Document Embeddings | Swedish |
| (62) | Information extraction (Drugs/Adverse events) | Ensemble Methods, Traditional/Statistical Learning Methods | Learnt Word-level Embeddings, Statistical Features | Swedish |
| (63) | Context analysis (abbreviation), Classification | Rule-Based Systems | NA | Swedish |
| (64) | Information extraction (Named entity recognition), De-identification, Classification (indexing and coding), Context analysis (uncertainty/assertion classification) | Transformers and Large Language Models | Learnt Sentence/Document Embeddings | Swedish |
| (65) | Information extraction (Named entity recognition), De-identification, Classification (indexing and coding) | Transformers and Large Language Models | Learnt Sentence/Document Embeddings | Swedish |
| (66) | Classification (Indexing and coding), De-identification, Information Extraction (Named entity recognition, Negation detection, Findings/Symptoms, Drugs/Adverse events), Context Analysis (Negation detection) | Transformers and Large Language Models | Learnt Sentence/Document Embeddings | Swedish |
| (67) | Resource Development (content analysis), Context Analysis (Abbreviation) | Traditional/Statistical Learning Methods | NA | Swedish |
| (68) | Information extraction (Findings/Symptoms, Medical Concepts, Specific characteristics), Classification (Phenotyping from EHR text) | Interpretable Methods | Statistical Features, Orthographic Word Features | Norwegian |
| (69) | Resource development (Corpora and annotation, methods, content analysis), Context analysis (Negation detection, Uncertainty/Assertion) | Traditional/Statistical Learning Methods | NA | English, Swedish |
| (70) | Resource development (Lexicons), Core NLP (Part of Speech tagging, Parsing) | Rule-Based Systems | Ontologies | Swedish |
| (71, 72) | Classification (Phenotyping from EHR text) | Transformers and Large Language Models | Learnt Sentence/Document Embeddings | Swedish |
| (73) | Classification (Phenotyping from EHR text) | Deep/Nonlinear Learning Methods | NA | Danish |
| (74) | Classification (Phenotyping from EHR text), Information extraction (Drugs/Adverse events) | Transformers and Large Language Models, Deep/Nonlinear Learning Methods, Ensemble Methods | Learnt Sentence/Document Embeddings, Statistical Features, Ontologies | Swedish |
| (75) | Resource development (Corpora and annotation), Information extraction (Named entity recognition), De-identification | Transformers and Large Language Models | Learnt Sentence/Document Embeddings | Norwegian |
| (76) | Information extraction (Named entity recognition), Classification, De-identification | Transformers and Large Language Models | Learnt Sentence/Document Embeddings | Norwegian, Danish, Swedish |
| (77) | Classification (Phenotyping from EHR text), Information Extraction (Findings/Symptoms) | Deep/Nonlinear Learning Methods, Rule-Based Systems | Learnt Word-level Embeddings | Danish |
| (78) | Classification (Phenotyping from EHR text), Information Extraction (Findings/Symptoms), Resource development (Corpora and annotation) | Transformers and Large Language Models, Deep/Nonlinear Learning Methods | Learnt Sentence/Document Embeddings, Learnt Word-level Embeddings | Danish |
| (79) | Information extraction (Named entity recognition, Classification (Phenotyping from EHR text) | Traditional/Statistical Learning Methods, Deep/Nonlinear Learning Methods, Interpretable Methods | Learnt Word-level Embeddings, Learnt Sentence/Document Embeddings | Spanish, Swedish |
| (80) | Resource development (Lexicons, methods), Information extraction (Named entity recognition) | Rule-Based Systems | Ontologies | Norwegian |
| **(81)** | Resource development (Corpora and annotation, methods)), Information extraction (Named entity recognition, Relations), Classification (Phenotyping from EHR text) | Interpretable Methods | NA | Norwegian |
| **(82)** | Classification (Indexing and coding) | Transformers and Large Language Models, Interpretable Methods | Statistical Features, Learnt Sentence/Document Embeddings | Swedish |
| **(83)** | Information extraction (Relations), Classification (Cohort stratification, Indexing and coding) | Traditional/Statistical Learning Methods | Statistical Features, Ontologies | Danish |
| **(84, 85)** | Classification (Phenotyping from EHR text), Information Extraction (Findings/Symptoms) | Interpretable Methods | Statistical Features | Norwegian |
| **(86)** | Context Analysis (Negation detection), Classification (Medical Concepts) | Rule-Based Systems | Ontologies | Swedish |
| **(87)** | Information extraction (Named entity recognition), resource development (Corpora and annotation) | Traditional/Statistical Learning Methods | NA | Swedish |
| **(88)** | Core NLP (Parsing) | Traditional/Statistical Learning Methods, Rule-Based Systems | NA | Swedish |
| **(89)** | Information extraction (Named entity recognition) | Traditional/Statistical Learning Methods | Learnt Word-level Embeddings, Statistical Features | Swedish |
| **(90)** | Resource development (Lexicons) | Traditional/Statistical Learning Methods, Rule-Based Systems | Learnt Word-level Embeddings, Statistical Features | Swedish |
| **(91)** | Information extraction (Named entity recognition), Resource Development (Lexicons), Classification (Indexing and coding) | Rule-Based Systems | NA | Swedish |
| **(92)** | Information extraction (Named entity recognition), Classification (Phenotyping from EHR text) | Traditional/Statistical Learning Methods | NA | Swedish |
| **(93, 94)** | Classification (Phenotyping from EHR text) | Interpretable Methods, Traditional/Statistical Learning Methods | Statistical Features | Norwegian |
| **(95)** | Resource Development (methods), Context Analysis (Negation detection) | Rule-Based Systems | Ontologies | Swedish |
| **(96)** | Resource development (Lexicons), Context analysis (Abbreviation) | Rule-Based Systems, Traditional/Statistical Learning Methods | Statistical Features | Swedish |
| **(97)** | Information extraction (Findings/Symptoms, Medical Concepts), Context Analysis (Negation detection) | Rule-Based Systems | Ontologies | Danish |
| **(98)** | Classification (Phenotyping from EHR text), Information Extraction (Drugs/Adverse events, Relations, Named entity recognition) | Transformers and Large Language Models | Learnt Sentence/Document Embeddings | Swedish |
| **(99)** | Resource development (Corpora and annotation) and Context Analysis (Negation detection, Uncertainty/Assertion) | Traditional/Statistical Learning Methods | NA | Swedish |
| **(100)** | Context Analysis (Temporality) | Rule-Based Systems | NA | Swedish |
| **(101)** | Classification (Phenotyping from EHR text, Indexing and coding), information extraction (Drugs/Adverse events) | Traditional/Statistical Learning Methods | NA | Swedish |
| **(102)** | Resource development (Lexicons), Context Analysis (Uncertainty/Assertion) | Rule-Based Systems | Ontologies | Swedish |
| **(103)** | Resource development (Corpora and annotation), Context Analysis (Temporality) | Rule-Based Systems | Statistical Features | Swedish, Finnish |
| **(104)** | Resource development (Corpora and annotation, methods), Context Analysis (Uncertainty/Assertion, Negation detection) | Traditional/Statistical Learning Methods, Rule-Based Systems | NA | Swedish |
| **(105)** | Information extraction (Findings/Symptoms) | Rule-Based Systems | Statistical Features | Norwegian |
| **(106)** | Information extraction (Findings/Symptoms), Context Analysis (Negation detection) | Rule-Based Systems | Ontologies | Swedish |
| **(107)** | Information extraction (Findings/Symptoms, Named entity recognition, Specific characteristics) | Rule-Based Systems | NA | Norwegian |
| **(108)** | Resource Development (Corpora and annotation, methods), Context Analysis (Uncertainty/Assertion) | Traditional/Statistical Learning Methods | NA | Swedish |
| **(109)** | Information extraction (Named entity recognition) | Ensemble Methods, Deep/Nonlinear Learning Methods, Interpretable Methods, Traditional/Statistical Learning Methods | Orthographic Word Features, Learnt Word-level Embeddings | Spanish, Swedish |
| **(110)** | Information extraction (Medical Concepts, Findings/Symptoms, and Drugs/Adverse events), Resource Development (Corpora and annotation) | Rule-Based Systems | Ontologies | Norwegian |
| **(111)** | Information extraction (Medical Concepts, Findings/Symptoms, and Drugs/Adverse events), Resource Development (Corpora and annotation) | Traditional/Statistical Learning Methods | Statistical Features, Learnt Sentence/Document Embeddings | Norwegian |
| **(112)** | Information extraction (Named entity recognition), De-identification | Traditional/Statistical Learning Methods | Learnt Word-level Embeddings, Statistical Features | Swedish |
| **(113)** | Information extraction (Named entity recognition), De-identification | Traditional/Statistical Learning Methods, Ensemble Methods | Orthographic Word Features | Swedish |
| **(114)** | Context Analysis (Negation detection), Resource development (Lexicons), Content Analysis | Rule-Based Systems | Ontologies | English, French, German, Swedish |
| **(115)** | Resource development (Corpora and annotation), Information Extraction (Named entity recognition, Relations) | Traditional/Statistical Learning Methods, Interpretable Methods | Orthographic Word Features | Norwegian |

***Table 4: Data extracted from the included studies (raw data). The table highlights the most important categories used in the analysis.***

| References | Author's Country | Health Domain | NLP Techniques | Model/Method | Accessibility | Adaption | Data source | Type of EHR documents | Language | Data Availability | Domain expert | Clinical Testing |
|---|---|---|---|---|---|---|---|---|---|---|---|---|
| (3) | Sweden | Infection | Information extraction (Findings/Symptoms, Named entity recognition) | Phrase detection, Iterative merging, Normalized (pointwise) mutual information | NO | NO | Karolinska University Hospital | Clinical notes | Swedish | Yes* | YES | NO |
| (4) | Sweden | N/A | Information extraction (Named entity recognition), De-identification | Rule-based, Edit distance (Levenshtein Distance), Qualitative analysis | NO | NO | Karolinska University Hospital | Clinical notes | Swedish | NO | YES | NO |
| (5) | Sweden | N/A | Context Analysis (Negation detection, Uncertainty/Assertion) | Random indexing, Euclidean distance, Single-linkage agglomerative hierarchical clustering with Euclidean distance | NO | NO | Läkartidningen (Journal of the Swedish Medical Association) corpus | N/A | Swedish | Yes | Yes | NO |
| (6) | Sweden, Finland, Norway, Australia | N/A | Resource development (content analysis) | Qualitative analysis, Quantitative analysis | NO | NO | Finnish university-affiliated hospital, Swedish University-affiliated hospital | Nursing narratives | Swedish, Finnish | NO | NO | NO |
| (7) | Sweden | N/A | Information extraction (Named Entity Recognition) | Rule-based, Cosine similarity, Character-based deep bidirectional recurrent neural network. | Yes | NO | Karolinska University Hospital | Diverse/Not specific | Swedish | Yes | Yes | NO |
| (8) | Sweden | Adverse events | Information extraction (Drugs/Adverse events), Classification (Phenotyping from EHR text) | Random forests, Support vector machines | NO | NO | Karolinska University Hospital | N/A | Swedish | NO | NO | NO |
| (9) | Sweden, Norway | N/A | Information extraction (Named entity recognition), De-identification | Long Short-Term Memory, Conditional Random Fields | NO | NO | Karolinska University Hospital, Läkartidningen (Journal of the Swedish Medical Association) | Clinical notes | Swedish | NO | YES | NO |
| (10) | Sweden | Infection, Oral, Musculoskeletal conditions | Information extraction (Named entity recognition), De-identification | Long Short-Term Memory, Conditional Random Fields | NO | NO | Karolinska University Hospital | N/A | Swedish | Yes* | YES | NO |
| (11) | Sweden | Cardiac disease, Musculoskeletal conditions | Information extraction (Named entity recognition), De-identification | Rule based, Conditional Random Fields, Self-training | NO | NO | Karolinska University Hospital | N/A | Swedish | NO | YES | Yes |
| (12) | Sweden | Infection, Oral, Musculoskeletal conditions | Information extraction (Named entity recognition), De-identification | Rule-based | NO | NO | Karolinska University Hospital | N/A | Swedish | NO | YES | Yes |
| (13) | Sweden | N/A | Information extraction (Named entity recognition), De-identification | Rule-based, Conditional Random Fields | Yes | NO | Karolinska University Hospital | N/A | Swedish | Yes | Yes | Yes |
| (14) | Sweden | Cancer, Adverse events | Information extraction (Named entity recognition), De-identification | Conditional Random Fields | NO | NO | Karolinska University Hospital | Clinical notes | Swedish | NO | YES | NO |
| (15) | Norway | Musculoskeletal conditions | Information extraction (Medical Concepts, Findings/Symptoms), Classification (Phenotyping from EHR text) | Semi-supervised clustering, Rule-based | NO | NO | Sørlandet Hospital Trust | Admission summaries, Discharge summaries, Progress notes, Outpatient clinical notes, Medication prescription records, Radiology reports, Laboratory data reports, Surgery notes, Assessment section | Norwegian | NO | YES | NO |
| (16) | Norway | Musculoskeletal conditions | Information extraction (Medical Concepts, Findings/Symptoms), Classification (Phenotyping from EHR text) | Semi-supervised clustering, Rule-based | NO | NO | Sørlandet Hospital Trust | Admission summaries, Discharge summaries, Progress notes, Outpatient clinical notes, Medication prescription records, Radiology reports, Laboratory data reports, Surgery notes, Assessment section | Norwegian | NO | YES | Yes |
| (17) | Sweden | N/A | Resource development (Corpora and annotation) | Not mentioned | Yes | NO | N/A | Chief complaint, Assessment section, Diagnostic notes | Swedish | Yes | YES | NO |
| (18) | Spain, Sweden, Norway | Gastrointestinal | Classification (Indexing and coding) | Convolutional Neural Networks, BERT, PlaBERT (BERT with per-label attention) | Yes | Yes | Karolinska University Hospital | Discharge Summaries | Spanish, Swedish, English | Yes* | NO | Yes |
| (19) | Spain, Sweden, Norway | Gastrointestinal | Classification (Indexing and coding) | Bidirectional Encoder Representations from Transformers (BERT) | Yes | Yes | Karolinska University Hospital | Discharge Summaries | Spanish, Swedish | N/A | NO | NO |
| (20) | Sweden | Musculoskeletal conditions, Infection, Oral, Cardiac disease | Information extraction (Named entity recognition), De-identification | SweDeClin-BERT | Yes | NO | Karolinska University Hospital, Linköping University Hospital | N/A | Swedish | N/A | Yes | NO |
| (21) | Sweden | Cardiac disease, Family history | Resource development (Corpora and annotation), Information Extraction (Named entity recognition), De-identification | Conditional Random Fields, Rule-based | Yes | Yes | NorSynthClinical PHI | Family History Notes | Norwegian | Yes | Yes | NO |
| (22) | Norway, Sweden | Gastrointestinal | Classification (Indexing and coding), Context Analysis (Negation detection) | KB/BERT, Support vector machines, Rule-based | NO | NO | Karolinska University Hospital | Discharge Summaries | Swedish | Yes* | NO | NO |
| (23) | Sweden | Patients' smoking status | Information extraction (Specific characteristics, Named entity recognition), Classification (Phenotyping from EHR text) | Sequential Minimal Optimization, K-NN, Naïve Bayes, Decision tree, Rule-based | NO | NO | N/A | N/A | Swedish | N/A | Yes | NO |
| (24) | Norway, Sweden | Gastrointestinal | Classification (Indexing and coding) | Swedish KB BERT, Fuzzy string matching, Levenshtein distance | Yes | NO | Karolinska University Hospital | Discharge Summaries | Swedish | Yes* | NO | NO |
| (27) | Norway, Sweden | N/A | Information extraction (Named entity recognition), De-identification | Bidirectional Long Short-Term Memory with conditional random fields | NO | Yes | Karolinska University Hospital, Läkartidningen (The Swedish scientific medical journal), Wikipedia | Clinical notes | Swedish | Yes* | Yes | NO |
| (28) | Sweden | Cancer | Classification (Phenotyping from EHR text) | Rule-based | NO | NO | The Cancer Registry of Norway | Pathology reports | Norwegian | N/A | Yes | NO |
| (29) | Norway, USA | N/A | Classification (Phenotyping from EHR text) | Support vector machines, Bidirectional long Short-Term Memory, Convolutional neural network | NO | NO | Akershus University Hospital | Radiology reports | Norwegian | NO | YES | NO |
| (25) | Norway, Sweden | Gastrointestinal | Classification (Indexing and coding) | SweDeClin-BERT, LIME, SHAP | NO | NO | Karolinska University Hospital | Discharge Summaries | Swedish | N/A | YES | NO |
| (30) | Denmark | N/A | Resource development (Corpora and annotation), Information Extraction (Named entity recognition), De-identification | Rule-based, Princeton University Relation Extraction system (PURE) NER, Danish BERT | Yes | Yes | Odense University Hospital | Narrative Clinical Text | Danish | NO | YES | NO |
| (31) | Denmark | Bleeding, Cardiac disease, Thorat, Musculoskeletal conditions, Gastrointestinal | Resource development (Corpora and annotation) | Bidirectional gated recurrent neural network | Yes | Yes | Odense University Hospital | Clinical notes | Danish | NO | YES | NO |
| (32) | Norway | Allergies, Musculoskeletal conditions | Information extraction (Medical Concept, and Findings/Symptoms), classification (Phenotyping from EHR text) | CNN 1D neural network, Naïve Bayes, Decision tree, Logistic regression, Random Forest, K-NN, Support Vector Machines, Multi-layer perceptron, Long Short-Term Memory, Long Short-Term Memory CNN, Bidirectional Long Short-Term Memory, Bidirectional Long Short-Term Memory CNN, Bidirectional Long Short-Term Memory with Attention Mechanism, Gated recurrent unit, Rule-based | NO | NO | Sørlandet Hospital Trust | Clinical notes | Norwegian | NO | Yes | Yes |
| (33) | Denmark | N/A | Resource development (Corpora and annotation), Information Extraction (Named entity recognition, and Relations) | Princeton University Relation Extraction system (PURE) NER, Danish Clinical ELECTRA | NO | Yes | Odense University Hospital | Clinical notes, Nursing notes, Ambulatory care contacts, Surgical notes | Danish | NO | YES | NO |
| (34) | Sweden | Gastrointestinal, Adverse events | Information extraction (Named entity recognition), Classification, De-identification | SweClin-BERT, SweDeClin-BERT | NO | Yes | Karolinska University Hospital | Discharge summaries | Swedish | NO | YES | NO |
| (35) | Denmark | Bleeding, Venous thromboembolism | Classification (Phenotyping from EHR text) | Danish clinical ELECTRA (Clin-ELECTRA), Long Short-Term Memory (LSTM) | NO | Yes | Odense University Hospital | EHR Narrative | Danish | NO | YES | NO |
| (36) | Denmark | Bleeding, Venous thromboembolism | Resource development (Models), Classification (Phenotyping from EHR text) | BERT, MeDa-BERT, LSTM+MeDa-WE, LSTM+General-WE | Yes | Yes | Odense University Hospital | EHR notes, Clinical guidelines | Danish | N/A | YES | NO |
| (37) | Norway | Sepsis, Adverse events | Resource development (Corpora and annotation, and methods) | Semantic annotation guideline design process | NO | NO | Norwegian Hospital, Synthetic adverse event notes | Adverse event reports | Norwegian | NO | YES | NO |
| **(38)** | Sweden | Infection | Classification (Phenotyping from EHR text) | Support vector machines, Gradient tree boosting | NO | NO | Karolinska University Hospital | Physician's notes, Microbiological results | Swedish | NO | NO | NO |
| **(39)** | Denmark | Psychiatric disorder | Resource development (Machine-readable dictionaries), Information Extraction (Named Entity recognition, Drugs/Adverse events) | Rule-based | NO | NO | Danish psychiatric hospital | Clinical narrative | Danish | NO | Yes | NO |
| **(40)** | Sweden | Adverse events | Information Extraction (Named Entity recognition, Drugs/Adverse events, Relations) | Decision tree, Random Forest, Support vector machines | NO | NO | Karolinska University Hospital | N/A | Swedish | NO | NO | NO |
| **(41)** | Sweden | N/A | Information extraction (Named Entity Recognition) | Balanced undersampling, Oversampling, Conditional Random Fields, Bidirectional Long Short-Term Memory neural network with conditional random fields | NO | NO | Karolinska University Hospital | N/A | Swedish | NO | NO | NO |
| **(42)** | Sweden | N/A | Information extraction (Named Entity Recognition) | BERT | NO | NO | Karolinska University Hospital | N/A | Swedish | NO | NO | NO |
| **(43)** | Sweden | N/A | Resource development (Lexicons, methods) | Rule-based, Swedish Clinical Abbreviation and Medical Terminology Matcher (SCATM), Swedish statistical language model-based compound analysis (NoCM), and misspelling resolution (SCATM+NoCAM) | NO | NO | Karolinska University Hospital | Physician's notes | Swedish | NO | NO | NO |
| **(44)** | Sweden | Cardiac disease, Infection | Resource development (Content analysis) | Rule-based, Quantitative analysis | NO | NO | Karolinska University Hospital | Clinical narrative, Dietician notes, Nurse notes, Physician notes | Swedish | NO | Yes | NO |
| **(45)** | Sweden | N/A | Core NLP (Parsing, Part of Speech tagging) | Rule+based, SweMalt (MaltParser), Granska tagger (PoS) | Yes | NO | N/A | Assessment entries | Swedish | NO | NO | NO |
| **(46)** | Sweden | Oral, Musculoskeletal conditions | Information Extraction (Named Entity Recognition) | Conditional Random Fields | NO | NO | Karolinska University Hospital | Clinical notes | Swedish | NO | Yes | NO |
| **(47)** | Sweden | Adverse events | Information Extraction (Named Entity Recognition, Drugs/Adverse events) | Random forest, Shallow neural networks (skip-gram model) | NO | NO | Karolinska University Hospital | N/A | Swedish | NO | NO | NO |
| **(48)** | Sweden | Infection, Oral, Musculoskeletal conditions | Information extraction (Named entity recognition), De-identification | Random indexing, Conditional Random Fields, Cosine distance | NO | NO | Karolinska University Hospital | Clinical notes | Swedish | NO | NO | NO |
| **(49)** | Sweden | N/A | Classification (Indexing and coding), Context Analysis (negation detection) | Random indexing, Rule-based | NO | NO | Karolinska University Hospital | Clinical notes | Swedish | NO | NO | NO |
| **(50)** | Sweden | N/A | Classification (Indexing and coding) | Random indexing, Manual dimension optimization | NO | NO | Karolinska University Hospital | N/A | Swedish | NO | NO | NO |
| **(51)** | Sweden | N/A | Classification (Indexing and coding) | Random indexing, Qualitative analysis | NO | NO | Karolinska University Hospital | N/A | Swedish | NO | NO | NO |
| **(52)** | Sweden | Infection, Oral, Musculoskeletal conditions | Information extraction (Named entity recognition), De-identification | Conditional Random Fields | NO | NO | Karolinska University Hospital | Nurse notes, Clinical notes | Swedish | NO | Yes | NO |
| **(53)** | Sweden | Adverse events, Allergies | Information extraction (Drugs/Adverse events, Named entity recognition, Relations), Context Analysis (Negation detection, Uncertainty/Assertion, Temporality), Resource Development (Corpora and annotation) | Conditional Random Fields | NO | NO | Karolinska University Hospital | Physician notes, Admission Notes, Patient History, Hypersensitivity and Drug Info, Assessment, Discharge Note | Swedish | NO | Yes | NO |
| **(54)** | Sweden | N/A | Information Extraction (Medical Concepts, Relations), Context Analysis (Abbreviation), Resource Development (methods) | Random indexing, Rrandom permutation | NO | NO | Karolinska University Hospital | EHR narrative, Clinical notes | Swedish | Yes* | NO | NO |
| **(55)** | Sweden | N/A | Resource Development (methods), Information Extraction (Medical concept, Relations) | Random indexing, Random permutation | NO | NO | Karolinska University Hospital | Clinical notes | Swedish | NO | NO | NO |
| **(56)** | Sweden | Adverse events | Resource Development (methods), Information Extraction (Drugs/Adverse events) | Random forest | NO | NO | Karolinska University Hospital | N/A | Swedish | NO | NO | NO |
| **(57)** | Sweden | Adverse events | Information extraction (Drugs/Adverse events) | Artificial neural network, Random Forest | NO | NO | Karolinska University Hospital | N/A | Swedish | NO | NO | NO |
| **(58)** | Sweden | Adverse events | Resource Development (methods), Information Extraction (Drugs/Adverse events) | Artificial neural network, Random Forest | NO | NO | Karolinska University Hospital | Clinical notes | Swedish | NO | NO | NO |
| **(59)** | Sweden | Infection | Classification (Phenotyping from EHR text) | Sparse autoencoders, Restricted Boltzmann machines | NO | NO | Karolinska University Hospital | N/A | Swedish | NO | Yes | NO |
| **(60)** | Norway | Cancer, Gastrointestinal | Classification (Phenotyping from EHR text) | Smith-Waterman matching algorithm, Naive Bayes classifiers | NO | NO | University Hospital of North Norway, Norwegian Death Registry | Admission journal, Nurse notes, Doctor notes, Descriptive surgical reports, Intensive care reports, Discharge notes | Norwegian | NO | NO | NO |
| **(61)** | Sweden | Cardiac disease | Information extraction (Medical Concepts) | KDTree, Generalist Swedish pre-trained KB-BERT, Swedish pre-trained SweDeClin-BERT | NO | NO | Linköping University Hospital | N/A | Swedish | NO | Yes | NO |
| **(62)** | Sweden | Skin disorder, Aplastic anemia, Adverse events | Information extraction (Drugs/Adverse events) | Random forest, Random indexing | NO | NO | Karolinska University Hospital | N/A | Swedish | NO | NO | NO |
| **(63)** | Sweden | N/A | Context analysis (abbreviation), Classification | Swedish Clinical Abbreviation Normalizer (SCAN), Rule-based, Heuristics-based, Lexicon-based | NO | NO | Karolinska University Hospital | Radiology reports, Assessment entries | Swedish | NO | NO | NO |
| **(64)** | Sweden | N/A | Information extraction (Named entity recognition), De-identification, Classification (indexing and coding), Context analysis (uncertainty/assertion classification) | Clinical KB-BERT model | NO | NO | Karolinska University Hospital | Clinical notes, Discharge summaries | Swedish | Yes | NO | NO |
| **(65)** | Sweden | N/A | Information extraction (Named entity recognition), De-identification, Classification (indexing and coding) | Clinical KB-BERT model with modified vocabulary | NO | NO | Karolinska University Hospital | Clinical notes, Discharge summaries | Swedish | NO | NO | NO |
| **(66)** | Sweden | Gastrointestinal | Classification (Indexing and coding), De-identification, Information Extraction (Named entity recognition, Negation detection, Findings/Symptoms, Drugs/Adverse events), Context Analysis (Negation detection) | Clinical BERT models for Swedish | NO | Yes | Karolinska University Hospital | Clinical notes, Discharge summaries, ICD-10 codes | Swedish | Yes* | NO | NO |
| **(67)** | Sweden | Nutrition | Resource Development (content analysis), Context Analysis (Abbreviation) | Frequency analysis, Context analysis | NO | NO | Karolinska University Hospital | Clinical notes, Radiology reports, Dietetic notes | Swedish | Yes* | Yes | NO |
| **(68)** | Norway, Spain | Postoperative delirium | Information extraction (Findings/Symptoms, Medical Concepts, Specific characteristics), Classification (Phenotyping from EHR text) | k-nearest neighbors (kNN) clustering, Elastic net logistic regression | NO | NO | University Hospital of North Norway | Surgical operation notes, Radiology report, Semi-structured nurses' notes, ICD-10 codes | Norwegian | NO | Yes | NO |
| **(69)** | USA, Sweden | Infection | Resource development (Corpora and annotation, methods, content analysis), Context analysis (Negation detection, Uncertainty/Assertion) | Qualitative comparison study, Grounded theory | NO | NO | N/A | Emergency department reports, Assessment entries | English, Swedish | Yes | NO | NO |
| **(70)** | Sweden | Infection, Oral, Musculoskeletal conditions | Resource development (Lexicons), Core NLP (Part of Speech tagging, Parsing) | Lexical lookup, Exact matching technique | NO | NO | Karolinska University Hospital | Physician notes, Nurse narratives | Swedish | Yes* | NO | NO |
| **(71)** | Sweden | Infection | Classification (Phenotyping from EHR text) | Clinical KB-BERT | NO | NO | Karolinska University Hospital | Clinical notes | Swedish | NO | NO | NO |
| **(72)** | Sweden | Infection | Classification (Phenotyping from EHR text) | Clinical KB-BERT | NO | NO | Hospitals in Stockholm | Clinical notes, Emergency department notes, Radiology notes | Swedish | NO | NO | NO |
| **(73)** | Denmark | Infection | Classification (Phenotyping from EHR text) | Universal Language Model Fine-Tuning, Average-Stochastic Gradient Descent Weight-Dropped Long Short-Term Memory | NO | NO | Hospitals in the Capital Region of Denmark | Postoperative chart notes, Admission notes, Operative notes, Progress notes, Radiology reports, Discharge notes | Danish | Yes* | Yes | NO |
| **(74)** | Sweden | Adverse events | Classification (Phenotyping from EHR text), Information extraction (Drugs/Adverse events) | Swedish BERT, AER-BERT, Long Short-Term Memory neural network, XGBoost | NO | NO | Swedish Medical Products Agency | Adverse event reports | Swedish | Yes* | Yes | NO |
| **(75)** | Norway | N/A | Resource development (Corpora and annotation), Information extraction (Named entity recognition), De-identification | GPT-4, Conditional random fields, Rule-based | Yes | Yes | NorSynthClinical-PHI, Synthetic test | Clinical notes | Norwegian | Yes | Yes | NO |
| **(26)** | Norway, Sweden | Gastrointestinal | Classification (Indexing and coding) | Clinical KB-BERT | NO | NO | Karolinska University Hospital | Discharge summaries | Swedish | Yes* | Yes | NO |
| **(76)** | Norway, Italy, Sweden | Gastrointestinal | Information extraction (Named entity recognition), Classification, De-identification | SweDeClin-BERT, NorBERT, ScandiBERT | NO | Yes | University Hospital of North Norway, Karolinska University Hospital, Statistisk sentralbyrå, Danish Dependency Treebank, NorSynthClinical PHI Corpus, | Gastro surgery notes | Norwegian, Danish, Swedish | NO* | Yes | NO |
| **(77)** | Denmark | Bleeding | Classification (Phenotyping from EHR text), Information Extraction (Findings/Symptoms) | Recurrent neural networks (RNNs), Convolutional neural networks (CNNs), Rule-based | NO | Yes | Hospitals in the Region of Southern Denmark | EHR notes, ICD codes | Danish | NO | Yes | NO |
| **(78)** | Denmark, Spain | Bleeding | Classification (Phenotyping from EHR text), Information Extraction (Findings/Symptoms), Resource development (Corpora and annotation) | Clinical ELECTRA, General BERT, Bidirectional Long Short-Term Memory | NO | Yes | Odense University Hospital | EHR notes | Danish | NO | Yes | NO |
| **(79)** | Sweden, Spain | Cancer | Information extraction (Named entity recognition, Classification (Phenotyping from EHR text) | Maximum probability tagger, Conditional random fields, Perceptron, Support vector machine, Brown trees, K-means clustering | Yes | Yes | Karolinska University Hospital | Clinical texts | Spanish, Swedish | Yes* | Yes | NO |
| **(80)** | Norway | N/A | Resource development (Lexicons, methods), Information extraction (Named entity recognition) | Suffix-based mapping, Keyword-based mapping, Manual evaluations | NO | NO | Medisinsk ordbok, FEST database for information, Synthetic notes | Family history notes, Clinical notes | Norwegian | Yes | Yes | NO |
| **(81)** | Norway, Spain | Cardiac disease | Resource development (Corpora and annotation, methods)), Information extraction (Named entity recognition, Relations), Classification (Phenotyping from EHR text) | Support vector machine | NO | NO | Synthetic corpus of clinical text | N/A | Norwegian | Yes | Yes | NO |
| **(82)** | Sweden | Gastrointestinal | Classification (Indexing and coding) | KB-BERT, Support Vector Machines, Decision Trees, K-nearest Neighbours | NO | Yes | Karolinska University Hospital | Clinical notes, Discharge Summaries | Swedish | Yes* | NO | NO |
| **(83)** | Denmark | Psychiatric disorder | Information extraction (Relations), Classification (Cohort stratification, Indexing and coding) | Hierarchical clustering, Cosine similarity | NO | NO | Sct. Hans Hospital | Epicrisis, Discharge note, Treatment note, Nursing note | Danish | NO | NO | NO |
| **(84)** | Norway | Infection | Classification (Phenotyping from EHR text), Information Extraction (Findings/Symptoms) | Support Vector Machines | Yes | NO | Akershus University Hospital | Clinical notes, Somatic nurse notes, Somatic physician notes, Discharge notes, Admission notes, Intensive nurse note, Ward transfer note | Norwegian | Yes | Yes | NO |
| **(85)** | Norway | Infection | Classification (Phenotyping from EHR text), Information Extraction (Findings/Symptoms) | Support Vector Machines | NO | NO | Akershus University Hospital | Clinical notes, Nursing notes, Surgical notes, Physician notes, Laboratory examination notes, Discharge summaries, Admission notes, Transfer notes, Reception notes, Palliative notes | Norwegian | NO | Yes | NO |
| **(86)** | Sweden | N/A | Context Analysis (Negation detection), Classification (Medical Concepts) | Rule-based, Negex | NO | Yes | Karolinska University Hospital | Assessment field | Swedish | Yes* | Yes | NO |
| **(87)** | Sweden | N/A | Information extraction (Named entity recognition), resource development (Corpora and annotation) | Conditional random fields | NO | NO | Karolinska University Hospital | Emergency notes | Swedish | NO | Yes | NO |
| **(88)** | Sweden | N/A | Core NLP (Parsing) | Malt-Parser, Dependency Parsing, Multilingual | NO | Yes | Karolinska University Hospital, Lakartidningen (journal from the Swedish Medical Association) | Assessment section | Swedish | NO | NO | NO |
| **(89)** | Sweden | Cardiac disease, Musculoskeletal conditions | Information extraction (Named entity recognition) | Conditional random fields, Ran-dom indexing, Clustering | NO | NO | Karolinska University Hospital | Emergency notes | Swedish | NO | Yes | NO |
| **(90)** | Sweden | N/A | Resource development (Lexicons) | Random indexing, Term Replacement, Cosine Addition | NO | NO | Karolinska University Hospital, Lakartidningen (journal from the Swedish Medical Association) | N/A | Swedish | Yes | NO | NO |
| **(91)** | Sweden | N/A | Information extraction (Named entity recognition), Resource Development (Lexicons), Classification (Indexing and coding) | Rule-based | NO | NO | Karolinska University Hospital | Assessment section | Swedish | Yes* | Yes | NO |
| **(92)** | Sweden | N/A | Information extraction (Named entity recognition), Classification (Phenotyping from EHR text) | Conditional random fields | NO | NO | Karolinska University Hospital | Clinical notes, Assessment section | Swedish | Yes* | Yes | NO |
| **(93)** | Norway, Spain | Cancer, Gastrointestinal | Classification (Phenotyping from EHR text) | Support vector machines, Naive Baye, Fisher discriminant analysis | NO | NO | University Hospital of North Norway | Nurses notes, Journal notes, Outpatient notes, Radiology reports, Referrals, Discharge letters, Admission notes | Norwegian | Yes | NO | NO |
| **(95)** | Sweden | N/A | Resource Development (methods), Context Analysis (Negation detection) | Rule-based, NegEx, PyConTextNLP, SynNeg | Yes | NO | Karolinska University Hospital | Assessment section | Swedish | Yes* | Yes | NO |
| **(96)** | Sweden | N/A | Resource development (Lexicons), Context analysis (Abbreviation) | Rule-based, Random Indexing, Random Permutation, Cosine similarity | NO | NO | Karolinska University Hospital, Lakartidningen (Journal of the Swedish Medical Association) | Radiology reports | Swedish | Yes | NO | NO |
| **(97)** | Denmark | Psychiatric disorder | Information extraction (Findings/Symptoms, Medical Concepts), Context Analysis (Negation detection) | Rule-based | NO | NO | Sct. Hans Hospital | Doctors notes, Psychiatrists notes, Nurses notes, Social workers notes | Danish | NO | Yes | NO |
| **(98)** | Sweden | Adverse events, Gastrointestinal | Classification (Phenotyping from EHR text), Information Extraction (Drugs/Adverse events, Relations, Named entity recognition) | BERT, KB-BERT | NO | Yes | Karolinska University Hospital | Discharge summaries | Swedish | Yes* | Yes | NO |
| **(99)** | Sweden | Cardiac disease, Infection, Musculoskeletal conditions | Resource development (Corpora and annotation) and Context Analysis (Negation detection, Uncertainty/Assertion) | Descriptive statistics, Pairwise Inter-Annotator Agreement | NO | NO | Karolinska University Hospital | Assessment section | Swedish | Yes* | Yes | NO |
| **(100)** | Sweden | N/A | Context Analysis (Temporality) | Rule-based, Manual evaluation | Yes, | Yes | Karolinska University Hospital | Nurse's notes, Doctors notes | Swedish | Yes | Yes | NO |
| **(101)** | Sweden | N/A | Classification (Phenotyping from EHR text, Indexing and coding), information extraction (Drugs/Adverse events) | Conditional Random Fields | NO | NO | Karolinska University Hospital | Assessment section | Swedish | Yes | Yes | NO |
| **(102)** | Sweden, USA | N/A | Resource development (Lexicons), Context Analysis (Uncertainty/Assertion) | Rule-based, pyConTextNLP, pyConTextSWE | NO | Yes | Karolinska University Hospital | Diagnostic notes | Swedish | NO | Yes | NO |
| **(103)** | Sweden, Finland, Norway | N/A | Resource development (Corpora and annotation), Context Analysis (Temporality) | Rule-based | NO | NO | Karolinska University Hospital | Nurse's notes, Doctors notes | Swedish, Finnish | Yes | Yes | NO |
| **(104)** | Sweden | N/A | Resource development (Corpora and annotation, methods), Context Analysis (Uncertainty/Assertion, Negation detection) | Conditional Random Fields, Rule-based, | NO | NO | Karolinska University Hospital | Radiology reports, Assessment section | Swedish | NO | NO | NO |
| **(105)** | Sweden | Cancer | Information extraction (Findings/Symptoms) | Rule-based | NO | NO | The Cancer Registry of Norway | Pathology reports | Norwegian | NO | Yes | NO |
| **(106)** | Sweden, Denmark | Cancer | Information extraction (Findings/Symptoms), Context Analysis (Negation detection) | Rule-based | Yes | NO | Karolinska University Hospital | Nurse's notes, Doctors notes | Swedish | NO | Yes | NO |
| **(107)** | Sweden, Norway | Cancer | Information extraction (Findings/Symptoms, Named entity recognition, Specific characteristics) | Rule-based, Levenshtein distance similarity | NO | NO | The Cancer Registry of Norway | Pathology reports | Norwegian | NO | Yes | NO |
| **(108)** | Sweden | N/A | Resource Development (Corpora and annotation, methods), Context Analysis (Uncertainty/Assertion) | Qualitative analysis | NO | NO | Karolinska University Hospital | Assessment section | Swedish | NO | Yes | NO |
| **(109)** | Sweden, Spain | N/A | Information extraction (Named entity recognition) | Ensemble, Perceptrons, Support vector machines, Conditional Random Fields | NO | NO | Karolinska University Hospital | N/A | Spanish, Swedish | NO | Yes | NO |
| **(110)** | Norway | Adverse events, Sepsis | Information extraction (Medical Concepts, Findings/Symptoms, and Drugs/Adverse events), Resource Development (Corpora and annotation) | Annotation guideline | NO | NO | St. Olavs University Hospital | Adverse event reports | Norwegian | NO | Yes | NO |
| **(111)** | Norway | Infection | Information extraction (Medical Concepts, Findings/Symptoms, and Drugs/Adverse events), Resource Development (Corpora and annotation) | Latent Dirichlet Allocation, Bayesian optimization | Yes | NO | St. Olavs University Hospital | Adverse event reports | Norwegian | NO | Yes | NO |
| **(112)** | Sweden | N/A | Information extraction (Named entity recognition), De-identification | Conditional Random Fields | NO | NO | Karolinska University Hospital | Clinical notes | Swedish | NO | Yes | NO |
| **(113)** | Sweden | Oral, Musculoskeletal conditions | Information extraction (Named entity recognition), De-identification | Conditional Random Fields, Random forests | NO | NO | Karolinska University Hospital | Clinical notes | Swedish | NO | No | NO |
| **(114)** | USA, Sweden, France | N/A | Context Analysis (Negation detection), Resource development (Lexicons), Content Analysis | Rule-based, NegEx, ConText | Yes | NO | Karolinska University Hospital | Assessment entries | English, French, German, Swedish | NO | Yes | NO |
| **(115)** | Norway, USA | Cardiac disease | Resource development (Corpora and annotation), Information Extraction (Named entity recognition, Relations) | Conditional Random Fields, Support vector machines | Yes | NO | Oslo University Hospital | Outpatient notes | Norwegian | Yes | Yes | NO |
| **(94)** | Spain, Norway, USA | Adverse events | Classification (Phenotyping from EHR text) | Support vector machines, Kernel methods | NO | NO | University Hospital of North Norway | Nurse notes, Journal notes, Outpatient notes, Radiology reports, Referrals, Discharge letters, Admission notes | Norwegian | NO | Yes | NO |

- ***Natural Language Processing Techniques - Methods/Model: Country-Specific Developments***

This section presents the country-specific developments and adoption of models for their respective clinical texts. Generally, the three countries had started publishing at different timeframes and had varied rates of adoption of various methods. A more detailed analysis of the three languages in terms of the adoption of different kinds of NLP models or methods is given below.

*<u>Norwegian</u>*: Interpretable methods (32.0%, n = 16) are the most widely used methods for Norwegian clinical text, followed by rule-based systems (24.0%, n = 12), and deep/nonlinear learning methods (20.0%, n = 10). The fourth most popular method is traditional/statistical learning methods (14.0%, n = 7), followed by transformers and large language models (8.0%, n = 4), and ensemble methods (2.0%, n = 1). Despite the recent success, studies focusing on Norwegian clinical text seem to start late in terms of adopting transformers and large language models, which only appeared in 2024, and deep/nonlinear learning methods also became visible only in 2021. Studies focusing on Norwegian clinical text started to appear in publications in 2014 with interpretable methods and traditional/statistical learning methods.

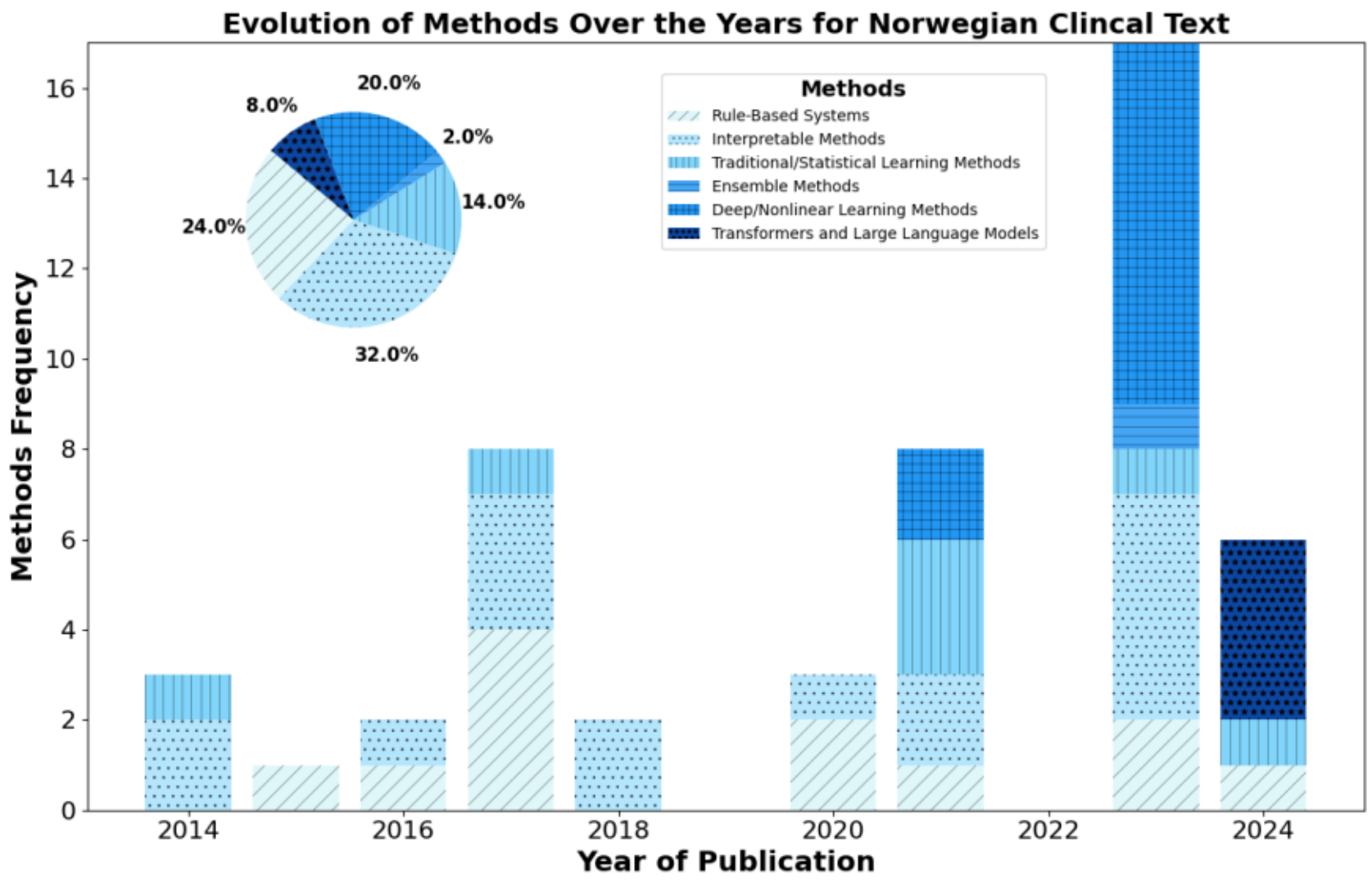

*Figure 1: Evolution of NLP model/methods for Norwegian clinical text.*

*Swedish*: Traditional/statistical learning methods (39.2%, n = 69) are the most widely used methods for Swedish clinical text, followed by rule-based systems (24.4%, n = 43). The third most used method is transformers and large language models (14.8%, n = 26) followed by interpretable methods (8.5%, n = 15). Deep/nonlinear learning methods (6.8%, n = 12), and ensemble methods (6.3%, 11) rank equally as the fifth most used method. Studies focusing on Swedish clinical text are leaders in terms of early adoption of transformers and large language models, and deep/nonlinear learning methods, which appeared in 2021, and 2015 respectively. Studies focusing on Swedish clinical text started appearing in publications in 2010 with rule-based systems and traditional/statistical learning methods.

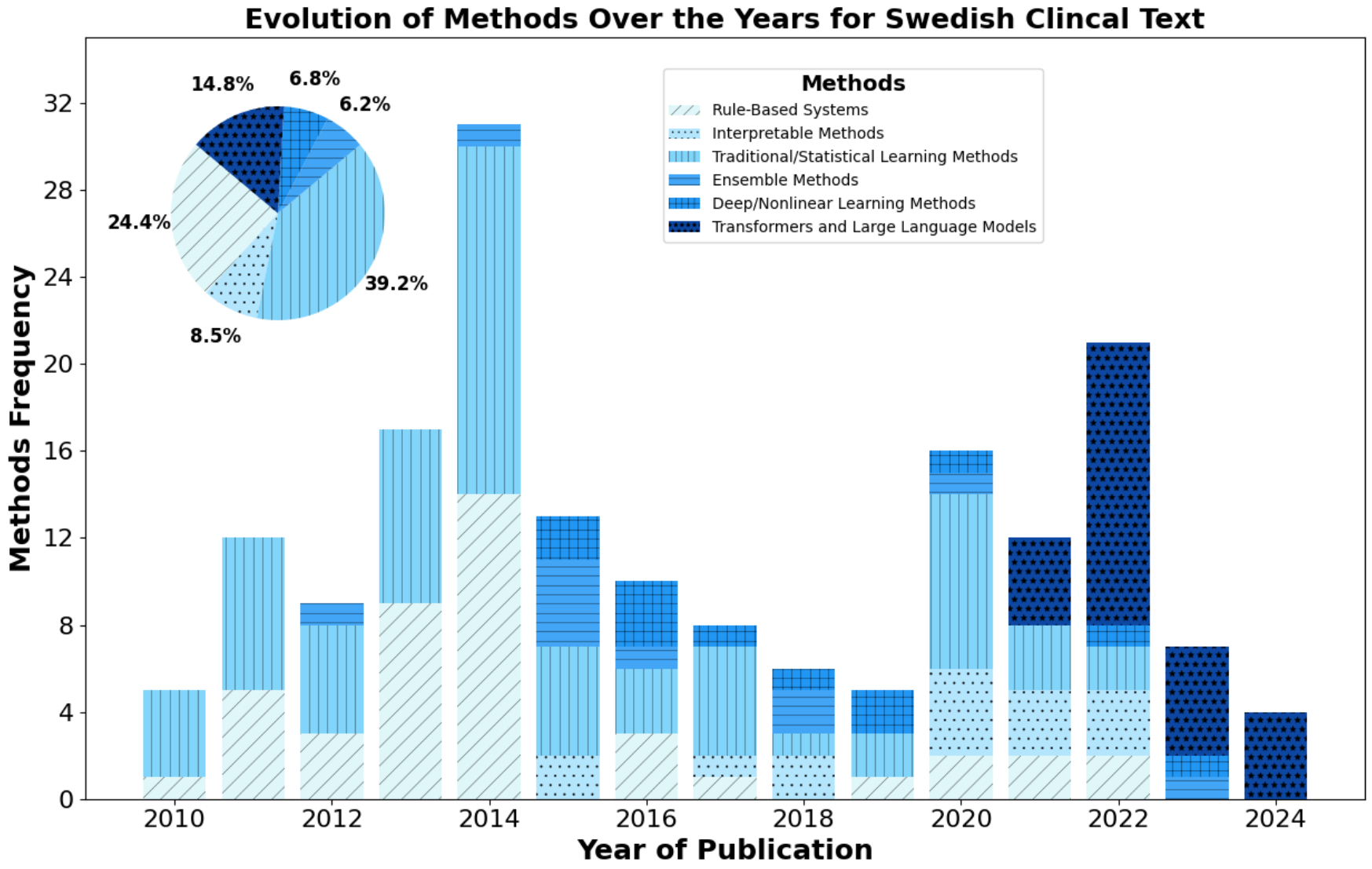

*Figure 2: Evolution of NLP models/methods for Swedish clinical text.*

*Danish*: Transformers and large language (40.7%, n = 11) models are the most widely used method for Danish clinical text, followed by deep/nonlinear learning methods (22.2%, n = 6). Rule-based systems (14.8%, n = 4), and hybrid methods (14.8%, n = 4) are ranked equally as the third most popular methods. It is worth mentioning that hybrid methods involving combining two or more methods only appeared in studies focusing on Danish clinical text in 2023. Traditional/Statistical Learning Methods have a limited appearance (7.4%, n = 2). Despite being a late starter compared to studies focusing on Swedish clinical text in adopting transformers and large language, which appeared in 2022, studies focusing on Danish clinical text has shown a significant increase afterward. Studies focusing on Danish clinical text started to appear in publication in 2011 with traditional/statistical learning methods.

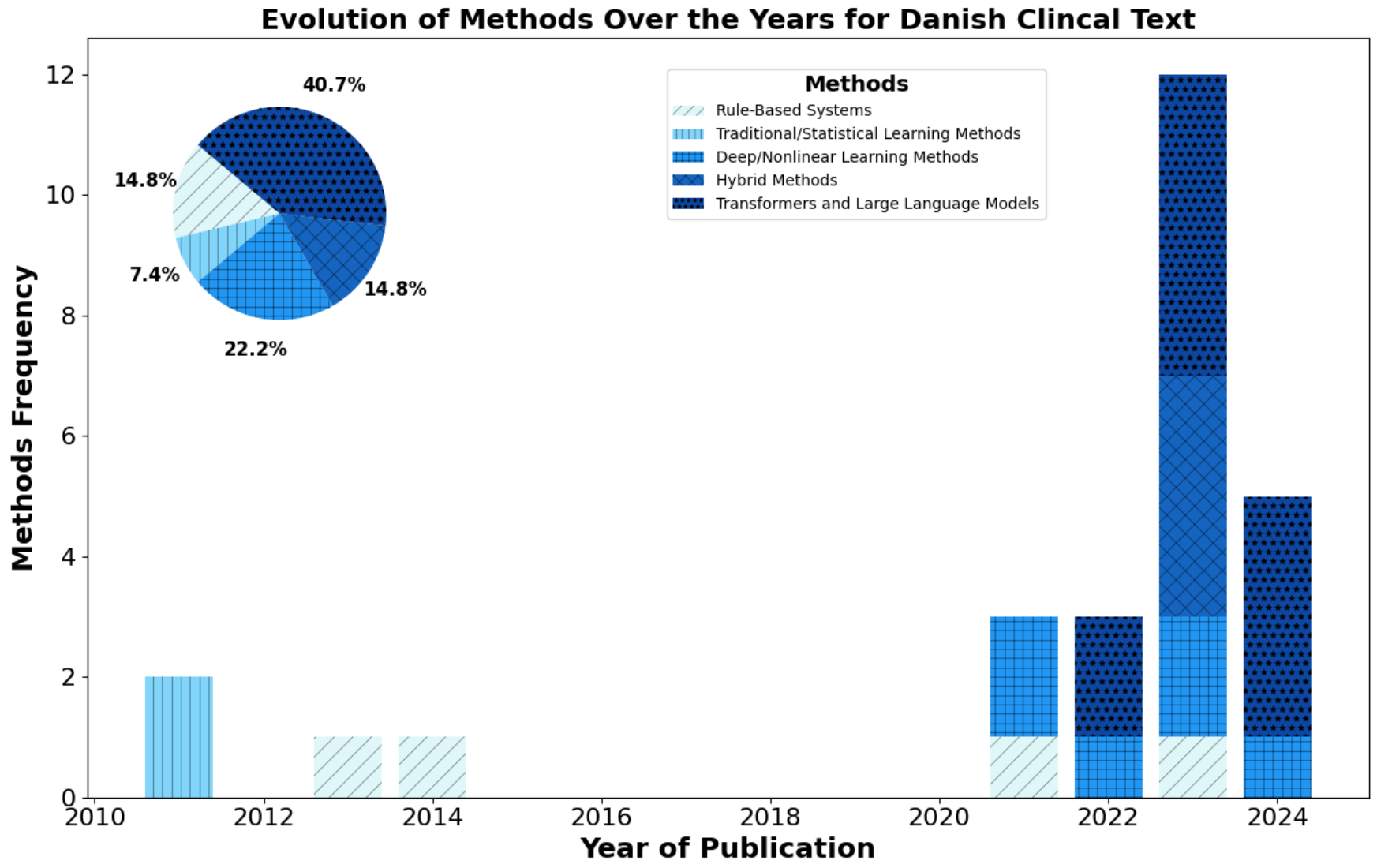

*Figure 3: Evolution of NLP models/methods for Danish clinical text.*

- ***Natural Language Processing Tasks Trend: Country-Specific Developments***

This section presents the type of NLP tasks considered across the three languages, as shown in Figures 4, 5, and 6. A detailed analysis providing highlights of each NLP task preference for each language is given below.

*Norwegian*: Information extraction (41.6%, n = 16) represents a significant portion of all the NLP tasks considered for Norwegian clinical text, followed by classification (30.8%, n = 12) task, and resource development (20.5%, n = 8). Despite the significant challenges associated with handling health data due to privacy and security reasons, a very limited number of studies focusing on Norwegian clinical text have attempted to tackle the task of de-identification (7.7%, n = 3). No research has been dedicated to context analysis and core NLP tasks for Norwegian clinical text.

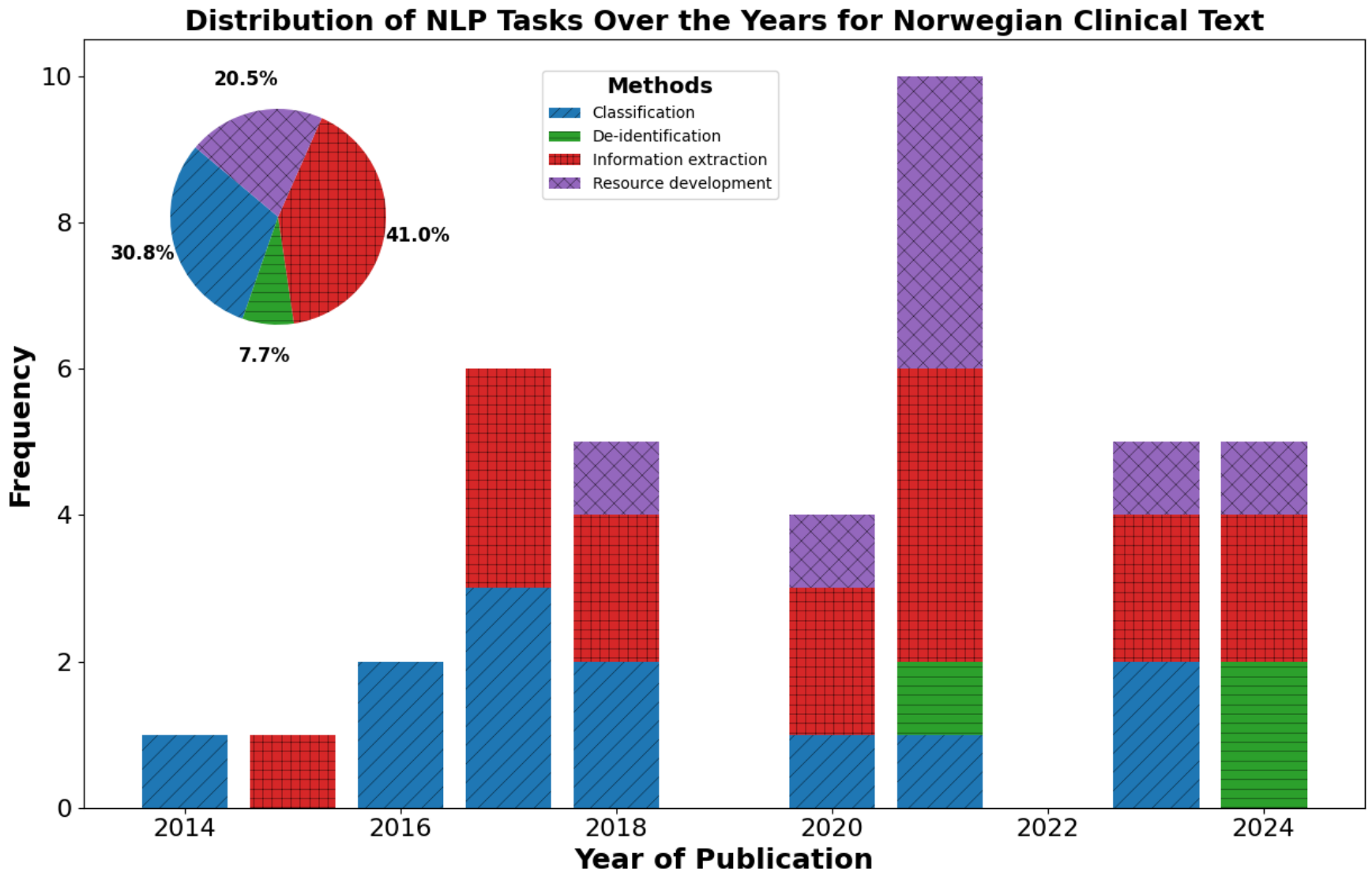

*Figure 4: Distribution of NLP tasks for Norwegian clinical text.*

*Swedish*: Like the Norwegian, information extraction (33.3%, n = 45) constitutes a notable portion of all the NLP tasks considered for Swedish clinical text, followed by classification (18.5%, n = 25) task. Context analysis (15.6%, n = 21) and resource development (17.0%, n = 23) ranked third equally representing a fair share of the studied NLP tasks. Compared to the studies focusing on Norwegian and Danish clinical text, de-identification (13.3%, n = 18) got significant attention, and there are also some research activities focusing on core NLP (2.2%, n = 3).

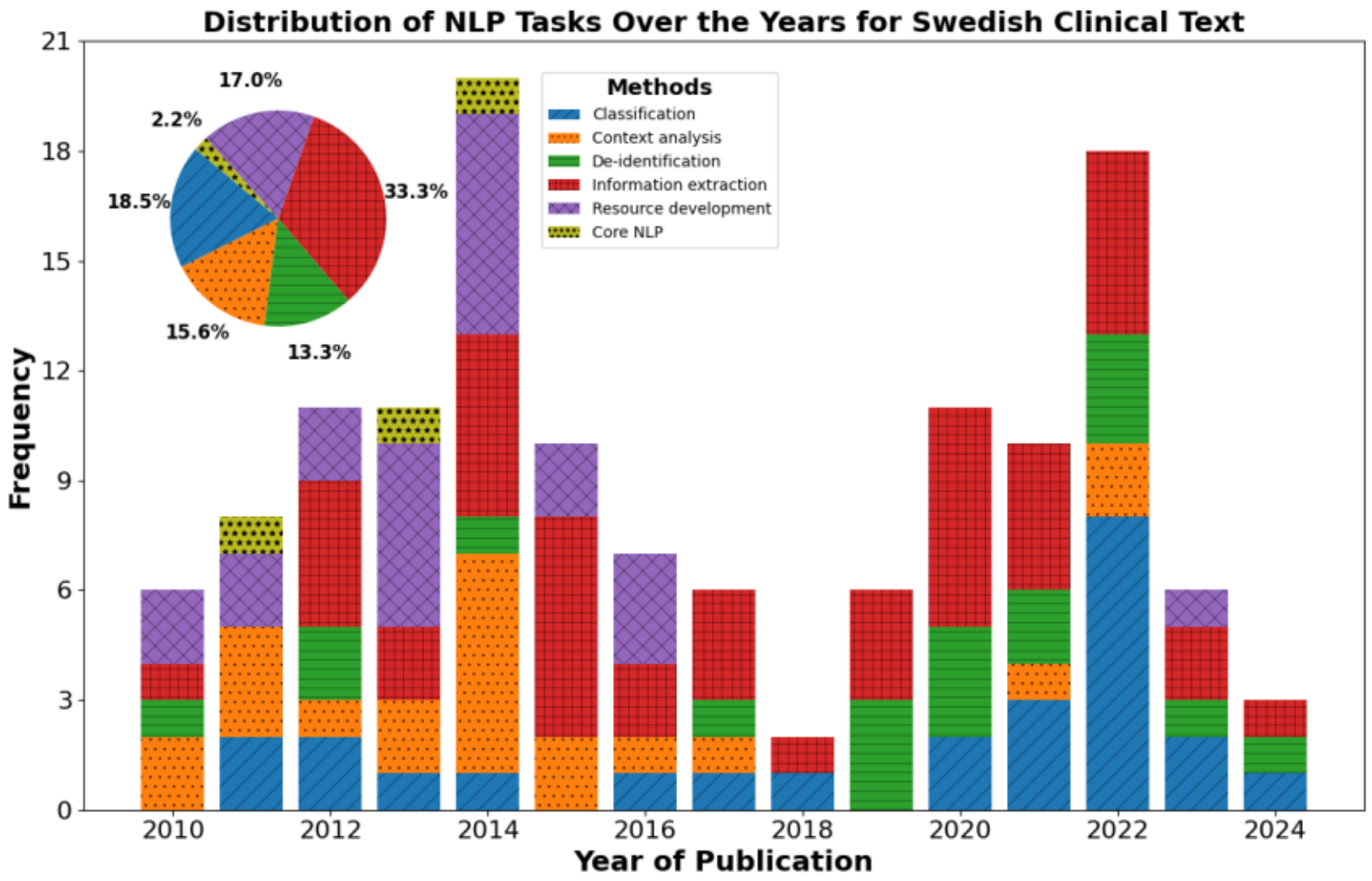

*Figure 5: Distribution of NLP tasks for Swedish clinical text.*

*Danish*: Both information extraction (32%, n = 8) and classification (32%, n = 8) constitute a significant portion of all NLP tasks considered for Danish clinical text, followed by resource development (24%, n = 6) tasks. Like the studies focusing on Norwegian clinical text, de-identification (8.0%, n = 2) receives little attention. There is little activity focusing on context analysis (4.0%, n = 1). No research has been dedicated to core NLP tasks for Danish clinical text.

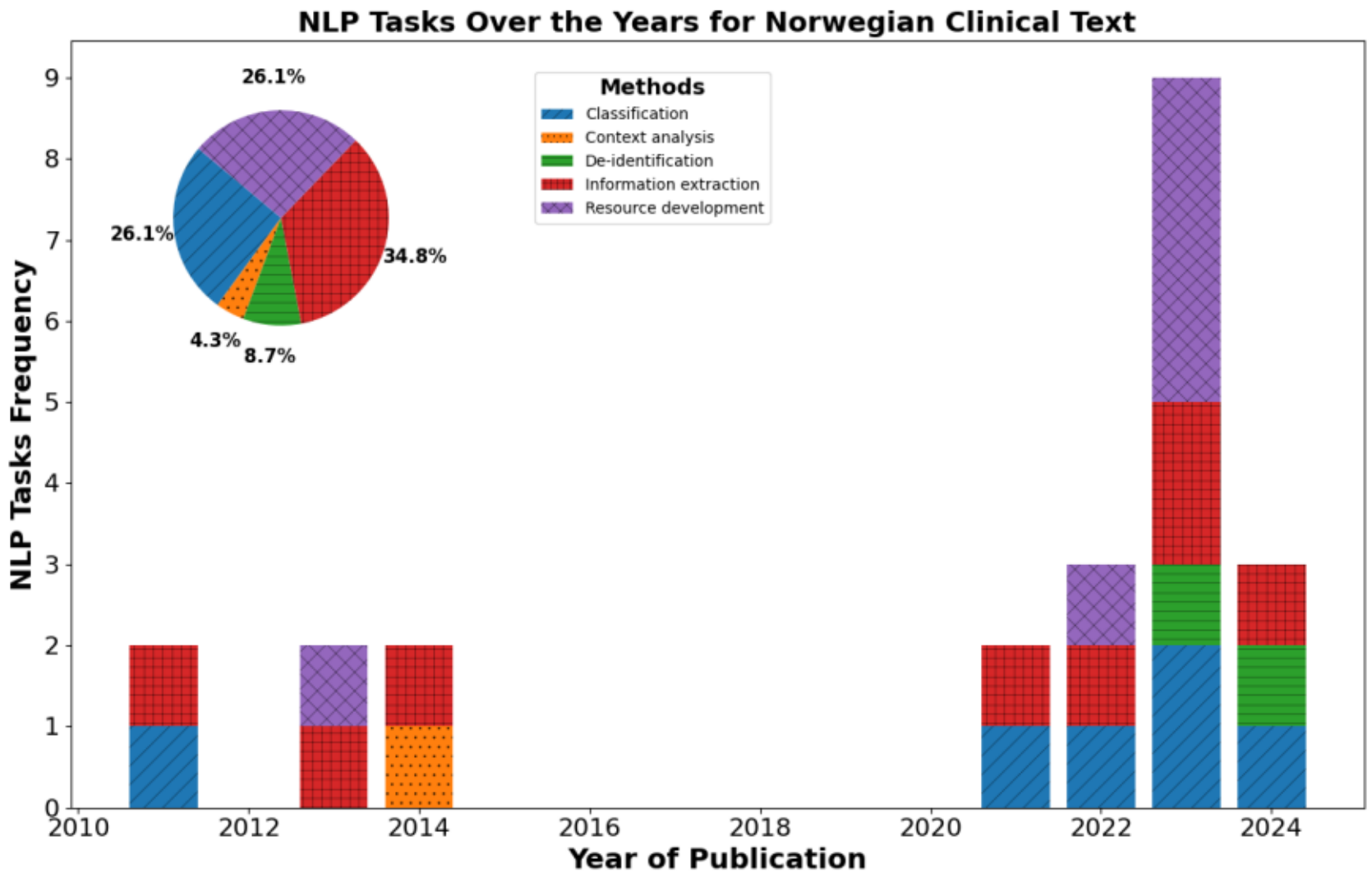

*Figure 6: Distribution of NLP tasks for Danish clinical text.*

- ***Analysis of model/methods usage in various Natural Language Processing Tasks***

*Deep/Nonlinear learning methods* are highly used in information extraction (48.8%, n = 21) and classification (44.2%, n = 19) tasks, and are also employed in de-identification tasks (7.0%, n = 3) to a lesser extent. *Ensemble methods* are mostly adopted in information extraction tasks (68.8%, n = 11) and have also been used in classification (25.0%, n = 4), and de-identification tasks (6.25%, n = 1) to a lesser extent. *Hybrid methods* have little appearance compared to the other models, and appeared in classification (40.0%, n = 2), de-identification (20.0%, n = 1), and information extraction (40.0%, n = 2). The reason behind the less visibility of hybrid methods in these NLP tasks could be related to its late adoption in the mainland Scandinavian clinical domain, which starts its visibility in 2023 and onwards. *Interpretable methods* are dominantly used in classification tasks (57.5%, n = 27), emphasizing the importance of model transparency and explainability, where understanding the model's decision-making process is vital. It also appeared significantly in the information extraction task (40.4%, n = 19), but to a lesser extent in the context analysis task (2.2%, n = 1). *Rule-based systems* are widely used across several NLP tasks, however, with a noticeable presence in context analysis (29.11%, n = 23), information extraction (30.4%, n = 24), and classification tasks (22.8%, n = 18). It also appeared in de-identification (10.2%, n = 8) and core NLP tasks (7.59%, n = 6) but to a lesser extent. *Traditional/Statistical learning methods* appeared across all the NLP tasks, with a high usage rate in the information extraction task (47.1%, n = 48.), and a noticeable presence in classification (17.6%, n = 18), context analysis (18.6%, n = 19), and de-identification tasks (14.7%, n = 15), but a reduced presence in core NLP task (2.0%, n = 2). The uniform presence of rule-based systems and traditional/Statistical learning methods across all the NLP tasks could be explained by early adoption and usage, which started during the initial period, and also the need to incorporate domain knowledge in hand-creating and engineering the features. *Transformers and large language models* are highly employed in classification tasks (44.4%, n = 28) and have a significant presence in information extraction (33.3%, n = 21), and de-identification tasks (17.5%, n = 11). These models also have appeared in context analysis tasks (4.8%, n = 3) but to a much lesser extent compared to the other tasks.

Table 5: Utilization of various models in clinical NLP tasks.

| *Methos* | Classification | Context analysis | Core NLP | De-identification | Information extraction |
|---|---|---|---|---|---|
| *Deep/Nonlinear Learning Methods* | 44.2% (19) | - | - | 7.0% (3) | 48.8% (21) |
| *Ensemble Methods* | 25.0% (4) | - | - | 6.25% (1) | 68.8% (11) |
| *Hybrid Methods* | 40.0% (2) | - | - | 20.0% (1) | 40.0% (2) |
| *Interpretable Methods* | 57.5% (27) | 2.2% (1) | - | - | 40.4% (19) |
| *Rule-Based Systems* | 15.3% (11) | 31.9% (23) | 8.3% (6) | 11.1% (8) | 33.3% (24) |
| *Traditional/Statistical Learning Methods* | 17.6% (18) | 18.6% (19) | 2.0% (2) | 14.7% (15) | 47.1% (48) |
| *Transformers and Large Language Models* | 39.7% (23) | 5.2% (3) | - | 19.0% (11) | 36.2% (21) |

- ***Distribution of NLP tasks – Type Description***

This section presents a detailed analysis of *NLP task type descriptions*, which are sub-categories of each of the NLP tasks described above. Figure 7 below depicts the overall distribution of the *NLP task type description* for the mainland Scandinavian clinical texts. As can be seen from the figure, the majority of the research focuses on Named Entity Recognition-NER (19.4%), followed by risk stratification from EHR text (phenotyping from EHR text) (12.6%). Satisfactory efforts have been made to corpora and annotation (8.6%), though much of the efforts came from studies focusing on Swedish clinical text. Findings/Symptoms classification and medical coding account for 7.2% of the research activities. Drug/Adverse events have also been given satisfactory attention. However, little attention has been given to resource development, especially for Norwegian and Danish clinical texts.

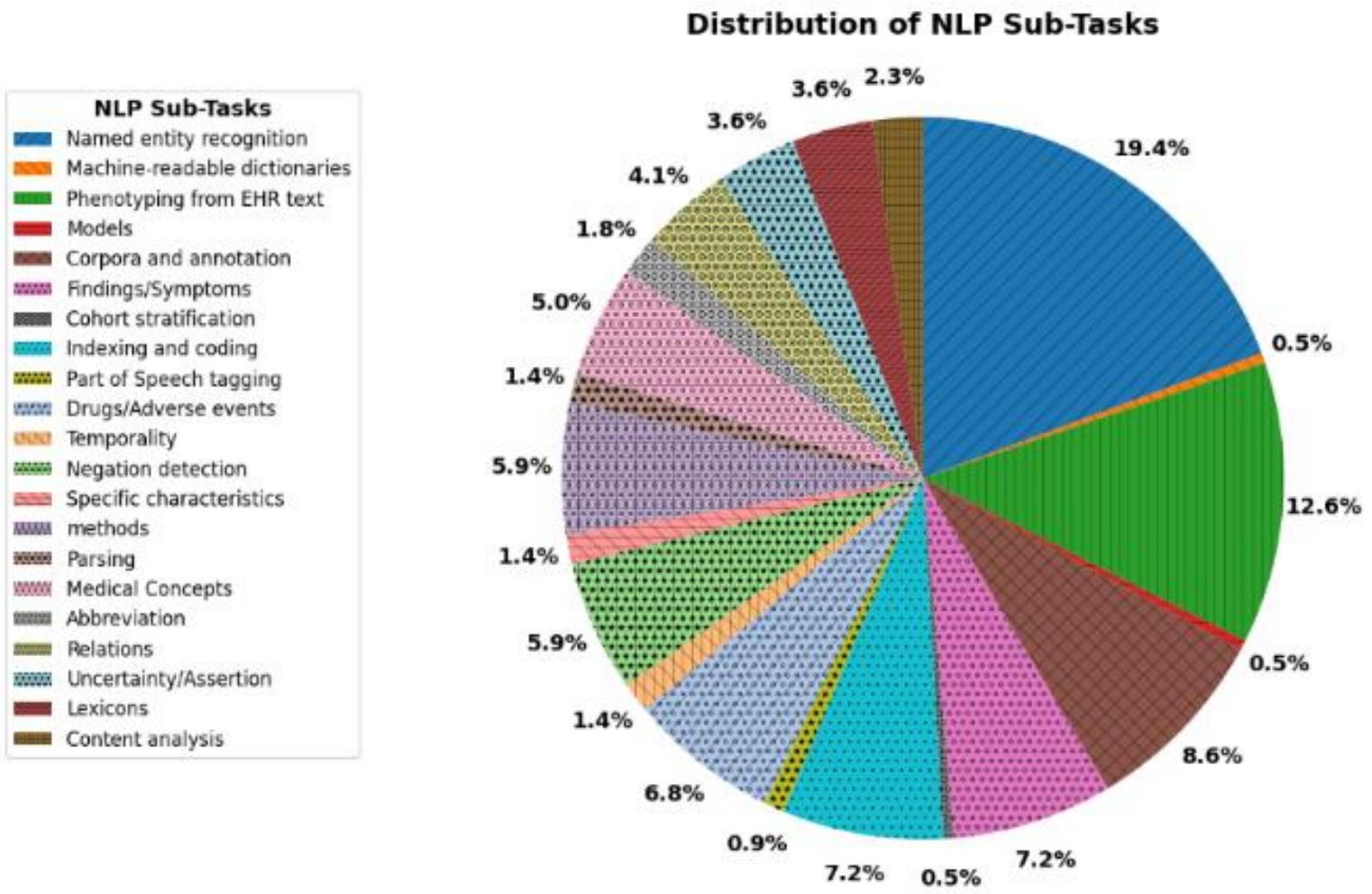

*Figure 7: Overall distribution of NLP tasks for the reviewed studies.*

- ***Hospital department/Clinical Unit distribution***

The following figures depict the distribution of the different hospital departments/clinical units targeted by the reviewed studies. Studies that didn't report this information are grouped under the "General services"; other department/clinical units that didn't fit into the available categories are grouped under the "Other specialties" category. As can be seen from the figure, Surgical specialties (22.8%) take up the majority followed by Medical specialties (19.2%). Intensive and critical care ranked third accounting for 11%, Diagnostic and therapeutic services, and Other specialties also got fair attention.

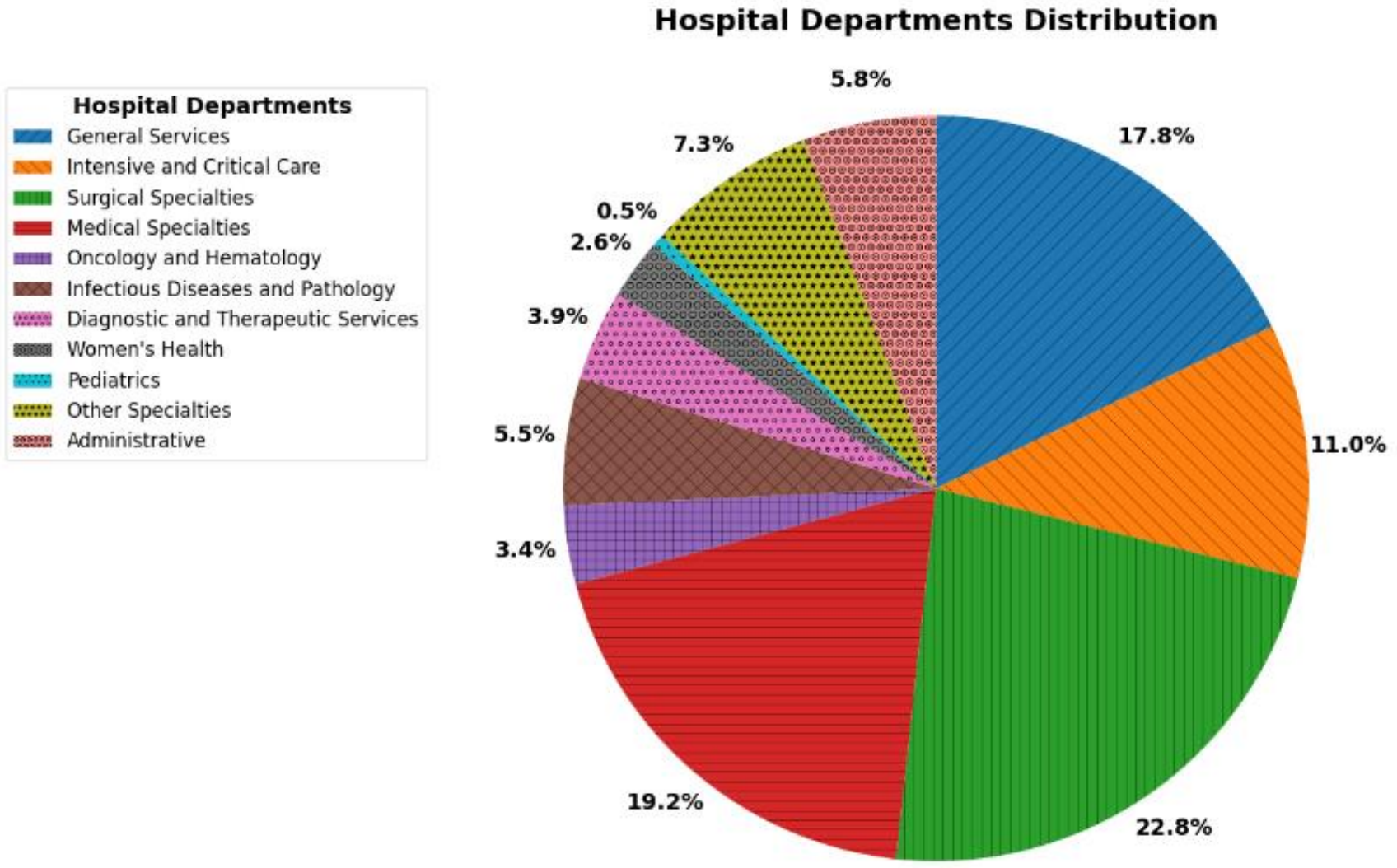

*Figure 8: Overall distribution of hospital departments/clinical units targeted by the reviewed studies.*

On a similar note, surgical specialties (29.8%) take the highest majority for studies focusing on Norwegian clinical text, followed by administrative documents accounting for 17.5%. Intensive and critical care (15.8%), along with medical specialties (14.0%) also take a significant proportion.

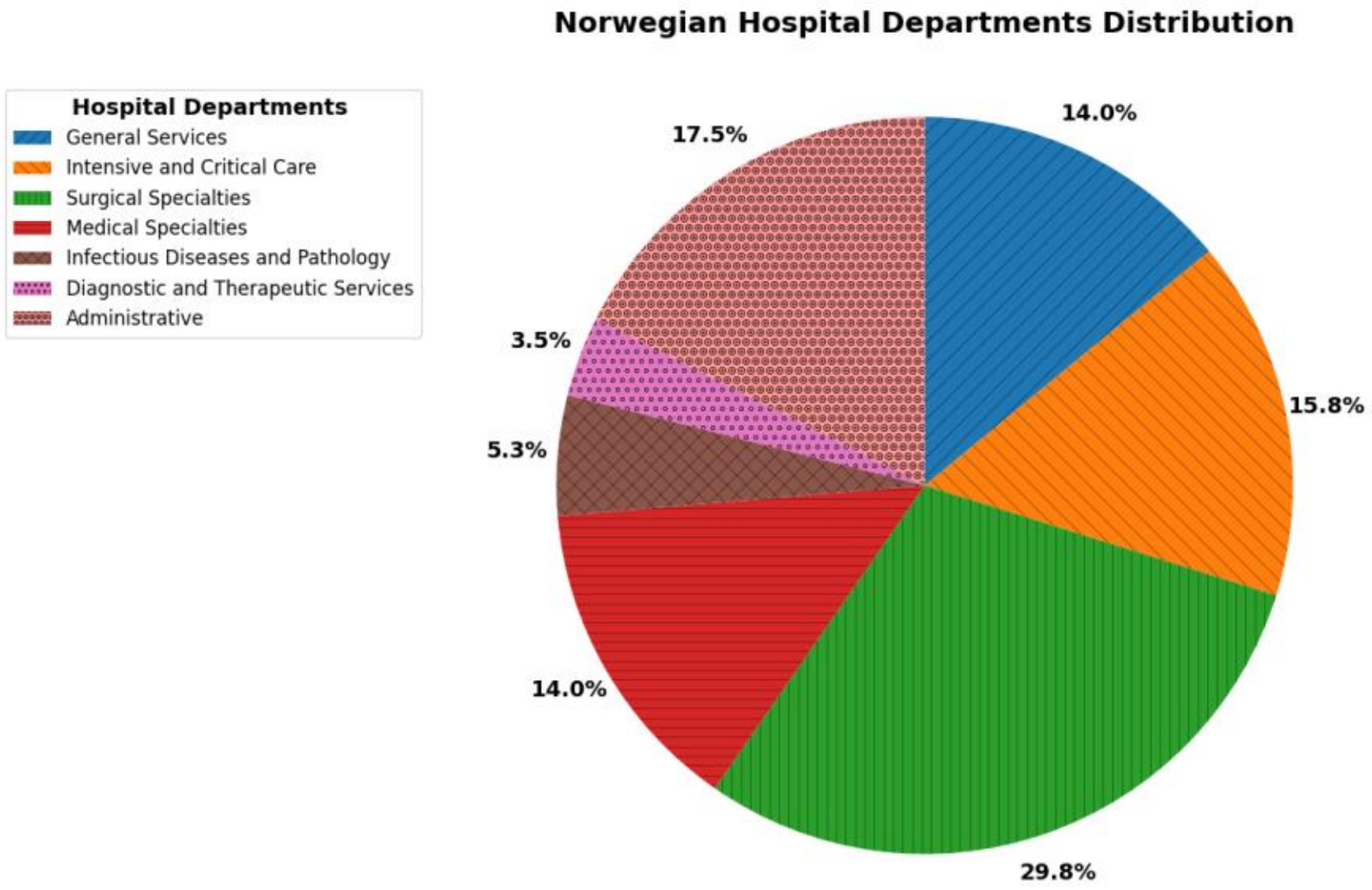

*Figure 9: Distribution of Norwegian hospital departments/clinical units targeted by the reviewed studies.*

For the Swedish clinical text, similar to the Norwegian clinical text, surgical specialties (20.3%) take up a significant proportion followed by medical specialists (18.1%). Intensive and critical care ranked fourth, accounting for 11.2%.

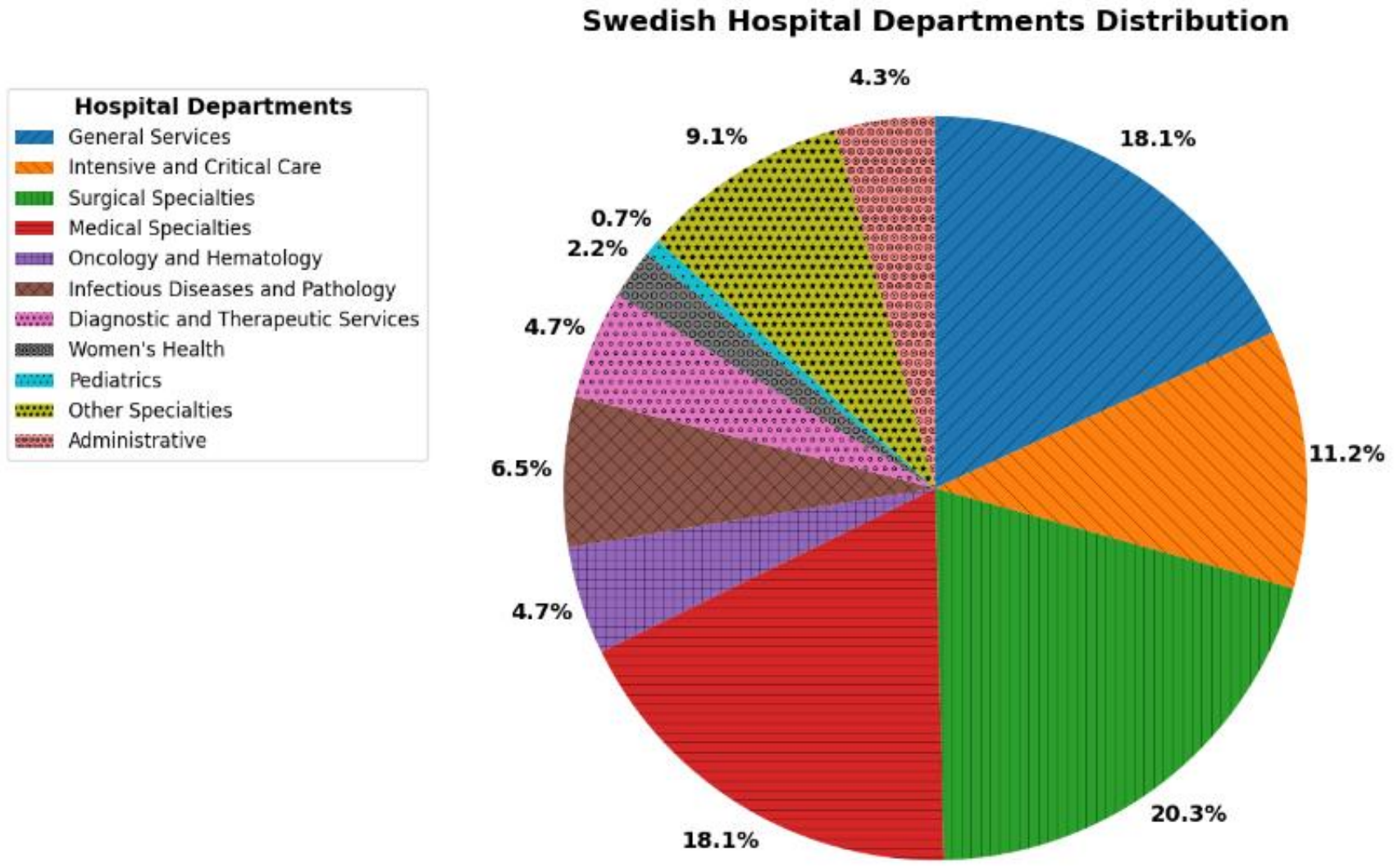

*Figure 10: Distribution of Swedish hospital departments/clinical units targeted by the reviewed studies.*

For Danish clinical text, surgical specialties (34.6%) and medical specialties (28.8%) cover the significant proportions. Unlike the other two, the review identified the utilization of data from women's health, which accounts for 7.7%. Compared to the other two – Norwegian and Swedish clinical text, this group seems to incorporate less diversified categories.

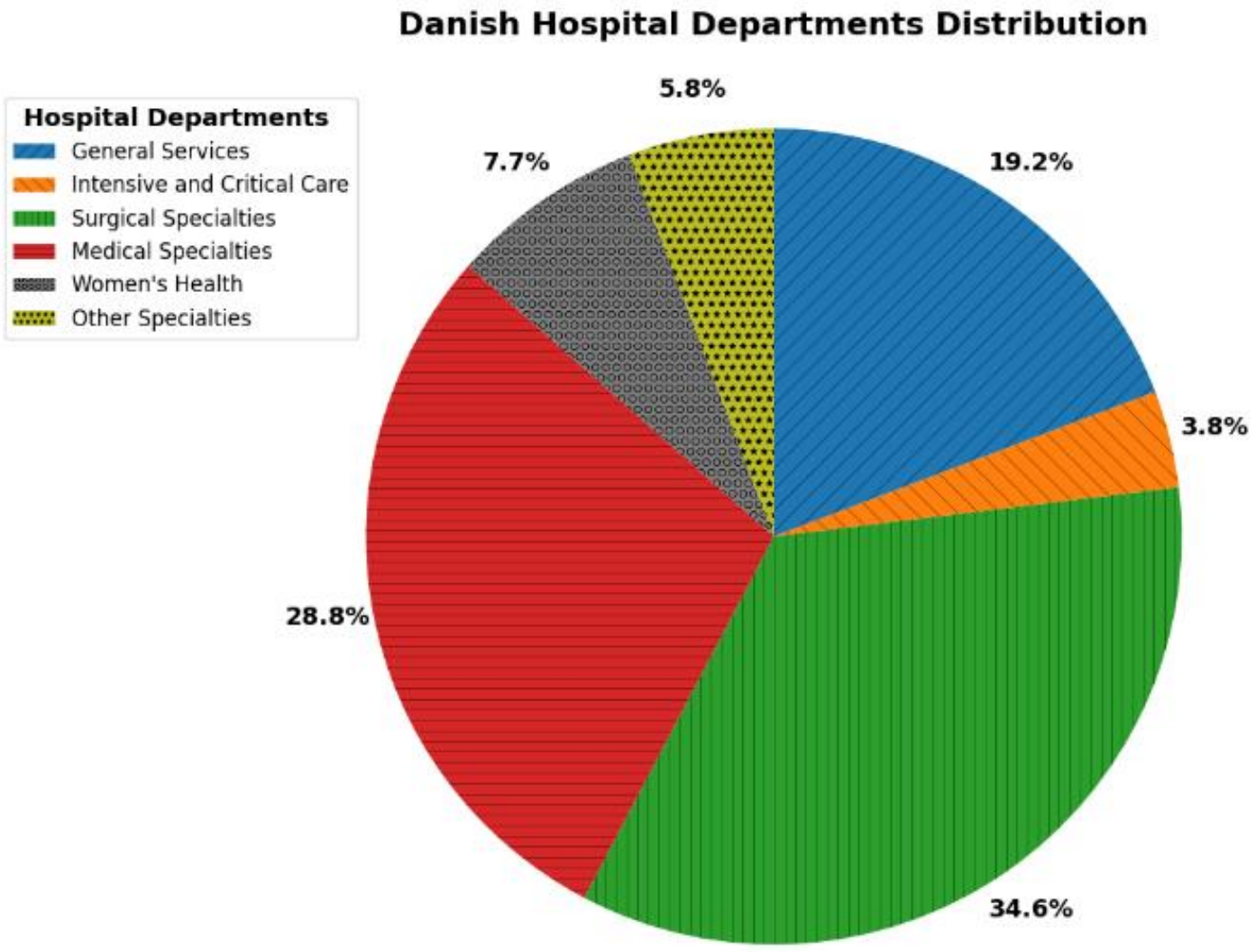

*Figure 11: Distribution of Danish hospital departments/clinical units targeted by the reviewed studies.*